\documentclass[manuscript,screen,acmsmall]{acmart}
\usepackage{amsmath}
\usepackage{acronym}
\usepackage{algorithmic}
\usepackage{booktabs}
\usepackage{multirow}

\usepackage{amssymb} 
\usepackage{enumitem}
\usepackage{adjustbox}
\usepackage{xspace}
\usepackage{hyperref}
\usepackage{natbib}
\usepackage{graphicx}
\usepackage{float}
\usepackage{subfigure}
\usepackage{url}
\usepackage{todonotes}
\usepackage{xcolor}
\usepackage{comment}
\usepackage{tikz}
\usepackage{pgfplots}
\usepackage{pgfplotstable}
\usepackage{makecell}
\usepackage{adjustbox}
\usepackage{pifont}
\usepackage{float}
\usepackage[edges]{forest}
\definecolor{hidden-draw}{RGB}{0,0,0}
\definecolor{hidden-pink}{RGB}{168,191,143}

\AtBeginDocument{%
  \providecommand\BibTeX{{%
    \normalfont B\kern-0.5em{\scshape i\kern-0.25em b}\kern-0.8em\TeX}}}

\setcopyright{acmcopyright}
\copyrightyear{2018}
\acmYear{2018}
\acmDOI{XXXXXXX.XXXXXXX}

\acmJournal{CSUR}

\begin{document}

\title[Survey on Factuality in Large Language Models]{Survey on Factuality in Large Language Models: Knowledge, Retrieval and Domain-Specificity}

\author{Cunxiang Wang}\orcid{0000-0002-3023-8082}
\authornote{The first three authors contribute equally.}
\email{wangcunxiang@westlake.edu.cn}
\affiliation{%
  \institution{Westlake University, Zhejiang University}
  \city{Hangzhou}
  \country{China}
}

\author{Xiaoze Liu}\orcid{0000-0002-9726-3397}
\authornotemark[1]
\affiliation{%
  \institution{Purdue University}
  \state{IN}
  \country{USA}
}

\author{Yuanhao Yue}\orcid{0009-0008-3612-5805}
\authornotemark[1]
\affiliation{%
  \institution{Fudan University}
  \city{Shanghai}
  \country{China}
}

\author{Qipeng Guo}\orcid{0000-0002-8805-8789}
\author{Xiangkun Hu}\orcid{0009-0006-2165-9236}
\affiliation{%
  \institution{Amazon AWS AI}
  \city{Shanghai}
  \country{China}
}

\author{Xiangru Tang}\orcid{0009-0006-2700-4513}
\affiliation{%
  \institution{Yale University}
  \state{CT}
  \country{USA}
}

\author{Tianhang Zhang}\orcid{0009-0006-6234-4409}
\affiliation{%
  \institution{Shanghai Jiao Tong University}
  \city{Shanghai}
  \country{China}
}

\author{Cheng Jiayang}\orcid{0000-0003-1140-6084}
\affiliation{%
  \institution{The Hong Kong University of Science and Technology}
  \city{Hong Kong}
  \country{China}
}

\author{Yunzhi Yao}\orcid{0000-0001-9458-696X}
\affiliation{%
  \institution{Zhejiang University}
  \city{Hangzhou}
  \country{China}
}

\author{Xuming Hu}
\orcid{0000-0001-6075-4224}
\author{Zehan Qi}
\orcid{0009-0007-5232-9130}
\affiliation{%
  \institution{Tsinghua University}
  \city{Beijing}
  \country{China}
}


\author{Wenyang Gao}\orcid{0000-0003-1143-064X}
\author{Yidong Wang}\orcid{0009-0007-9969-8259}
\author{Linyi Yang}\orcid{0000-0003-0667-7349}
\affiliation{%
  \institution{Westlake University}
  \city{Hangzhou}
  \country{China}
}


\author{Jindong Wang}\orcid{0000-0002-4833-0880}
\author{Xing Xie}\orcid{0000-0002-8608-8482}
\affiliation{%
  \institution{Microsoft Research}
  \city{Beijing}
  \country{China}
}


\author{Zheng Zhang}\orcid{0009-0007-2007-2019 }
\affiliation{%
  \institution{Amazon AWS AI}
  \city{Shanghai}
  \country{China}
}

\author{Yue Zhang}\orcid{0000-0002-5214-2268}
\email{zhangyue@westlake.edu.cn}
\affiliation{%
  \institution{Westlake University}
  \city{Hangzhou}
  \country{China}
}

\renewcommand{\shortauthors}{Wang, et al.}


\begin{abstract}
This survey addresses the crucial issue of factuality in Large Language Models (LLMs). As LLMs find applications across diverse domains, the reliability and accuracy of their outputs become vital. We define the ``factuality issue" as the probability of LLMs to produce content inconsistent with established facts. We first delve into the implications of these inaccuracies, highlighting the potential consequences and challenges posed by factual errors in LLM outputs. Subsequently, we analyze the mechanisms through which LLMs store and process facts, seeking the primary causes of factual errors. Our discussion then transitions to methodologies for evaluating LLM factuality, emphasizing key metrics, benchmarks, and studies. We further explore strategies for enhancing LLM factuality, including approaches tailored for specific domains. We focus two primary LLM configurations—standalone LLMs and Retrieval-Augmented LLMs that utilizes external data—we detail their unique challenges and potential enhancements. Our survey offers a structured guide for researchers aiming to fortify the factual reliability of LLMs. We consistently maintain and update the related open-source materials at \url{https://github.com/wangcunxiang/LLM-Factuality-Survey}. 
\end{abstract}

\ccsdesc[500]{Computing methodologies~Natural language generation}




\keywords{ Factuality, factuality in LLM, retrievel augmented LLM, domain factuality enhanced LLM}


\maketitle

\section{Introduction}
\definecolor{mycolor}{RGB}{215, 245, 200}

\tikzstyle{my-box}=[
    rectangle,
    draw=hidden-draw,
    rounded corners,
    text opacity=1,
    minimum height=1.5em,
    minimum width=5em,
    inner sep=2pt,
    align=center,
    fill opacity=.5,
    line width=0.8pt,
]
\tikzset{
leaf/.style={
my-box,
minimum height=1.5em,
fill=mycolor, 
text=black,
align=left,
font=\footnotesize,
inner xsep=2pt,
inner ysep=4pt,
line width=0.8pt
}
}
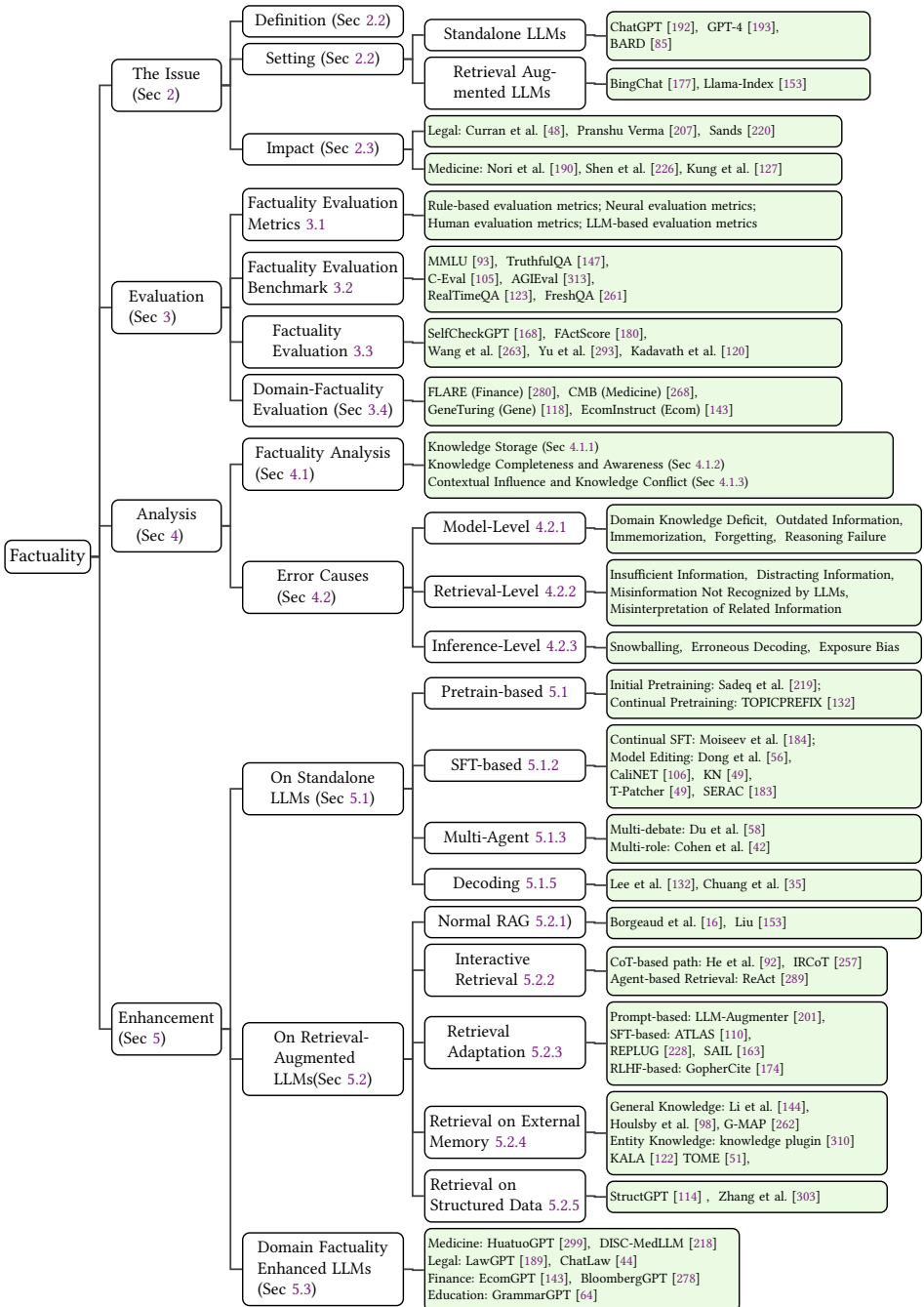
\begin{figure*}[t!]
    \centering
    \begin{adjustbox}{width=0.9\textwidth}
        \begin{forest}
            forked edges,
            for tree={
                grow=east,
                reversed=true,
                anchor=base west,
                parent anchor=east,
                child anchor=west,
                base=center,
                font=\large,
                rectangle,
                draw=hidden-draw,
                rounded corners,
                align=left,
                text centered,
                minimum width=5em,
                edge+={darkgray, line width=1pt},
                s sep=3pt,
                inner xsep=2pt,
                inner ysep=3pt,
                line width=0.8pt,
                ver/.style={rotate=90, child anchor=north, parent anchor=south, anchor=center},
            },
            where level=1{text width=6em,font=\normalsize,}{},
            where level=2{text width=9em,font=\normalsize,}{},
            where level=3{text width=9em,font=\normalsize,}{},
            where level=4{text width=7em,font=\normalsize,}{},
            [
                Factuality
                [
                    The Issue\\(Sec \ref{sec:the_issue})
                    [
                        Definition (Sec \ref{sec:definition})
                    ]
                    [
                        Setting (Sec \ref{sec:definition})
                        [
                            Standalone LLMs
                            [
                                ChatGPT~\citep{chatgpt}{, } GPT-4~\citep{GPT4}{, }\\BARD \citep{bard}\\
                                 , leaf, text width=15em
                            ]
                        ]
                        [
                            Retrieval Aug-\\mented LLMs
                            [
                                BingChat \citep{BingChat}{,} Llama-Index \citep{LlamaIndex}
                                , leaf, text width=15em
                            ]
                        ]
                    ]
                    [
                        Impact (Sec \ref{sec:impact})
                        [
                            Legal: \citet{curran2023hallucination}{, } \citet{Pranshu2023}{, } \citet{news_mayor}
                                , leaf, text width=24em
                        ]
                        [
                            Medicine: \citet{nori2023capabilities}{,} \citet{shen2023chatgpt}{,} \citet{USMLE_evaluation}
                                , leaf, text width=24em
                        ]
                    ]
                ]
                [
                    Evaluation\\(Sec \ref{sec:evalution})
                    [
                        Factuality Evaluation\\Metrics \ref{sec:eval_metrics}
                        [
                            Rule-based evaluation metrics;
                            Neural evaluation metrics; \\
                            Human evaluation metrics; LLM-based  evaluation metrics, leaf, text width=24em
                        ]
                    ]
                    [
                        Factuality Evaluation\\Benchmark \ref{sec:eval_benchmark}
                        [
                            MMLU~\citep{MMLU}{, } TruthfulQA~\citep{TruthfulQA}{, }\\C-Eval~\citep{C-Eval}{, } AGIEval~\citep{AGIEval}{, } \\ RealTimeQA~\citep{kasai2022realtimeqa}{, } FreshQA~\citep{vu2023freshllms}, leaf, text width=24em
                        ]
                    ]
                    [
                        Factuality\\Evaluation \ref{sec:factuality-eval}
                        [
                            SelfCheckGPT~\citep{SelfCheckGPT}{, } FActScore~\citep{Min2023-FActScore}{, } \\ \citet{wang2023evaluating}{, } \citet{yu2023generate}{, } \citet{kadavath2022language}, leaf, text width=24em
                        ]
                    ]
                    [
                        Domain-Factuality\\Evaluation (Sec \ref{sec:domain-eval})
                        [ 
                            FLARE (Finance) \citep{xie2023pixiu}{, } CMB (Medicine) \citep{wang2023cmb}{, }\\GeneTuring (Gene) \citep{jin2023genegpt}{, } EcomInstruct (Ecom) \citep{li2023ecomgpt}, leaf, text width=24em
                        ]
                    ]
                ]
                [
                    Analysis\\
                    (Sec \ref{sec:analysis})
                    [
                        Factuality Analysis\\
                        (Sec \ref{sec:factuality_analysis})
                        [
                            Knowledge Storage (Sec \ref{sec:analy_store})\\
                            Knowledge Completeness and Awareness (Sec \ref{sec:analy_aware})\\
                            Contextual Influence and Knowledge Conflict (Sec \ref{sec:analy_confict}), leaf, text width=27em 
                        ]
                    ]
                    [
                        Error Causes\\(Sec \ref{sec:causes})
                        [
                            Model-Level \ref{sec:model-causes}
                            [
                                Domain Knowledge Deficit{, } Outdated Information{, }\\ Immemorization{, } Forgetting{, } Reasoning Failure, leaf, text width=18em
                            ]
                        ]
                        [
                            Retrieval-Level \ref{sec:retrieval-causes}
                            [
                                Insufficient Information{, } Distracting Information{, } \\ Misinformation Not Recognized by LLMs{, }\\
                                Misinterpretation of Related Information, leaf, text width=18em
                            ]
                        ]
                        [
                            Inference-Level \ref{sec:inference-causes}
                            [
                                Snowballing{, } Erroneous Decoding{, } Exposure Bias, leaf, text width=18em
                            ]
                        ]
                    ]
                ]
                [
                    Enhancement\\ (Sec \ref{sec:enhancement})
                    [
                        On Standalone\\LLMs
                        (Sec \ref{sec:enhace_pure})
                        [
                            Pretrain-based
                            \ref{sec:enhace_pure}
                            [
                                Initial Pretraining: \citet{sadeq-etal-2023-unsupervised}{; }\\
                                Continual Pretraining: TOPICPREFIX \citep{lee2022factuality}\\
                                , leaf, text width=18em
                            ]
                        ]
                        [
                            SFT-based \ref{sec:enhace_pure_sft}
                            [
                                Continual SFT: \citet{moiseev-etal-2022-skill}{;}\\
                                Model Editing: \citet{dong-etal-2022-calibrating}{, }\\ CaliNET \citep{huang2023transformerpatcher}{, }  KN \citep{dai-etal-2022-knowledge}{, }\\ T-Patcher \citep{dai-etal-2022-knowledge}{, }
                                SERAC \citep{Mitchell2022MemoryBasedME}\\
                                , leaf, text width=18em
                            ]
                        ]
                        [
                            Multi-Agent \ref{sec:enhace_pure_multi}
                            [
                                Multi-debate: \citet{multiagent_debate}\\
                                Multi-role: \citet{cohen2023lm}\\
                                , leaf, text width=18em
                            ]
                        ]
                        [
                            Decoding \ref{sec:enhace_pure_decoding}
                            [
                                \citet{lee2022factuality}{,} \citet{chuang2023dola}\\
                                , leaf, text width=18em
                            ]
                        ]
                    ]
                    [
                        On Retrieval-\\Augmented\\LLMs(Sec \ref{sec:enhace_rag})
                        [
                            Normal RAG
                            \ref{sec:enhace_rag_normal})
                            [
                                \citet{borgeaud2022improving}{, } \citet{LlamaIndex} \\
                                , leaf, text width=18em
                            ]
                        ]
                        [
                            Interactive\\Retrieval \ref{sec:enhace_rag_inter}
                            [
                                CoT-based path: \citet{he2022rethinking}{, } IRCoT \citep{trivedi-etal-2023-interleaving} \\
                                Agent-based Retrieval: ReAct \citep{yao2023react}
                                , leaf, text width=16em
                            ]
                        ]
                        [
                            Retrieval\\Adaptation  \ref{sec:enhace_rag_adapt}
                            [
                                Prompt-based: LLM-Augmenter \citep{peng2023check}{, } \\
                                SFT-based: ATLAS \citep{Atlas}{, } \\REPLUG \citep{REPLUG}{, } SAIL \citep{luo2023sail}  \\
                                RLHF-based: GopherCite \citep{GopherCite}\\
                                , leaf, text width=16em
                            ]
                        ]
                        [
                            Retrieval on External \\ Memory \ref{sec:enhace_rag_mem}
                            [
                                 General Knowledge: \citet{li2022decoupled}{,}\\ \citet{houlsby2019parameterefficient}{,} G-MAP \citep{wan2022gmap}  \\
                                Entity Knowledge: knowledge plugin \citep{zhang-etal-2023-plug}\\KALA \citep{KALA} TOME \citep{jong2022mention}{, }  \\
                                , leaf, text width=16em
                            ]
                        ]
                        [
                            Retrieval on\\Structured Data
                            \ref{sec:enhace_rag_kg}
                            [
                                StructGPT \citep{StructGPT} {, } \citet{zhang2023mitigating}
                                , leaf, text width=16em
                            ]
                        ]
                    ]
                    [
                        Domain Factuality\\Enhanced LLMs
                        \\(Sec \ref{sec:domain_llms})
                        [
                            Medicine: HuatuoGPT \citep{zhang2023huatuogpt}{, } DISC-MedLLM \citep{ross2022large}\\
                            Legal: LawGPT \citep{nguyen2023brief}{, } ChatLaw \citep{cui2023chatlaw}\\
                            Finance: EcomGPT \citep{li2023ecomgpt}{, } BloombergGPT \citep{BloombergGPT} \\
                            Education: GrammarGPT \citep{fan2023grammargpt}\\
                                , leaf, text width=18em
                        ]
                    ]
                ]
            ]
        \end{forest}
    \end{adjustbox}
        \vspace{-4mm}
    \caption{Taxonomy of research on factuality in Large Language Models that consists of the issue, evaluation, analysis and enhancement.}
    \label{fig:tree}
\end{figure*}
The quest for mastery of knowledge has been a foundational aspiration in the development of artificial intelligence systems. Historically, seminal works by \citet{McCarthy1963} and \citet{Newell1976} have underscored the significance of knowledge representation and reasoning in AI systems. For instance, the Cyc project embarked on an ambitious journey to codify common-sense knowledge, aiming to provide AI systems with a comprehensive understanding of the world \citep{Lenat1995}. Concurrently, endeavors like the WordNet project by \citet{Miller1990} sought to create lexical databases that capture semantic relationships between words, thereby aiding AI systems in grasping the nuances of human language. 

Amidst these pioneering efforts, the emergence of Large Language Models (LLMs), such as ChatGPT \citep{chatgpt}, GPT-4 \citep{GPT4} and LLaMA \citep{llama,llama2}, has been seen as a significant leap in both academics and industries, especially towards AI systems possessing vast factual knowledge \citep{LAMA,de-cao-etal-2021-editing,GPT4}. 
The advantages of using LLMs as knowledge bases carriers are manifold. Firstly, they reduce the overhead and costs associated with building and maintaining dedicated knowledge bases \citep{petroni2019language,alkhamissi2022review,KSSM}. Additionally, LLMs offer a more flexible approach to knowledge processing and utilization, allowing for context-aware reasoning and the ability to adapt to novel information or prompts \citep{sun2023beamsearchqa,huang2023reasoning}. 
Yet, with their unparalleled capabilities, concerns have arisen about the potential of LLMs to generate non-factual or misleading content \citep{Bender2021,GPT4,bubeck2023sparks}. In light of these advancements and challenges, this survey seeks to delve deeply into the LLMs, exploring both their potential and the concerns surrounding their factual accuracy.

Understanding the factuality of Large Language Models is more than just a technical challenge; it's essential for the responsible use of these tools in our daily lives. As LLMs become more integrated into services like search engines \citep{BingChat}, chatbots \citep{chatgpt,bard}, and content generators \citep{ChatLaw}, the information they provide directly influences decisions, beliefs, and actions of millions of people. If an LLM provides incorrect or misleading information, it can lead to misunderstandings, spread false beliefs, or even cause harm, especially for those domains that demand high factual accuracy~\citep{ling2023beyond}, such as health ~\citep{thirunavukarasu2023large, tang-etal-2023-aligning}, law~\citep{huang2023lawyer}, and finance~\citep{BloombergGPT}. For instance, a physician relying on an LLM for medical guidance might inadvertently jeopardize patient health, a corporation leveraging LLM insights might make ill-informed market decisions, or an attorney misinformed by an LLM might falter in legal proceedings \citep{curran2023hallucination}.
In addition, with the advancement of LLM-based agents, the factuality of LLMs is becoming even more potent. A driver or an autonomous driving car might rely on LLM-based agents for planning or driving, where serious factual mistakes made by LLMs could cause irreversible damage.
By studying the factuality of LLMs, we aim to ensure that these models are both powerful and trustworthy.

A surge of research has been directed towards evaluating LLMs' factuality, which encompasses diverse tasks like factoid question answering and fact checking. Beyond evaluation, efforts to improve the factual knowledge of LLMs have been notable. Strategies have ranged from retrieving information from external knowledge bases to continual pretraining and supervised finetuning. Yet, despite these burgeoning efforts, a holistic overview that covers the full spectrum of factuality in LLMs remains elusive. 
While there are existing surveys in the field, such as those by \citet{chang2023survey} and \citet{aligning_llm_human}, that delve into the evaluation of LLMs and their factuality, they only scratch the surface of the broader landscape. There are also a bunch of recent studies focusing on hallucinations in LLMs \citep{rawte2023survey,zhang2023sirens,rawte2023survey,ye2023cognitive,Huang2023ASO}. But we differentiate between the hallucination issue and the factuality issues in Sec \ref{sec:definition}. Moreover, these surveys often overlook key areas we emphasize, like domain-specific factuality or the challenge of outdated information. While \citet{ling2023domain} explores domain specialization in LLMs, our survey takes a more expansive look at the broader issues of factuality.
To the best of our understanding, our work is the first comprehensive study on the factuality of large language models.

This survey aims to offer an exhaustive overview of the factuality studies in LLMs, delving into four key dimensions: Sec 2) The definition and impact of the factuality issue \citep{Pranshu2023,nori2023capabilities}; Sec 3) Techniques for evaluating factuality and its quantitative assessment~\citep{C-Eval, Min2023-FActScore}; Sec 4) Analyzing the underlying mechanisms of factuality in LLMs and identifying the root causes of factual errors~~\citep{liu2023lost, kotha2023understanding}; and Sec 5) Approaches to enhance the factuality of LLMs~\citep{multiagent_debate, he2022rethinking}. Notably, we categorize the use of LLMs into two primary settings: LLMs without external knowledge, such as ChatGPT \citep{chatgpt} and Retrieval-Augmented LLMs, such as BingChat \citep{BingChat}. The complete structure of this survey is illustrated in Figure \ref{fig:tree}.
Through a detailed examination of existing research, we seek to shed light on this critical aspect of LLMs, helping researchers, developers, and users harness the power of these models responsibly and effectively.

\section{Factuality Issue}
\label{sec:the_issue}

In this section, we describe the issue of factuality in large language models, as well as the impact.

\subsection{Large Language Models}
There is no well-accepted and exact definition of large language models in the literature \citep{chang2023survey,LLMSurvey,huang2022towards}. We mainly consider the decoder-only generative pre-trained language modes with emergent abilities, such as ChatGPT \citep{chatgpt} and LLaMA \citep{llama,llama2}. We also include some work that is based on models with encoder-decoder architectures, such as T5 \citep{t5}. We do not talk about work that is only based on the Encoder-only models, such as BERT \citep{bert} and RoBERTa \citep{RoBERTa}, in this survey. To be specific, our survey includes the following LLMs:

\begin{itemize}[topsep=0pt,itemsep=0pt,parsep=0pt,partopsep=0pt,leftmargin=*]
    \item \textbf{General Domain LLMs:}
GPT-2~\citep{GPT2},
GPT-3~\citep{GPT-3}, 
ChatGPT~\citep{chatgpt}, 
GPT-4~\citep{GPT4},
GPT-Neo~\citep{gpt-neo},
OPT~\citep{zhang2022opt}, 
LLaMA~\citep{llama}, 
LLaMA-2~\citep{llama2}, Incite~\citep{together2023redpajama, incite}, 
Claude \citep{Claude},
Falcon~\citep{falcon40b}, 
MPT~\cite{MosaicML2023Introducing}, 
Vicuna~\citep{vicuna2023}, 
FLAN-T5~\cite{chung2022scaling}, BLOOM~\citep{scao2022bloom}, 
Baichuan \&   Baichuan2~\citep{yang2023baichuan}, 
PaLM~\citep{chowdhery2022palm}, 
Gopher~\citep{Gopher}, 
Megatron-LM~\citep{shoeybi2019megatron}, 
SAIL~\citep{luo2023sail}, 
Codex~\citep{chen2021evaluating}, 
Bard~\citep{bard}, 
GLM \&
ChatGLM~\citep{glm},
InternLM~\citep{team2023internlm},
StableBeluga~\citep{StableBelugaModels},
Claude~\citep{Claude},
Alpaca~\citep{alpaca},
New Bing~\citep{BingChat},
Ziya-LLaMA~\citep{fengshenbang},
BLOOMZ~\citep{muennighoff2022crosslingual},
Chinese-LLaMA~\citep{chinese-llama-alpaca},
Phoenix~\citep{phoenix-2023}, and others.

\item \textbf{Domain-specify LLMs:} 
BloombergGPT~\citep{BloombergGPT}, 
EcomGPT~\citep{li2023ecomgpt}, 
BioGPT~\citep{luo2022biogpt},
LawGPT~\citep{nguyen2023brief}, 
Lawyer LLaMA~\citep{huang2023lawyer}, ChatLaw~\citep{ChatLaw}, BioMedLM~\citep{venigalla2022biomedlm}, HuatuoGPT~\citep{zhang2023huatuogpt}, ChatDoctor~\citep{li2023chatdoctor}, MedicalGPT~\citep{MedicalGPT}, 
Bentsao (Huatuo as its original name)~\citep{huatuo}, 
Zhongjing~\citep{yang2023zhongjing},
LLM-AMT~\citep{wang2023augmenting},
DISC-MedLLM~\citep{bao2023disc},
Cohortgpt ~\citep{guan2023cohortgpt},
Deid-gpt~\citep{liu2023deid},
Doctorglm~\citep{xiong2023doctorglm},
MedChatZH~\citep{tan2023medchatzh},
K2~\citep{deng2023learning},
HouYi~\citep{bai2023houyi},
GrammarGPT~\citep{fan2023grammargpt},
FoodGPT~\citep{qi2023foodgpt},
ChatHome~\citep{wen2023chathome},
and others.
\end{itemize}


\subsection{Factuality}
\label{sec:definition}

\begin{table*}[]\small
\centering
\caption{Examples of different kinds of factual error produced by large language models. We category the factual error types by the causes of them, whose details can be found in Sec \ref{sec:causes}}.
\label{tab:causes}
\vspace{-4mm}  \begin{adjustbox}{max width=\textwidth}\setlength{\tabcolsep}{0.3mm}
{
\begin{tabular}{llll}
\toprule
Category &
  Cause &
  Example Dialog &
  Notes and references 
  \\\midrule

   &
  \begin{tabular}[c]{@{}l@{}}Domain \\knowledge\\ deficit \end{tabular} &
  \begin{tabular}[c]{@{}l@{}}
  
  \textbf{Q:} CEO of Assicurazioni Generali?\\
   \midrule
  \textbf{BloombergGPT: }Philippe Donnet\\  \textbf{GPT-NeoX: }Antonio De Lorenzo, Simone Gambarini, Enrico Zanetti\\  \textbf{FLAN-T5-XXL:} John M Forsyth, Christopher K Peters, \{empty string\}
  \\ \midrule
  \end{tabular} &
  \begin{tabular}[c]{@{}l@{}} BloombergGPT is a finance \\  domain-specific language model.\\
    \cite{BloombergGPT}
  \end{tabular} \\
 &\begin{tabular}[c]{@{}l@{}}
  Outdated \\information \end{tabular} &
\begin{tabular}[c]{@{}l@{}}\textbf{Q:} When was Kyiv attacked by Russia?\\ 
\midrule
\textbf{ChatGPT:}As of my last knowledge update in September 2021, Russia \\had not launched an attack on Kyiv.
\\ \midrule
\end{tabular} &
  \begin{tabular}[c]{@{}l@{}}
  Kyiv was attacked by Russia on 25\\
  February 2022. 
  \end{tabular} \\
\multirow{-15}{*}{\begin{tabular}[c]{@{}l@{}}Model-\\level \\causes\end{tabular}} &
  \begin{tabular}[c]{@{}l@{}}Reasoning \\error\end{tabular} &
  \begin{tabular}[c]{@{}l@{}}
  \textbf{Q:} Who is Tom Cruise’s mother?\\ 
    \midrule
  \textbf{A:} Mary Lee Pfeiffer \\
    \midrule
  \textbf{Q:} Who is Mary Lee Pfeiffer’s son?\\ 
    \midrule
  \textbf{A:} There is no widely known information about...  \\
  \end{tabular} &
 \begin{tabular}[c]{@{}l@{}}
 From \citet{berglund2023reversal}.\\ It is clear that the model knows \\Tom Cruise’s mother is Lee Pfeiffer,\\  but it fails to reason that Lee Pfeiffer\\ has a son named Tom Cruise.
  \end{tabular}
  
   \\ \midrule
\multicolumn{1}{c}{} &
  \begin{tabular}[c]{@{}l@{}}
  Distracted by \\
  the retrieval \\
  information
    \end{tabular}
  &
  \begin{tabular}[c]{@{}l@{}} \textbf{Retrieved context:} \emph{``Alan Melikdjanian - Age, Bio, Faces and Birthday''}\\
 Melikdjanian was the Director, Writer (with his father), \\Editor, and Composer of the 2006 direct-to-DVD \underline{Russian}-language \\film Citizen Mavzik, the story of a Russian immigrant family's \\ assimilation into American culture. The film was produced by \\Vilalan Productions, named after Melikdjanian and his father Vilen.\\
  \midrule
    \textbf{Q:} What is the nationality of the director of Citizen Mavzik?\\
    \midrule
    
    \textbf{A:} (1) The director of the film Citizen Mavzik is Alan Melikdjanian.\\  (2) Alan Melikdjanian is from Russia.  So, the answer is Russia.
    \\\midrule
    \end{tabular} &
    \begin{tabular}[c]{@{}l@{}}
    Alan Melikdjanian, known by the alias \\Captain Disillusion, is a Soviet-born \\\underline{American} independent filmmaker \\and YouTuber~\citep{Captain_disillusion_2018}.\end{tabular} 
   \\
\multicolumn{1}{c}{} &  \begin{tabular}[c]{@{}l@{}}
  Misunderstood \\
  the retrieval \\
  information
    \end{tabular} &
  \begin{tabular}[c]{@{}l@{}}
  \textbf{Retrieved context:} 
  \emph{``Gerard III, Count of Guelders - Wikipedia''}\\   Gerard married Margaretha of Brabant, the daughter of \\
  Duke Hendrik I of Brabant and Matilda of Flanders, Duchess of  \\Brabant, in 1206. Gerard and Margaret's Tomb ...\\ 
   \midrule
  \textbf{Q:} Who is Gerard I, Count Of Guelders's wife?\\\midrule  \textbf{A:} ... the name of Gerard I, Count of Guelders's wife, which is\\ Margaretha of Brabant.
  \\ \midrule
  \end{tabular} &
  \begin{tabular}[c]{@{}l@{}}
  

    The retrieved context accurately \\
    provided the name of Gerard III, \\
    Count of Guelders' wife, which is \\
    Margaretha of Brabant. However, the \\
    model erroneously referred to him as \\
    Gerard I, resulting in an incorrect \\
    answer.\end{tabular}
   \\
\multicolumn{1}{c}{\multirow{-20}{*}{\begin{tabular}[c]{@{}l@{}}Retrieval-\\level \\causes\end{tabular}}} & \begin{tabular}[c]{@{}l@{}}
  Fail to address \\
  misinformation \\
   in retrieved \\documents
    \end{tabular}
    
     &
  \begin{tabular}[c]{@{}l@{}}
  \textbf{Retrieved context:} \\
    
 \emph{{[}0{]} ``James Vincent - Death Notice and Service Information - Legacy.com''}\\   
 James Vincent died on August 9, 2023 in Long Beach, \\California Obituary. James Joseph Vincent was the perfect husband, \\father, grandfather,...\\     
 \emph{{[}1{]} ``UPDATE: Jury convicts James Vincent of murder after deliberating ''}\\
A jury convicted James Vincent of murder in the 2018 \\shooting death of Luka Grabb. The jury reached its verdict at 4:30 \\p.m. Friday.\\
  \midrule
  \textbf{Q:} When did James Vincent die?\\

  \midrule
  \textbf{A:} Based on the information provided in the two articles, James \\Vincent died on August 9, 2023, in Long Beach, California.\end{tabular} &
 \begin{tabular}[c]{@{}l@{}} In the presence of contradictory\\ information from retrieved documents,\\ the model should thoroughly\\ outline the response, while also\\ noting the need for clarity.\\ For instance, the model could state:\\ ``According to the first document, \\James Vincent passed away on August \\9, 2023, in Long Beach. However, the \\second document suggests that\\ James Vincent was murdered in 2018."
   \end{tabular}\\
   \midrule

 &
  Snowballing &
  \begin{tabular}[c]{@{}l@{}}
  \textbf{Q:} Was there ever a US senator who represented the state of New\\ Hampshire and whose alma mater was the University of Pennsylvania?\\ 
  \midrule
  \textbf{A:} Yes... His name was John P. Hale \\

  \midrule
  \end{tabular} 

  &\begin{tabular}[c]{@{}l@{}}
  John P. Hale was graduated from Bowdoin\\ College. \cite{Snowball} \end{tabular} \\
\multirow{-5}{*}{\begin{tabular}[c]{@{}l@{}}Inference-\\level causes\end{tabular}} &
  Exposure bias &
  \begin{tabular}[c]{@{}l@{}}
  \textbf{Q:}
  Aamari was very stoic. {[}PRONOUN{]} rarely showed any emotion.
  \\  \midrule
  \textbf{A:} He.
  \end{tabular} &
  \begin{tabular}[c]{@{}l@{}}
The correct answer was\\ Xe according to \cite{hossain-etal-2023-misgendered}. 
\end{tabular}

  \\\bottomrule
\end{tabular}
}
\end{adjustbox}
\end{table*}
By factuality in LLMs we refer to the capability of large language models for generating contents that follow factual information, which encompasses commonsense, world knowledge and domain facts. The factual information can be grounded to reliable sources, such as dictionaries, Wikipedia or textbooks from different domains \footnote{We only consider cases where the meaning is clear and the truthfulness can be determined in this survey. Furthermore, we only take into account undisputed facts. Minor errors that may exist in reliable sources are not within the scope of our consideration. }. A series of work have discussed whether LLMs can serve as knowledge bases to store factual knowledge \citep{alkhamissi2022review,yu2023generate,pan2023unifying}

\begin{table*}[]\small
\caption{Comparison between the factuality issue and the hallucination issue.}
\centering
\label{tab:comp-hallu}
\vspace{-2mm}
\begin{adjustbox}{max width=\textwidth}
    
  \setlength{\tabcolsep}{0.9mm}
{

\begin{tabular}{ll}
\toprule
Factual and Non-Hallucinated & Factually correct outputs. \\ \midrule
Non-Factual and Hallucinated & Entirely fabricated outputs.  \\  \midrule
Hallucinated but Factual  & 
\begin{tabular}[c]{@{}l@{}} 
1. Outputs that are unfaithful to the prompt but remain factually correct \citep{cao-etal-2022-hallucinated}.  \\
2. Outputs that deviate from the prompt's specifics but don't touch on factuality, e.g.,\\ a prompt asking for a story about a rabbit and wolf becoming friends, but the LLM\\ produces a tale about a rabbit and a dog befriending each other.\\
3. Outputs that provide additional factual details not specified in the prompt, e.g.,\\ a prompt asking about the capital of France, and the LLM responds with "Paris, which\\ is known for the Eiffel Tower."
\end{tabular}\\ \midrule
Non-Factual but Non-Hallucinated &  
\begin{tabular}[c]{@{}l@{}} 
1. Outputs where the LLM states``I don't know" or avoids a direct answer.  \\
2. Outputs that are partially correct, e.g., for the question, "Who landed on the moon\\ with Apollo 11?" If the LLM responds with just "Neil Armstrong," the answer is\\ incomplete but not hallucinated.  \\
3. Outputs that provide a generalized or vague response without specific details, e.g.,\\ for a question about the causes of World War II, the LLM might respond with ``It was\\ due to various political and economic factors."
\end{tabular}
 \\
\bottomrule
\end{tabular}}
\end{adjustbox}
\end{table*}
Existing work focus on measuring factuality in LLMs qualitatively \citep{TruthfulQA,chern2023factool}, discussing the mechanism for storing knowledge \citep{meng2022locating,chen2023journey} and tracing the source of knowledge issues \citep{gou2023critic,kandpal2023large}. The factuality issue for LLMs receive relatively the most attention. Several instances are shown in Table \ref{tab:causes}. 
For instance, an LLM might be deficient in domain-specific factual knowledge, such as medicine or law domain. Additionally, the LLM might be unaware of facts that occurred post its last update. There are also instances where the LLM, despite possessing the relevant facts, fails to reason out the correct answer. In some cases, it might even forget or be unable to recall facts it has previously learned.
The factuality problem is closely related to several hot topics in the field of Large Language Models, including \textbf{Hallucinations} \citep{Hallucination_Survey,LLMSurvey}, \textbf{Outdated Information} \citep{WebGPT,WebCPM}, and \textbf{Domain-Specificity} (e.g., Health \citep{DoctorGLM,huatuo}, Law \citep{ChatLaw}, Finance \citep{BloombergGPT}). At their core, these topics address the same issue: the potential for LLMs to generate contents that contradicts certain facts, whether those contents arise out of thin air, outdated information, or a lack of domain-specific knowledge. 

Therefore, we consider these three topics to fall within the scope of the factuality problem.
However, it is important to note that while these topics are related, they each have a unique focus. 
Both hallucinations and factuality issues in LLMs pertain to the accuracy and reliability of generated content, they address distinct aspects. Hallucinations primarily revolve around LLMs generating baseless or unwarranted content. Drawing from definitions by \citet{GPT4,Hallucination_Survey}, hallucinations can be understood as the model's inclination to "produce content that is nonsensical or untruthful in relation to certain sources." This is different from factuality concerns, which emphasize the model's ability to learn, acquire, and utilize factual knowledge.
To illustrate the distinction:
If an LLM, when prompted to craft "a fairy tale about a rabbit and a wolf making friends," produces a tale about "a rabbit and a dog becoming friends," it's exhibiting hallucination. However, this isn't necessarily a factuality error.
If the generated content contains accurate information but diverges from the prompt's specifics, it's a hallucination but not a factuality issue. For instance, if the LLM's output includes more details or different elements than the prompt specifies but remains factually correct, it's a case of hallucination.
Conversely, if the LLM avoids giving a direct answer, states "I don't know," or provides a response that's accurate but omits some correct details, it's addressing factuality, not hallucination.
Furthermore, it's worth noting that hallucination can sometimes produce content that, while deviating from the original input, remains factually accurate.
For a more structured comparison between the factuality issue and hallucination, refer to Table \ref{tab:comp-hallu}.
Outdated information, on the other hand, focuses on instances where previously accurate information has been superseded by more recent knowledge. Lastly, domain-specificity emphasize the generation of content that requires specific, specialized knowledge. Despite these differences, all three topics contribute to our understanding of the broader factuality problem in LLMs.

 
\paragraph{Setting}
In this survey, our primary focus is on two specific settings:
1. Standard LLMs: Directly using LLMs for answering and chatting \citep{GPT4,chatgpt};
2. Retrieval-Augmented LLMs: The retrieval-augmented generation \citep{BingChat,LlamaIndex}.
The latter is of particular interest as retrieval mechanisms are among the most prevalent methods for enhancing the factuality of LLMs. This involves not just generating accurate responses but also correctly selecting pertinent knowledge snippets from the myriad of retrieved sources.

While summarization tasks—where the goal is to produce summaries that stay true to the source input—have seen research on factuality \citep{maynez-etal-2020-faithfulness,tam-etal-2023-evaluating,tang-etal-2022-confit}, we opted not to focus heavily on this domain in our survey. There are a few reasons for this decision. Firstly, the source inputs for summarization often contain content that is not factual. Secondly, summarization introduces unique challenges like ensuring coherence, conciseness, and relevance, which deviate from the focus of this survey.
It is also worth noting that \citet{pu2023summarization} found LLMs to produce fewer factual errors or hallucinations compared to humans across various summarization benchmarks. However, we will still discuss some works in this area, particularly those that overlap with retrieval settings.

\subsection{Impact}
\label{sec:impact}
The factuality problem significantly impacts the usability of LLMs. Some of these issues have even led to losses at the societal or economic level \citep{Pranshu2023,news_mayor}, drawing the attention of many users, developers, and researchers \citep{Hallucination_Survey}.

Factuality issues also impacted the legal field, with a lawyer in the United States facing sanctions for submitting hallucinated case law in court. One court has mandated that lawyers indicate the portions generated by generative AI in their submitted materials \citep{curran2023hallucination}.
In addition, as part of a research study, a fellow lawyer asked ChatGPT to generate a list of legal scholars with a history of sexual harassment. ChatGPT generated a list that included a law professor. ChatGPT claimed that the professor attempted to touch a student on a class tripnd referenced an article from The Washington Post in March 2018. However, the fact is that this article does not exist, nor does the mentioned class trip \citep{Pranshu2023}.
Besides, a mayor in Australia, discovered false claims made by ChatGPT stating that he was personally convicted of bribery, confessed to charges of bribery and corruption, and received a prison sentence. In response, he plans to initiate legal action against the company responsible for ChatGPT, accusing them of defamation for disseminating untrue information about him. This could potentially be the first defamation case of its kind involving an artificial intelligence chatbot \citep{news_mayor}.

A recent study \citep{nori2023capabilities} provides a comprehensive evaluation of GPT-4's performance in medical competency examinations and benchmark datasets. The evaluation utilizes the text-only version of GPT-4 and investigates its ability to address medical questions without any training or fine-tuning. The assessment is conducted using the United States Medical Licensing Examination (USMLE) \cite{USMLE_evaluation} and the MultiMedQA benchmark \cite{MultiMedQA}, comparing GPT-4's performance against earlier models like GPT-3.5 and models specifically fine-tuned on medical knowledge. The results demonstrate that GPT-4 significantly outperforms its predecessors, achieving scores on the USMLE that exceed the passing threshold by more than 20 points and delivering the best overall performance without specialized prompt crafting or domain-specific fine-tuning. 

While large language models show promise on medical datasets, the introduction of automation in the healthcare field still requires extreme caution \cite{shen2023chatgpt}. Existing metrics and benchmarks are often developed for highly focused problems. Evaluating LLM outputs in supporting real-world decision-making poses challenges \cite{MultiMedQA}, including the stability and robustness of personalized recommendations and inferences in real-world contexts.  Using large language models carries significant risks, including inaccurate ranking recommendations \cite{hirosawa2023diagnostic} (e.g., differential diagnosis) and sequencing \cite{lm_bio} (e.g., information gathering and testing), as well as factual errors \cite{shen2023chatgpt}, particularly important omissions and erroneous responses.

\section{Factuality Evaluation}
\label{sec:evalution}

Evaluating the factuality of LLMs is pivotal for ensuring the reliability and trustworthiness of their generated content \citep{pezeshkpour2023measuring,lee2022factuality}. As LLMs become increasingly integrated into various applications, from information retrieval to content generation, the accuracy of their outputs becomes paramount. In this section, we delve into the evaluation metrics and benchmarks used for assessing the factuality of LLMs, studies that have undertaken such evaluations, and domain-specific evaluation.

\subsection{Factuality Evaluation Metrics}
\label{sec:eval_metrics}
\begin{table*}[t]\small
\centering
\caption{Evaluation Metrics for the Factuality of LLMs.}
\vspace{-2mm}
\label{tab:eval-metrics}
\begin{adjustbox}{max width=\textwidth}
  \setlength{\tabcolsep}{.4mm}
{
\begin{tabular}{l|l|l}
\toprule
\textbf{Evaluation Type} & \textbf{Metric} & \textbf{Description, Purpose/Usage, and Reference/Findings} \\ 
\midrule
\multirow{10}{*}{Rule-based} & Exact Match & Correct matches between generated and input/reference text. \\ \\
& \begin{tabular}[c]{@{}l@{}}Common Metrics \\ (Accuracy, Precision,\\
Recall,  F-Measure,\\ Calibration, Brier)\end{tabular} & \begin{tabular}[c]{@{}l@{}}\textbf{Accuracy}: Measures overall correctness of model prediction.\\ \textbf{Precision}: Ability of model to classify instances positively.\\ \textbf{Recall}: Ability of model to identify all positive instances.\\ \textbf{F-Measure}: Combines both precision and recall into a single score.\\ \textbf{Calibration}: Agreement between predicted probabilities and observed frequencies.\\ \textbf{Brier}: Measures accuracy of probabilistic predictions.\end{tabular} \\ \\
& MC1 and MC2 \citep{TruthfulQA} & Used to identify the correct answer from multi-choice questions. \\ 
& BLEU \citep{papineni2002bleu} & Determines frequency of phrase co-occurrence in two sentences. \\ 
& ROUGE \citep{lin2004rouge}& Calculates similarity between reference and generated text. \\ 
& METEOR\citep{lowe2017towards} & Comprehensive measure based on harmonic mean of unigram precision and recall. \\  \\
& QUIP-Score \citep{weller2023according}& Calculates how much of generated text consists of exact spans found in a text corpus. \\ 
\midrule
\multirow{5}{*}{Neural} & ADEM \citep{lowe2017towards} & \begin{tabular}[c]{@{}l@{}}Uses a Hierarchical RNN to evaluate the quality of responses in language models.\end{tabular}  \\
& BERTScore \citep{zhang2020bertscore}& \begin{tabular}[c]{@{}l@{}}Pre-trained embeddings from BERT used to evaluate sentence similarity.\end{tabular} \\ 
& BLEURT \citep{sellam2020bleurt}& \begin{tabular}[c]{@{}l@{}}Trains BERT on larger dataset and rate similarity of sentences .\end{tabular} \\ 
& BARTScore  \citep{yuan2021bartscore}& \begin{tabular}[c]{@{}l@{}}Evaluates quality using pre-trained sequence-to-sequence models.\end{tabular} \\ 
\midrule
\multirow{2}{*}{Human} & \begin{tabular}[c]{@{}l@{}}AIS \citep{rashkin2023measuring}, \\Auto-AIS\citep{ gao2023rarr}\end{tabular} & \begin{tabular}[c]{@{}l@{}}Evaluates if output is backed by evidence.\end{tabular} \\ \\
& FActScore \citep{Min2023-FActScore}& \begin{tabular}[c]{@{}l@{}}Measures factual accuracy of LLMs breakpointing generated content into atomic facts.\end{tabular} \\ \midrule
\multirow{7}{*}{LLM-based} & GPTScore \citep{fu2023gptscore} & \begin{tabular}[c]{@{}l@{}}Evaluates quality of AI output. It is efficient and avoids the annotation requirement.\end{tabular} \\ \\
& GPT-judge  \citep{TruthfulQA}& \begin{tabular}[c]{@{}l@{}}Evaluates truthfulness of LLM answers.  It is used in evaluating truthfulness\\ and informativeness.\end{tabular} \\ \\
& \begin{tabular}[c]{@{}l@{}}Truthfulness and\\ Informativeness \\\citep{TruthfulQA}\end{tabular} & \begin{tabular}[c]{@{}l@{}}Truthfulness Measures the honesty of LLM information. \\ It is used to evaluate the factuality of information.\\
Informativeness Evaluates the relevance and value of  LLM responses 
\end{tabular}\\ \\
& LLM-Eval \citep{lin2023llm} & \begin{tabular}[c]{@{}l@{}}Evaluates the quality of a conversation. It is adaptable in various scenarios.\end{tabular} \\ \bottomrule
\end{tabular}
}
\end{adjustbox}
\end{table*}

In this subsection, we delve into metrics established for evaluating the factuality of LLMs. As the problem formulation is akin to natural language generation (NLG)
~\citep{celikyilmaz2021evaluation,ji2023survey}, we introduce several automatic evaluation metrics typically used for NLG, as well as specifically examining the metrics for factuality.

We categorize these metrics into the following groups:
(1) Rule-based evaluation metrics;
(2) Neural evaluation metrics;
(3) Human evaluation metrics; and
(4) LLM-based evaluation metrics.

We list those metrics in Table \ref{tab:eval-metrics}.
\subsubsection{Rule-based evaluation metrics}

Most assessments of factuality in large language models use rule-based evaluation metrics, due to their consistency, predictability, and ease of implementation; they allow for reproducible outcomes through a systematic method. However, they can be rigid and may not account for nuances or variations in language use, context interpretation, or colloquial expressions. This means language models rated highly by these metrics may still produce content that feels unnatural or inauthentic to human readers.

\paragraph{Exact Match} An ``exact match" refers to a situation where the generated text precisely matches a specific input or reference text. This means that the LLM produces output that is identical, word-for-word, to the provided input or reference text. Exact matches are often used in NLG when you want to replicate or repeat a piece of text without any variations or alterations. Exact Match is commonly used in Open-domain Question Answering \citep{Fusion-in-Decoder,rfid}.

\paragraph{Common metrics}
Many factuality evaluation measurements use commonly adapted metrics, such as Accuracy, Precision, Recall, AUC, F-Measure, Calibration score,  Brier score, and other common metrics used in probabilistic forecasting and machine learning, specifically in tasks involving probabilistic predictions. The common definition of these metrics involves the use of correctly predicted labels and ground-truth labels. As the input and output of LLMs are human-readable sentences, there is no unified method to convert the sentences into labels. Most evaluation will define their own way. That is, these scores are frequently not used in isolation, but rather in combination. For instance, BERTScore~\citep{BERTscore} uses BERT for determining the Precision and Recall, then uses F-measures to get a final weighted score. In the following, we will describe the most simple form of these scores.

The \emph{Calibration score}, used in  \citep{lin2022teaching, kadavath2022language} measures the agreement between predicted probabilities and observed frequencies. A perfectly calibrated model should, over a large number of instances, see the predicted probability of an outcome match the relative frequency of that outcome.

The \emph{Brier Score}, used in ~\citep{kadavath2022language} is a metric used in probabilistic forecasting to measure the accuracy of probabilistic predictions. It calculates the mean squared difference between the predicted probability assigned to an event and the actual outcome of the event. 
The Brier Score ranges from 0 to 1, where 0 indicates a perfect prediction and 1 indicates the worst possible prediction. In other words, the lower the Brier Score, the better the accuracy of the prediction. It's worth noting that this metric is appropriate for binary and categorical outcomes, but not for ordinal outcomes. 
For binary outcomes, the Brier Score can be calculated as follows:

\begin{equation}
BS = \frac{1}{N} \sum_{i=1}^{N} (forecast_{i} - actual_{i})^2
\end{equation}

where \(forecast_{i}\) is the predicted probability, \(actual_{i}\) is the actual outcome (0 or 1), and \(N\) is the total number of forecasts made.

\paragraph{MC1 (Single-true) and MC2 (Multi-true)} are widely recognized metrics in multi-choice question answering, particularly in TruthfulQA \citep{TruthfulQA}.
MC1: For a given question accompanied by several answer choices, the objective is to identify the sole correct answer. The model's selection is determined by the answer choice to which it allocates the highest log probability of completion, independent of the other choices. The score is calculated as the straightforward accuracy across all questions.
MC2: Presented with a question and multiple reference answers labeled as true or false, the score is derived from the normalized total probability assigned to the collection of true answers.

\paragraph{BLEU}~\citep{papineni2002bleu}, also known as Bilingual Evaluation Understudy metric is commonly employed in the context of factuality evaluation. This metric calculates the co-occurrence frequency of phrases in two sentences, based on a weighted average of matched n-gram phrases. This helps in quantitatively assessing the factual consistency between the generated text and its reference.



\paragraph{ROUGE}
The Recall-Oriented Understudy for Gisting Evaluation (ROUGE) metric \citep{lin2004rouge} serves as a measure of the similarity between the generated text and the reference text, with the similarity grounded on recall scores. Primarily used in the field of text summarization, the ROUGE metric incorporates four distinct types. These include ROUGE-n, which assesses n-gram co-occurrence statistics, and ROUGE-l, which measures the longest common subsequence. ROUGE-w provides evaluation based on the weighted longest common subsequence, while ROUGE-s measures skip-bigram co-occurrence statistics. These diverse metrics collectively provide a comprehensive measure of the factual accuracy of generated text.



\paragraph{METEOR}
The Metric for Evaluation of Translation with Explicit Ordering (METEOR) \citep{banerjee2005meteor} aims to address several shortcomings presented by BLEU. These include deficiencies in recall, the absence of higher order n-grams, an absence of explicit word-matching between the generated and reference text, and the use of geometric averaging of n-grams. METEOR overcomes these by introducing a comprehensive measure, calculated based on the harmonic mean of the unigram precision and recall. This offers potentially enhanced appraisals of factuality in the generated text.

\paragraph{QUIP-Score}~\citep{weller2023according}  is an n-gram overlap measure. It quantifies the degree to which a generated passage consists of exact spans found in a text corpus. The QUIP-Score serves to evaluate the 'grounding' ability of LLMs, specifically assessing whether model-generated answers can be directly located within the underlying text corpus. It is defined by comparing the precision of the character n-gram from the generated output to the pre-training corpus. 


\subsubsection{Neural  Evaluation Metrics}
These metrics operate by comparing the output of these models with a standard or reference text by learning evaluator models. This category primarily comprises three prominent metrics, namely ADEM, BLEURT, and BERTScore. Each metric approaches the evaluation slightly differently, yet all aim to assess the semantic and lexical alignment between machine-generated text and its reference counterpart, thus ensuring the factuality of the generated content.

\paragraph{ADEM}
The automatic Dialogue Evaluation Model (ADEM) metric is a cogent tool utilized to gauge the quality of responses generated by language models in a conversation. Developed by \citet{lowe2017towards}, this metric trains a Hierarchical RNN, in a semi-supervised fashion, to predict ratings for the machine-generated responses. The ADEM's assessment primarily hinges on a dialogue context, designated as $\mathbf{c}$, alongside the model's response, labeled as $\mathbf{\hat{r}}$, and a reference response, specified as $\mathbf{r}$. These elements, when encoded via the hierarchical RNN, inform the ADEM which then predicts a score, reflecting the proximity of the model's response to factual accuracy and relevance.

\paragraph{BERTScore}
the BERTScore metric, as introduced by \citet{zhang2020bertscore}, utilizes pre-trained embeddings from BERT to gauge the similarity between two sentences. This is done by assigning embeddings to a referenced sentence, "x", and a model-generated sentence, "$\hat{x}$", denoted as "$\mathbf{x}$" and "$ \mathbf{\hat{x}}$", respectively. The recall, precision, and F1-scores are then calculated to quantify the similarity between "$\mathbf{x}$" and "$ \mathbf{\hat{x}}$". This score provides a measure of the generated sentence's factuality.

\paragraph{BLEURT}
the Bilingual Evaluation Understudy with Representations from Transformers (BLEURT) metric \citep{sellam2020bleurt} employs a unique pre-training scheme, where BERT is initially trained on a significant corpus of synthetic sentence pairs. This is reinforced by multiple lexical and semantic-level supervision signals, used concurrently. Following this pre-training stage, BERT is further fine-tuned on rating data, and its objective is to estimate human rating scores accurately. The initial stage of pre-training is essential for this metric because it significantly enhances the model's robustness, thereby effectively transforming it into a secure bulwark against quality drifts inherent in generative systems.

\paragraph{BARTScore} ~\citep{yuan2021bartscore} is a metric proposed to evaluate the quality of the generated text, such as in applications like machine translation and summarization. This metric conceives this evaluation as a problem of text generation, modeled using pre-trained sequence-to-sequence models. BARTScore uses BART, an encoder-decoder-based pre-trained model, to translate generated text to and from a reference point, earning higher scores when the text is more accurate and fluent. This metric offers several variants that can be flexibly applied in an unsupervised manner for different perspectives of text evaluation, such as fluency or factuality. Tests have shown that BARTScore can outperform existing top-scoring metrics in the majority of test settings across multiple datasets and perspectives.

\subsubsection{Human Evaluation Metrics}

Human evaluation in factuality assessment is crucial due to its sensitivity to nuanced elements of language and context that may elude automated systems. Human evaluators excel at interpreting abstract concepts and emotional subtleties that can significantly inform the accuracy of evaluation. However, they are subject to limitations such as subjectivity, inconsistency, and potential for error. On the other hand, automated evaluations offer consistent results, and efficient processing of large data sets, and are ideal for tasks needing quantitative measurements. They also provide an objective benchmark for model performance comparison. Overall, an ideal evaluation framework might blend automated evaluation's scalability and consistency with human evaluation's ability to interpret complex linguistic concepts.

\paragraph{Attribution} is a metric to verify that the output of LLMs is sharing only verifiable information about the external world. 
As proposed by \citep{rashkin2023measuring}, Attributable to Identified Sources (AIS) is a human evaluation framework that adopts a binary concept of attribution. A text passage $y$ is deemed attributable to a set $A$ of evidence if, and only if, an arbitrary listener would agree with the statement "According to $A$, $y$" within the context of $y$. The AIS framework awards a full score (1.0) if every element of content in passage $y$ can be linked to the evidence set $A$. Conversely, it gives a score of zero (0.0) if this condition is not met.

Based on AIS,~\citet{gao2023rarr} propose a more fine-grained, sentence-level extension of AIS called Auto-AIS, where annotators assign AIS scores to each sentence, and an average score across all sentences is reported. This procedure effectively measures the percentage of sentences that are fully attributable to the evidence. Context, such as surrounding sentences and the question the text answers, is provided to annotators for more informed judgment. A limit is also set for the number of evidence snippets in the attribution report to maintain conciseness.

During model development, an automated AIS metric is defined to approximate human AIS evaluations, using a natural language inference model, which correlates well with AIS scores. Before computing the scores, they improve accuracy by decontextualizing each sentence in context. 

\paragraph{FActScore} \citep{Min2023-FActScore} is a novel evaluation metric designed to assess the factual precision of long-form text generated by LLMs. The challenge of evaluating the factuality of such text arises from two main issues: (1) the generated content often contains a mix of supported and unsupported information, making binary judgments insufficient, and (2) human evaluation is both time-consuming and expensive.
To address these challenges, FActScore breaks down a generated text into a series of atomic facts—short statements that each convey a single piece of information. Each atomic fact is then evaluated based on its support from a reliable knowledge source. The overall score represents the percentage of atomic facts that are supported by the knowledge source.
The paper conducted extensive human evaluations to compute FActScores for biographies generated by several state-of-the-art commercial LLMs, including InstructGPT, ChatGPT, and the retrieval-augmented PerplexityAI. The results revealed that these LLMs often contain factual inaccuracies, with FActScores ranging from 42\% to 71\%. Notably, the factual precision of these models tends to decrease as the rarity of the entities in the biographies increases.

\subsubsection{LLM-based Metrics}
Using LLMs for evaluation offers efficiency, versatility, reduced reliance on human annotation, and the capability of evaluating multiple dimensions of conversation quality in a single model call, which improves scalability. However, potential issues include a lack of established validation, which can lead to bias or accuracy problems if the LLM used for evaluation is not thoroughly vetted. The decision process to identify suitable LLMs and decoding strategies can be complex and pivotal to obtaining accurate evaluations. The range of evaluation may also be limited, as the focus is often on open-domain conversations, possibly leaving out assessments in specific or narrow domains. While reducing human input can be beneficial, it can also miss out on crucial interaction quality aspects better evaluated by human judges, such as emotional resonance or nuanced understandings.

\paragraph{GPTScore}~\citep{fu2023gptscore} is a new evaluation framework designed to assess the quality of output from generative AI models. To provide these evaluations, GPTScore taps into the emergent capabilities of 19 different pre-trained models, such as zero-shot instruction, and uses them to judge the generated texts. These models vary in scale from 80M to 175B. Testing across four text generation tasks, 22 aspects of evaluation, and 37 related datasets, has demonstrated that GPTScore can effectively evaluate text per instructions in natural language. This attribute allows it to sidestep challenges traditionally encountered in text evaluation, like the need for sample annotations and achieving custom, multi-faceted evaluations.

\paragraph{GPT-judge}~\citep{TruthfulQA} is a finetuned model based on the GPT-3-6.7B, which is trained to evaluate the truthfulness of answers to questions in the TruthfulQA dataset. The training set consists of triples in the form of question-answer-label combinations, where the label can be either true or false. The model's training set includes examples from the benchmark and answers generated by other models assessed by human evaluation. In its final form, the GPT-judge uses examples from all models to evaluate the truthfulness of responses. This training includes all questions from the dataset, with the goal being to evaluate truth, not generalize new questions.

The study conducted by the authors~\citep{TruthfulQA} focuses on the application of GPT-judge in assessing \emph{Truthfulness} and \emph{Informativeness} using the TruthfulQA Dataset. The authors undertook the fine-tuning of two distinct GPT-3 models to evaluate two essential aspects: Truthfulness, which pertains to the accuracy and honesty of information provided by the LLM, and Informativeness, which measures how effectively the LLM conveys relevant and valuable information in its responses.
From these two fundamental concepts, the authors derived a combined metric denoted as \emph{truth * info}. This metric represents the product of scalar scores for both truthfulness and informativeness. It not only quantifies the extent to which questions are answered truthfully but also incorporates the assessment of informativeness for each response. This comprehensive approach prevents the model from generating generic responses like "I have no comment" and ensures that responses are not only truthful but also valuable.
These metrics have found widespread deployment in evaluating the factuality of information generated by LLMs~\citep{chuang2023dola, li2023inferencetime}.

\paragraph{LLM-Eval} \citep{lin2023llm} is a novel evaluation methodology for open-domain dialogues with LLMs. Unlike conventional evaluation methods which rely on human annotations, ground-truth responses, or multiple LLM prompts, LLM-Eval uses a unique prompt-based evaluation process employing a unified schema to assess various elements of a conversation's quality during a single model function. Extensive evaluations of LLM-Eval's performance using multiple benchmark datasets indicate it is effective, efficient, and adaptable compared to traditional evaluation practices. Further, the authors stress the necessity of selecting appropriate LLMs and decoding strategies for precise evaluation outcomes, underscoring LLM-Eval's versatility and dependability in assessing open-domain conversation systems across a variety of circumstances.

\begin{table*}[]\small
\caption{Benchmarks for Evaluating Factuality in LLMs. This table details each benchmark's task type, associated sub-datasets, evaluation metrics, and the performance of select representative LLMs.}
\centering
\label{tab:eval_benchmark}
\vspace{-2mm}
\begin{adjustbox}{max width=\textwidth}
  \setlength{\tabcolsep}{2mm}
  \renewcommand{\arraystretch}{1.25}
{

\begin{tabular}{lllll}
\toprule
Reference & Task Type & Dataset & Metrics & \makecell[l]{Performance of\\ Representative LLMs}  \\
\midrule 

MMLU \citep{MMLU}
&
\begin{tabular}[c]{@{}l@{}}
Multi-Choice QA
\end{tabular}&
\begin{tabular}[c]{@{}l@{}}
Humanities,\\
Social,\\Sciences,\\
STEM...
\end{tabular}&
ACC&
\begin{tabular}[c]{@{}l@{}}
(ACC, 5-shot)\\
GPT-4: 86.4\quad  GPT-3.5: 70  \\LLaMA2-70B: 68.9
\end{tabular} 
\\
\hline 

TruthfulQA \citep{TruthfulQA} &
\begin{tabular}[c]{@{}l@{}}QA \end{tabular}&
\begin{tabular}[c]{@{}l@{}}Health, Law,\\ Conspiracies,\\ Fiction...\end{tabular}&
\begin{tabular}[c]{@{}l@{}}Human Score,\\ GPT-judge, \\ ROUGE, BLEU, \\ MC1,MC2... \end{tabular}&
\begin{tabular}[c]{@{}l@{}}
(zero-shot)\\
GPT-4: $\sim$29 (MC1)\\ GPT-3.5: $\sim$28 (MC1),  79.92(\%true)  \\
LLaMA2-70B: 53.37 (\%true)
\end{tabular}
\\
\hline

C-Eval \citep{C-Eval} &
Multi-Choice QA &
\begin{tabular}[c]{@{}l@{}} 
STEM,\\Social Science,\\Humanities...
\end{tabular} &
\begin{tabular}[c]{@{}l@{}} 
ACC
\end{tabular} &
\begin{tabular}[c]{@{}l@{}} 
(zero-shot, average ACC)\\
GPT-4: 68.7\quad GPT-3.5: 54.4\\
LLaMA2-70B: 50.13
\end{tabular} 
 \\
 \hline

AGIEval \citep{C-Eval} &
Multi-Choice QA &
\begin{tabular}[c]{@{}l@{}} 
Gaokao, \\(geometry, Bio,\\ history...),\\SAT, Law...
\end{tabular} &
\begin{tabular}[c]{@{}l@{}} 
ACC
\end{tabular} &
\begin{tabular}[l]{@{}l@{}} 
(zero-shot, average ACC)\\
GPT-4: 56.4\quad GPT-3.5: 42.9\\
LLaMA2-70B: 40.02
\end{tabular} 
\\
\hline

HaluEval \citep{HaluEval} &
\begin{tabular}[c]{@{}l@{}} 
Hallucination\\Evaluation \end{tabular}&
HaluEval &
ACC&
\begin{tabular}[c]{@{}l@{}} 
(general ACC)\\
GPT-3.5: 86.22
\end{tabular}
\\
\hline

BigBench \citep{BigBench} &
\begin{tabular}[c]{@{}l@{}} 
Multi-tasks
(QA, NLI, \\Fact Checking,\\ Reasoning...)
\end{tabular}&
BigBench
&
\begin{tabular}[c]{@{}l@{}} 
Metric to each\\type of task
\end{tabular}&
\begin{tabular}[c]{@{}l@{}} 
(Big-Bench Hard)\\
GPT-3.5: 49.6 \\
LLaMA-65B: 42.6
\end{tabular}
\\
\hline

ALCE \citep{gao2023enabling} &
\begin{tabular}[c]{@{}l@{}} 
Citation\\Generation
\end{tabular} &
\begin{tabular}[c]{@{}l@{}} 
ASQA, ELI5,\\QAMPARI\\
\end{tabular} &
\begin{tabular}[c]{@{}l@{}} 
MAUVE,\\
Exact Match,\\
ROUGE-L...
\end{tabular} &
\begin{tabular}[c]{@{}l@{}} 
(ASQA, 3-psg, citation prec)\\
GPT-3.5: 73.9\\
LLaMA-33B: 23.0
\end{tabular} 
\\
\hline

QUIP \citep{weller2023according} &
Generative QA &
\begin{tabular}[c]{@{}l@{}}
 TriviaQA,\\ NQ, ELI5,\\HotpotQA 
 \end{tabular} &
\begin{tabular}[c]{@{}l@{}}
 QUIP-Score \\ Exact match
 \end{tabular} &
 \begin{tabular}[c]{@{}l@{}}
 (ELI5, QUIP, null prompt)\\
 GPT-4: 21.0\\ GPT-3.5: 27.7
 \end{tabular} 
 \\
  \hline

 PopQA \citep{PopQA} &
 Multi-Choice QA &
\begin{tabular}[c]{@{}l@{}}
PopQA,\\
EntityQuestions
 \end{tabular} &
 ACC &
\begin{tabular}[c]{@{}l@{}}
(overall ACC)\\
GPT-3.5: $\sim$37.0
\end{tabular} 
 \\
 \hline

UniLC \citep{UniLC} &
Fact Checking &
\begin{tabular}[c]{@{}l@{}}
Climate,\\
Health, MGFN
 \end{tabular} &
 \begin{tabular}[c]{@{}l@{}}
ACC\\F1
 \end{tabular} &
\begin{tabular}[c]{@{}l@{}}
(zero-shot, fact tasks, average F1)\\
GPT-3.5: 51.62
\end{tabular} 
\\
\hline

Pinocchio \citep{Pinocchio} &
\begin{tabular}[c]{@{}l@{}}
Fact Checking\\QA, Reasoning \end{tabular}&
\begin{tabular}[c]{@{}l@{}}
Pinocchio
 \end{tabular} &
 \begin{tabular}[c]{@{}l@{}}
ACC\\F1
 \end{tabular} &
\begin{tabular}[c]{@{}l@{}}
GPT-3.5: (Zero-shot ACC: 46.8, F1:44.4)\\
GPT-3.5: (Few-shot ACC: 47.1, F1:45.7)
\end{tabular} 
\\
\hline

SelfAware \citep{yin-etal-2023-large} &
 Self-evaluation&
\begin{tabular}[c]{@{}l@{}}
SelfAware
 \end{tabular} &
 \begin{tabular}[c]{@{}l@{}}
ACC
 \end{tabular} &
\begin{tabular}[c]{@{}l@{}}
(instruction input, F1)\\
GPT-4: 75.47\quad GPT-3.5: 51.43\\
LLaMA-65B: 46.89
\end{tabular}
\\
\hline

RealTimeQA \citep{kasai2022realtimeqa} &
\begin{tabular}[c]{@{}l@{}} 
Multi-Choice QA,\\Generative QA \end{tabular}&
RealTimeQA &
ACC, F1 &
\begin{tabular}[c]{@{}l@{}}
(original setting, GCS retrieval)\\
GPT-3: 69.3 (ACC for MC)\\
GPT-3: 39.4 (F1 for generation)
\end{tabular}
\\
 \hline

FreshQA \citep{vu2023freshllms} &
Generative QA&
FRESHQA&
ACC (Human)&
\begin{tabular}[c]{@{}l@{}}
(strict ACC, null prompt)\\
GPT-4: 28.6\quad GPT-3.5: 26.0
\end{tabular}
\\

\bottomrule
\end{tabular}}
\end{adjustbox}
\end{table*}


\subsection{{Benchmarks for Factuality Evaluation}}
\label{sec:eval_benchmark}

In this section, we delve into the benchmarks that are prominently employed to assess the factuality of LLMs.
Specific benchmarks tailored for evaluating factuality in LLMs are tabulated in Table \ref{tab:eval_benchmark}. 

\begin{table*}[]\small
\caption{Factuality Evaluation Studies of LLMs. This table shows studies focused on factuality evaluation, and those providing valuable insights into such evaluations. The column `Human Eval' denotes if the evaluation incorporated human evaluation. The `Granularity' column specifies the level of evaluation, with token-level (T) and sentence-level (S) distinctions.}
\centering
\label{tab:evaluation1}
\vspace{-2mm}
\begin{adjustbox}{max width=\textwidth}
  \setlength{\tabcolsep}{0.9mm}
  \renewcommand{\arraystretch}{1}
{

\begin{tabular}{llllclc}
\toprule
Reference & Task & Dataset & Metrics & \makecell[l]{Human\\Eval} & Evaluated LLMs & Granularity \\
\midrule 
\begin{tabular}[c]{@{}l@{}} 
FActScore\\
\citep{Min2023-FActScore} 
\end{tabular}&
\begin{tabular}[c]{@{}l@{}}Biography\\Generation \end{tabular}&
\begin{tabular}[c]{@{}l@{}}183 people\\entities \end{tabular}&
F1&
\checkmark 
&
\begin{tabular}[c]{@{}l@{}} 
GPT-3.5\\
ChatGPT...
\end{tabular}
&
T
\\
\hline

\begin{tabular}[c]{@{}l@{}} 
SelfCheckGPT\\ \citep{SelfCheckGPT}
\end{tabular}&
Bio Generation&
WikiBio&
\begin{tabular}[c]{@{}l@{}}
AUC-PR,\\
Human Score
\end{tabular}&
\checkmark
&
\begin{tabular}[c]{@{}l@{}}
GPT-3,\\
LLaMA,\\
OPT,\\
GPT-J...
\end{tabular}
&
S
\\
\hline

\citet{wang2023evaluating} &
Open QA &
\begin{tabular}[c]{@{}l@{}}NQ, TQ  \\ \end{tabular}&
\begin{tabular}[c]{@{}l@{}}ACC, \\ EM\end{tabular} &
\checkmark &
\begin{tabular}[c]{@{}l@{}}GPT-3.5\\ ChatGPT\\GPT-4\\Bing Chat\end{tabular}&
S 
\\
\hline

\citet{pezeshkpour2023measuring} &
\begin{tabular}[c]{@{}l@{}}Knowledge \\Probing \end{tabular}&
\begin{tabular}[c]{@{}l@{}}T-REx, \\LAMA\end{tabular}&
ACC &
&
GPT3.5 &
T 
\\
\hline
\citet{de-cao-etal-2021-editing} &
\begin{tabular}[c]{@{}l@{}} 
QA, \\
Fact Checking \end{tabular}&
\begin{tabular}[c]{@{}l@{}} 
KILT, \\FEVER, \\zsRE
\end{tabular} &
ACC &
&
\begin{tabular}[c]{@{}l@{}} GPT-3\\
FLAN-T5\end{tabular} &
\begin{tabular}[c]{@{}l@{}} S/
T \end{tabular} 
\\
\hline

 \citet{varshney2023stitch}&
 \begin{tabular}[c]{@{}l@{}} Article\\Generation\end{tabular}&
 \begin{tabular}[c]{@{}l@{}} Unnamed\\Dataset\end{tabular}&
 \begin{tabular}[c]{@{}l@{}} 
 ACC, \\AUC
 \end{tabular}&
 &
 \begin{tabular}[c]{@{}l@{}} 
 GPT3.5\\
 Vicuna
 \end{tabular} &
 S 
 \\
 \hline

\begin{tabular}[c]{@{}l@{}} 
FactTool\\
\citep{chern2023factool}
\end{tabular}&
KB-based QA&
\begin{tabular}[c]{@{}l@{}} 
RoSE
\end{tabular}&
\begin{tabular}[c]{@{}l@{}}
ACC\\F1...
\end{tabular}&
&
\begin{tabular}[c]{@{}l@{}} 
GPT-4\\
ChatGPT \\
FLAN-T5
\end{tabular}&
S
\\
\hline

\citet{kadavath2022language} &
Self-evaluation &
\begin{tabular}[c]{@{}l@{}} 
BIG Bench,\\
MMLU, 
LogiQA,\\
TruthfulQA,\\
QuALITY\\
TriviaQA\\
Lambada
\end{tabular}&
\begin{tabular}[c]{@{}l@{}} 
ACC, \\
Brier Score, \\
RMS\\Calibration\\Error...
\end{tabular}&
&
Claude&
T
\\ \midrule
\citet{chen2023beyond} &
QA &
\begin{tabular}[c]{@{}l@{}} 
NQ,\\
WoW
\end{tabular}&
\begin{tabular}[c]{@{}l@{}} 
Factuality, \\Relevance,\\ Coherence,\\ Informativeness,\\ Helpfulness,\\ and Validity
\end{tabular}&
&
LLaMA, FLAN-T5, ...&
T
\\
\midrule
RAGAS \citep{es2023ragas} &
QA &
\begin{tabular}[c]{@{}l@{}} 
WikiEval
\end{tabular}&
\begin{tabular}[c]{@{}l@{}} 
Faithfulness, \\Answer Relevance,\\ Context Relevance
\end{tabular}&
&
ChatGPT&
T
\\
\bottomrule
\end{tabular}}
\end{adjustbox}
\end{table*}

MMLU \citep{MMLU} and TruthfulQA \citep{TruthfulQA} stand as two pivotal benchmarks in the realm of evaluating the factuality of LLMs \citep{GPT4,llama2,Open-LLM-Leaderboard} . 
The MMLU benchmark is proposed to measure a text model's multitask accuracy across a diverse set of 57 tasks. These tasks span a wide range of subjects, from elementary mathematics to US history, computer science, law, and more. The benchmark is designed to test both the world knowledge and problem-solving ability of models. The findings from the paper suggest that while most recent models perform at near random-chance accuracy, the largest GPT-3 model showcased a significant improvement. However, even the best models still have a long way to go before achieving expert-level accuracy across all tasks \citep{GPT4}.
TruthfulQA is a benchmark designed to assess the truthfulness of a language model's generated answers. The benchmark consists of 817 questions spanning 38 categories, including health, law, finance, and politics. These questions were crafted in such a way that some humans might answer them falsely due to misconceptions or false beliefs. The goal for models is to avoid generating these false answers that they might have learned from imitating human texts. The TruthfulQA benchmark serves as a tool to highlight the potential pitfalls of relying solely on LLMs for accurate information and emphasizes the need for continued research in this area.

HaluEval \citep{HaluEval} is a benchmark designed to understand and evaluate the propensity of LLMs like ChatGPT to generate hallucinations. A hallucination, in this context, refers to content that either conflicts with the source or cannot be verified based on factual knowledge.
The HaluEval benchmark offers a vast collection of generated and human-annotated hallucinated samples, aiming to evaluate the performance of LLMs in recognizing such hallucinations. The benchmark utilizes a two-step framework, termed "sampling-then-filtering", based on ChatGPT to generate these samples. Additionally, human labelers were employed to annotate hallucinations in ChatGPT responses. The HaluEval benchmark is a comprehensive tool that not only evaluates the hallucination tendencies of LLMs but also provides insights into the types of content and the extent to which these models are prone to hallucinate. 

BigBench \citep{BigBench} focuses on the capabilities and limitations of LLMs. It comprises 204 tasks from diverse domains such as linguistics, childhood development, math, common-sense reasoning, biology, physics, social bias, software development, and more. The benchmark is designed to evaluate tasks believed to be beyond the capabilities of current language models. The study evaluates the performance of various models, including OpenAI's GPT models, on BIG-bench and compares them to human expert raters. Key findings suggest that model performance and calibration improve with scale but are still suboptimal when compared to human performance. Tasks that involve a significant knowledge or memorization component show predictable improvement, while tasks that exhibit "breakthrough" behavior at a certain scale often involve multiple steps or components. 

\citet{C-Eval} propose C-Eval, the first comprehensive Chinese evaluation suite. It can be used to evaluate the advanced knowledge and reasoning abilities of foundational models within a Chinese context. The evaluation suite comprises multiple-choice questions spanning 52 diverse disciplines, with four levels of difficulty: middle school, high school, college, and professional. Additionally, C-Eval Hard is introduced for very challenging subjects within the C-Eval suite, which demands advanced reasoning abilities to solve. Their evaluations of state-of-the-art LLMs, including both English and Chinese-oriented models, have shows that there is significant room for improvement as only GPT-4 managed to achieve an average accuracy of over 60\%.
The authors focus on assessing LLMs' advanced abilities within a Chinese context. The researchers assert that LLMs intended for a Chinese environment should be evaluated based on their knowledge of Chinese users' primary interests, such as Chinese culture, history, and laws.
With C-Eval, the authors aim to guide developers in understanding the abilities of their models from multiple dimensions to facilitate the development and growth of foundational models for Chinese users. Simultaneously, C-Eval has not just introduced a whole suite but subsets that can serve as individual benchmarks, thereby assessing certain model abilities and analyzing key strengths and limitations of foundational models. Experiment results have shown that although GPT-4, ChatGPT, and Claude were not exclusively tailored for Chinese data, they emerged as the top performers on C-Eval.

SelfAware \citep{yin-etal-2023-large} aims to investigate whether models recognize what they don't know. This dataset encompasses two types of questions: unanswerable and answerable. The dataset comprises 2,858 unanswerable questions gathered from various websites and 2,337 answerable questions extracted from sources such as SQuAD, HotpotQA, and TriviaQA. Each unanswerable question is confirmed as such by three human evaluators. In the conducted experiments, GPT-4 achieves the highest F1 score of 75.5, compared to a human score of 85.0. Larger models tend to perform better and in-context learning can enhance performance.

The Pinocchio benchmark \citep{Pinocchio} serves as an extensive evaluation platform, emphasizing factuality and reasoning for LLMs. This benchmark encompasses 20,000 varied factual queries from diverse sources, timeframes, fields, regions, and languages. It tests an LLM's capability to discern combined facts, process both organized and scattered evidence, recognize the temporal evolution of facts, pinpoint minute factual disparities, and withstand adversarial inputs. Each reasoning challenge within the benchmark is calibrated for difficulty to allow for detailed analysis.

\citet{kasai2022realtimeqa} introduce a dynamic QA platform named REALTIMEQA. This platform is unique in that it announces questions and evaluates systems on a regular basis, specifically weekly. The questions posed by REALTIMEQA pertain to current events or novel information, challenging the static nature of traditional open domain QA datasets. The platform aims to address instantaneous information needs, pushing QA systems to provide answers about recent events or developments. Their preliminary findings indicate that while GPT-3 can often update its generation results based on newly-retrieved documents, it sometimes returns outdated answers when the retrieved documents lack sufficient information.

FreshQA \citep{vu2023freshllms} is a dynamic benchmark designed to evaluate up-to-date world knowledge of LLMs. Its questions range from those that are never-changing to those that are fast-changing, as well as questions based on false premises. The aim is to challenge the static nature of LLMs and test their adaptability to ever-changing knowledge through human evaluations. They develop a reliable evaluation protocol that uses a two-mode system: RELAXED and STRICT for a comprehensive understanding of model performance, ensuring that answers are confident, definitive, and accurate. They also provide a strong baseline named FRESHPROMPT, which seeks to enhance LLM performance by integrating real-time data from search engines. Initial experiments reveal that, outdated training data weakens the performance of LLMs and the FRESHPROMPT method can significantly enhance it. The research underscores the need for LLMs to be refreshed with current information to ensure their relevance and accuracy in a constantly evolving world.

Several benchmarks, such as BigBench \citep{BigBench} and C-Eval \citep{C-Eval}, encompass subsets that extend beyond the realm of factual knowledge or factuality. In this work, we specifically emphasize and focus on those subsets related to factuality.

There are benchmarks primarily designed for Pre-trained Language Models (PLMs) that can also be adapted for LLMs. Some studies use them for evaluating LLM's factuality, but they are not so widely-used, so we have chosen to exclude them from the table for clarity. These benchmarks predominantly encompass knowledge-intensive tasks, as highlighted by \citep{kilt}. Those benchmarks include NaturalQuestions (NQ) \citep{NQ}, TriviaQA (TQ) \citep{TQ}, OTT-QA~\citep{OTT-QA}, AmbigQA~\cite{ambigQA} and WebQuestion (WQ) \citep{WQ} of Open-domain Question Answering (QA) task, the HotpotQA~\citep{hotpotqa}, 2WikiMultihopQA~\citep{2WikiMultihopQA}, IIRC~\citep{IIRC}, MuSiQue~\citep{musique} of multi-step
QA, the ELI5 \citep{ELI5} of the Long-form QA task, the FEVER \citep{fever}, FM2 \citep{FM2}, HOVER \citep{HOVER}, FEVEROUS \citep{FEVEROUS} of the Fact Checking task, the T-REx \citep{TREx}, zsRE \citep{ZsRE} and LAMA \citep{LAMA} to examine factual knowledge contained in pretrained language models, the WikiBio \citep{wikiBio} of biography generation task, the RoSE \citep{Rose}, WikiAsp \citep{wikiasp} of summarization task, the KILT \citep{kilt} of comprehensive knowledge intensive tasks, the MassiveText \citep{Gopher}, Curation Corpus \citep{curation}, Wikitext103 \citep{Wikitext103}, Lambada \citep{lambada}, C4 \citep{C4}, Pile \citep{pile} of langauge modeling task, the WoW \citep{WoW}, DSTC7 track2~\citep{dstc7}, DSTC11 track5~\citep{dstc11} of dialogue task, the RealToxicityPrompts \cite{RealToxicityPrompts} of toxicity reduction, the CommaQA \citep{CommaQA}, StrategyQA \citep{StrategyQA}, TempQuestions \citep{TempQuestions}, IN-FOTABS~\citep{INFOTABS} of diverse reasoning tasks.


Some studies \citep{Min2023-FActScore, SelfCheckGPT} also provide a small dataset, but they mainly concentrate on the evaluation metrics or methods for factuality, we choose to discuss them in the next subsection.

\subsection{Factuality Evaluation Studies}
\label{sec:factuality-eval}

\begin{table*}[]\small
\caption{Factuality Evaluation of LLMs. This table lists the benchmarks that used in the enhancement work. The column `Human Eval' denotes if the evaluation incorporated human evaluation. The `Granularity' column specifies the level of evaluation, with token-level (T) and sentence-level (S) distinctions.}
\centering
\label{tab:eval-table2}
\vspace{-2mm}
\begin{adjustbox}{max width=\textwidth}
  \setlength{\tabcolsep}{1.2mm}
    \renewcommand{\arraystretch}{0.88}
{

\begin{tabular}{llllclc}
\toprule
Reference & Task & Dataset & Metrics & \makecell[l]{Human\\Eval} & Evaluated LLMs & Granularity \\
\midrule 
\begin{tabular}[c]{@{}l@{}}
Retro\\
\citep{borgeaud2022improving} 
\end{tabular} &
\begin{tabular}[c]{@{}l@{}} QA, \\Language\\Modeling\end{tabular} &
\begin{tabular}[c]{@{}l@{}}
MassiveText, \\
Curation Corpus, \\
Wikitext103, \\
Lambada, \\
C4,Pile, NQ
\end{tabular} &
\begin{tabular}[c]{@{}l@{}} 
PPL, \\ACC, \\Exact Match
\end{tabular} &
   \checkmark&
\begin{tabular}[c]{@{}l@{}} 
Retro
\end{tabular} &
T 
\\
\hline
\begin{tabular}[c]{@{}l@{}} 
GenRead\\
 \citep{yu2023generate} 
\end{tabular} &
\begin{tabular}[c]{@{}l@{}} QA, \\ Dialogue, \\ Fact Checking \end{tabular} &
\begin{tabular}[c]{@{}l@{}} NQ, TQ, WebQ, \\FEVER, \\FM2, WoW  \end{tabular} &
\begin{tabular}[c]{@{}l@{}} 
EM, ACC, \\F1, Rouge-L
\end{tabular} &
   &
   \begin{tabular}[c]{@{}l@{}} 
GPT3.5, Codex\\GPT-3, Gopher\\FLAN, GLaM\\PaLM\end{tabular} &
S 
\\
\hline
\begin{tabular}[c]{@{}l@{}}
GopherCite\\ \citep{menick2022teaching}  
\end{tabular}
&
\begin{tabular}[c]{@{}l@{}} Self-supported QA\end{tabular} &
\begin{tabular}[c]{@{}l@{}} 
NQ, ELI5, \\
 TruthfulQA\\
(Health\\Law, Fiction\\ Conspiracies)

\end{tabular} &
\begin{tabular}[c]{@{}l@{}} 
Human Score
\end{tabular} &
   \checkmark&
   \begin{tabular}[c]{@{}l@{}} 
GopherCite
\end{tabular} &
S 
\\
\hline
\citet{trivedi-etal-2023-interleaving} &
\begin{tabular}[c]{@{}l@{}} 
 QA
\end{tabular} &
\begin{tabular}[c]{@{}l@{}} 
HotpotQA, IIRC\\2WikiMultihopQA, \\MuSiQue(music)
\end{tabular} &
\begin{tabular}[c]{@{}l@{}} 
Retrieval recall, \\
Answer F1
\end{tabular} &
 &
 \begin{tabular}[c]{@{}l@{}} 
 GPT-3\\
FLAN-T5
 \end{tabular} &
  \begin{tabular}[c]{@{}l@{}} 
 S/
T
 \end{tabular} 
 \\
\hline
 \citet{peng2023check} &
\begin{tabular}[c]{@{}l@{}} 
QA, \\
Dialogue
\end{tabular} &
\begin{tabular}[c]{@{}l@{}} 
DSTC7 track2, \\
DSTC11 track5, \\
OTT-QA
\end{tabular} &
\begin{tabular}[c]{@{}l@{}} 
ROUGE, chrF, \\
BERTScore, \\
Usefulness, \\
Humanness...
\end{tabular} &
\checkmark &
ChatGPT &
  \begin{tabular}[c]{@{}l@{}} 
 S/
T
 \end{tabular} 
 \\
\hline
\begin{tabular}[c]{@{}l@{}}
CRITIC\\
\citep{gou2023critic}
\end{tabular} &
\begin{tabular}[c]{@{}l@{}}
QA,\\Toxicity Reduction
 \end{tabular} &
\begin{tabular}[c]{@{}l@{}}
 AmbigNQ, \\TriviaQA, \\HotpotQA, 
\\RealToxicityPrompts
 \end{tabular}  &
 \begin{tabular}[c]{@{}l@{}}
Exact Match, \\
maximum toxicity, \\
perplexity, \\
n-gram diversity, \\
AUROC...,
\end{tabular} &
&
\begin{tabular}[c]{@{}l@{}}
GPT-3.5\\
ChatGPT
\end{tabular} &
T 
\\
\hline
\citet{khot2023decomposed} &
\begin{tabular}[c]{@{}l@{}}
QA, \\
long-context\\ QA
\end{tabular} &
\begin{tabular}[c]{@{}l@{}}
CommaQA-E, \\
2WikiMultihopQA, \\ MuSiQue, \\HotpotQA
\end{tabular} &
\begin{tabular}[c]{@{}l@{}}
Exact Match, \\
Answer F1
\end{tabular} &
&
\begin{tabular}[c]{@{}l@{}}
GPT-3\\
FLAN-T5
\end{tabular} &
T 
\\
\hline
\begin{tabular}[c]{@{}l@{}} 
ReAct\\
\citep{yao2023react} 
\end{tabular} &
\begin{tabular}[c]{@{}l@{}} 
QA,\\ Fact Verification 
\end{tabular} &
\begin{tabular}[c]{@{}l@{}}
HotpotQA, \\FEVER
\end{tabular} &
\begin{tabular}[c]{@{}l@{}}
Exact Match, \\
ACC
\end{tabular} &
 &
\begin{tabular}[c]{@{}l@{}}
PaLM\\
GPT-3
\end{tabular} &
\begin{tabular}[c]{@{}l@{}}
S/T
\end{tabular}
\\
\hline
\citet{jiang2023active} &
\begin{tabular}[c]{@{}l@{}}
QA, \\Commonsense\\Reasoning, \\
long-form QA...
\end{tabular} &
\begin{tabular}[c]{@{}l@{}} 
2WikiMultihopQA, \\StrategyQA, 
\\ASQA, WikiAsp
\end{tabular} &
\begin{tabular}[c]{@{}l@{}} 
Exact Match, \\
Disambig-F1, \\ROUGE, \\
entity F1...
\end{tabular} &
&
GPT-3.5 &
T 
\\
\hline
\citet{lee2022factuality} &
\begin{tabular}[c]{@{}l@{}}
Open-ended\\Generation 
\end{tabular}&
FEVER &
\begin{tabular}[c]{@{}l@{}} 
Entity score, \\
Entailment\\Ratio, ppl...
\end{tabular}&
 &
 Megatron-LM &
 T
\\
\hline
\begin{tabular}[c]{@{}l@{}}
SAIL\\
\citep{luo2023sail} 
\end{tabular}&
\begin{tabular}[c]{@{}l@{}}
QA,\\
Fact Checking 
\end{tabular}&
UniLC&
\begin{tabular}[c]{@{}l@{}} 
ACC\\ F1
\end{tabular}&
&
\begin{tabular}[c]{@{}l@{}}
LLaMA
Vicuna\\
SAIL
\end{tabular}&
T
\\
\hline
\citet{he2022rethinking} &
\begin{tabular}[c]{@{}l@{}} 
Commonsense\\ Reasoning, \\Temporal\\ Reasoning, \\
Tabular\\Reasoning
\end{tabular} &
\begin{tabular}[c]{@{}l@{}}
StrategyQA, \\ TempQuestions, \\ IN-FOTABS
\end{tabular}&
ACC&
&
GPT-3 &
T
\\
\hline
\citet{pan-etal-2023-fact} &
Fact Checking&
\begin{tabular}[c]{@{}l@{}}
HOVER \\ FEVEROUS-S
\end{tabular}&
Macro-F1 &
&
\begin{tabular}[c]{@{}l@{}}
Codex\\
FLAN-T5
\end{tabular}&
S
\\
\hline
\citet{multiagent_debate}&
\begin{tabular}[c]{@{}l@{}} 
Biography,\\
MMLU
\end{tabular}&
\begin{tabular}[c]{@{}l@{}} 
Unnamed\\Biography\\Dataset, \\
MMLU
\end{tabular}&
\begin{tabular}[c]{@{}l@{}} 
ChatGPT\\Evaluator, \\ACC
\end{tabular}&
&
\begin{tabular}[c]{@{}l@{}} 
Bard\\ChatGPT
\end{tabular}&
S
\\
\hline 
\citet{asai2023selfrag} &
\begin{tabular}[c]{@{}l@{}} 
QA,\\
CItation 
Generation
\end{tabular}&
\begin{tabular}[c]{@{}l@{}} 
PopQA,\\Pubhealth,\\ASQA,\\ARC,\\Unamed\\Biography
\end{tabular}&
\begin{tabular}[c]{@{}l@{}} 
ACC,\\Factscore,\\Exact Match,\\
ROUGE,\\MAUVE...
\end{tabular}&
&
\begin{tabular}[c]{@{}l@{}} 
LLaMA2\\ChatGPT\\Perplexity.ai\\Alpaca\\SAIL
\end{tabular}&
S/T
\\
    
\bottomrule
\end{tabular}}
\end{adjustbox}
\end{table*}

In this section, we delve into studies that evaluate factuality in LLMs without introducing a specific benchmark, focusing primarily on those whose main contribution lies in the evaluation methodology. We spotlight works that have pioneered evaluation techniques, metrics, or have offered distinctive insights into the factuality evaluation of LLMs.

\citet{SelfCheckGPT} use an evaluation process that encompasses several key steps. Initially, synthetic Wikipedia articles are generated using GPT-3, focusing on individuals from the Wikibio dataset. Subsequently, manual annotation is performed at the sentence level, classifying sentences as ``Major Inaccurate" , ``Minor Inaccurate", or ``Accurate", with ``Major Inaccurate" denoting sentences unrelated to the topic. Passage-level scores are derived by averaging sentence-level labels, and identifying cases of total inaccuracies through score distribution analysis. Inter-annotator agreement is assessed using Cohen's $\kappa$ ~\cite{cohen1960coefficient}. Evaluation metrics primarily employ precision-recall curves (PR-Curves), distinguishing between ``Non-Factual Sentences," ``Non-Factual* Sentences" (a specific subset), and ``Factual Sentences." These PR-Curves elucidate the trade-off between precision and recall for different detection methods.

BSChecker \citep{BSChecker} is a factual errors detection framework that detects inaccuracies on triplet-level granularity. Its pipeline comprises a claim extractor and a checker. The claim extractor extracts triplets from model’s response as claims formed with (subject, predicate, object). Meanwhile, the checker compare these claims triplets with references, categorizing each triplet to one of three hallucination labels: Entailment, Contradiction or Neutral, drawing inspiration from the Natural Language Inference task. BSChecker offers a benchmark that distinguishes three task settings: Zero Context, Noisy Context and Accurate Context covering tasks of closed-book QA, retrieval-augmented generation, summarization, closed-QA and information extraction.

\citet{pezeshkpour2023measuring} propose a new metric for measuring whether a certain type of knowledge is present in LLM. This metric is based on information theory and measures knowledge by analyzing the probability distribution of predictions made by LLM before and after injecting target knowledge. The accuracy of GPT-3.5 in the Knowledge Probing task is tested on the T-REx \citep{TREx} and LAMA \citep{LAMA} datasets. 


\citet{varshney2023stitch} point out a common real-world occurrence where users often ask questions based on false premises. These questions are challenging for state-of-the-art models.
This necessitated the creation of a new evaluation dataset. To this end, the authors have conducted a case study and compiled a set of 50 such adversarial questions, all of which the GPT-3.5 model answered incorrectly. The aim is to create a challenging experimental setup to assess the performance of models faced with such questions.
In order to enhance the evaluation, corresponding true premise questions were created for each of the false premise questions. This allowe for a holistic evaluation of the model's performances, taking into consideration both correct and incorrect premises.
The authors make sure to evaluate the complete answers given by the model for correct and incorrect questions - in this context, it's not enough for an answer to be partially correct, the entire answer needs to be accurate to be marked as correct. For example: For the false premise question ``Why does Helium have an atomic number of 1?", the corresponding true premise question is ``Why does Hydrogen have an atomic number of 1?".

FACTOOL \citep{chern2023factool} is a tool designed to function as a factuality detector, with the primary purpose of auditing generative chatbots and assessing the reliability of their outputs. This tool is employed to evaluate several contemporary chatbots, including GPT-4, ChatGPT, Claude, Bard \citep{bard}, and Vicuna \citep{vicuna2023}. Notably, FACTOOL itself leverages the capabilities of GPT-4.
For the evaluation process, the researchers have curated a diverse set of prompts: 30 from knowledge-based question answering (KB-QA), 10 each from code, math, and scientific domains. The KB-QA prompts were sourced from a prior study, code prompts were taken from HumanEval, math prompts from another distinct study, while the scientific prompts were crafted by the authors themselves.
The evaluation metrics included both claim-level and response-level accuracies for each chatbot. To offer a more comprehensive and equitable evaluation, a weighted claim-level accuracy is used. The weighting is determined based on the proportion of prompts from each category.
The findings are illuminating. GPT-4 emerge as the top performer in terms of both weighted claim-level factual accuracy and response-level accuracy among all the chatbots assessed. Another intriguing observation is that chatbots that underwent supervised fine-tuning, such as Vicuna-13B, exhibited commendable performance in standard scenarios like KB-QA. However, their performance dip in more intricate scenarios, including those involving math, code, and scientific queries.

\citet{wang2023evaluating} ask several LLMs, including ChatGPT, GPT-4 \citep{GPT4}, BingChat \citep{BingChat} to answer open questions from NaturalQuestions \citep{NQ} and TriviaQA \citep{TQ}. They manually estimate the accuracy of those LLMs on open question answering, and find that though LLMs can achieve nice performance but still far away from perfect. Besides, they evaluate whether the GPT-3.5 can assess the correctness of LLM-generated responses, and find negative results, even if the golden answer is also presented.
Similarity, \citet{fu2023large} ask LLMs, such as GPT-2 and GPT-4, to directly score the factuality of a summary, and find  no significant correlation  between LLM's factuality indicators and human evaluations.

\citet{kadavath2022language} investigate whether language models can evaluate the accuracy of their own assertions and predict which questions they can answer correctly. It is found that larger models are well-calibrated on diverse multiple-choice and true/false questions if given in the appropriate format.
The approach to self-evaluation on open-ended tasks is to ask the models to initially suggest answers, and then evaluate the probability (P[True]) that their answers are correct. This resulted in compelling performance, calibration, and scaling on a diverse range of tasks. Furthermore, self-evaluation performance improved when the models are allowed to consider many of their own suggestions before predicting the validity of a specific one.

\citet{yu2023generate} explore whether the internal knowledge of LLMs can replace the retrieved documents on knowledge intensive tasks. They ask LLMs, such as InstructGPT \citep{gpt35}, to directly generate contexts given a question rather than retrieving from database. They find the generated documents contain the golden answers more often than the top retrieved documents. Then they feed the generated docs and retrieved docs to the Fusion-in-Decoder model \citep{Fusion-in-Decoder} for knowledge-intensive tasks such as Open-domain QA \citep{NQ} and find the generated docs are more effective than the retrieved docs, suggesting that the LLMs contain enough knowledge for knowledge-intensive tasks.

\citet{menick2022teaching} propose a task named \textit{Self-supported QA} to evaluate LLMs' ability in also producing citations when generating answers. Authors ask humans to evaluate whether the responses of their proposed model GopherCite are plausible and whether they are supported by the accompanying quote evidence on datasets such as NQ, ELI5, TruthfulQA.

CONNER \citep{chen2023beyond}, a framework that evaluates LLMs as generators of knowledge. It focuses on six areas: Factuality, Relevance, Coherence, Informativeness, Helpfulness, and Validity. It evaluates whether the generated information can be backed by external proof (Factuality), is relevant to the user's query (Relevance), and is logically consistent (Coherence). It also checks if the knowledge provided is novel or surprising (Informativeness). The Extrinsic evaluation measures whether the knowledge enhances downstream tasks (Helpfulness) and its results are factually accurate (Validity). 

In the realm of factuality evaluation, the Model Editing task holds a unique position, focusing on refining the internal knowledge of models. This task comes with its own set of specialized evaluation metrics. Predominantly, the Zero-Shot Relation Extraction (zsRE) \cite{ZsRE} and CounterFact \cite{meng2022locating} serve as the primary benchmarks for assessing Model Editing techniques. When evaluating these methods, several key criteria emerge:
Reliability: Post-editing, the model should consistently generate the intended output.
Generalization: The model should be adept at producing the target output even when presented with paraphrased inputs.
Locality: Edits should be localized, ensuring that facts not related to the specific edit remain intact.
However, given the intricate web of interconnected facts, recent studies \citet{zhong2023mquake,yao2023editing,cohen2023evaluating} have advocated for a more holistic evaluation approach. They introduce broader criteria for fact updates, encompassing aspects like portability, logical generalization, among others.

RAGAS \citep{es2023ragas} proposes a framework for the reference-free evaluation of Retrieval-Augmented Generation (RAG) systems in Large Language Models (LLMs). This framework primarily assesses the RAG system's ability to identify relevant and key context paragraphs, the LLM's fidelity in utilizing these paragraphs, and the overall quality of the generated content. The framework focuses on three aspects of quality:
Faithfulness: The generated answers should be based on the given context. To evaluate faithfulness, the generated content is broken down into sentences, which are then transformed into one or more short assertions. Each assertion's faithfulness is assessed against the retrieved content, with the faithfulness score being the ratio of faithful assertions to the total number of assertions.
Answer Relevance: The generated answers should appropriately address the posed questions. This is evaluated by having the LLM generate potential questions from each piece of generated content. An embedding model then calculates the similarity between each potential question and the original question. The answer relevance score is the average of these similarities.
Context Relevance: The retrieved context should be highly relevant, containing as little irrelevant information as possible. This is assessed by using the LLM to extract sentences from the retrieved content that are relevant to the question. The context relevance score is then calculated based on the proportion of these relevant sentences to the total number of sentences retrieved.
The authors created a WikiEval dataset, consisting of question-context-answer triplets with human judgments. This allows for the measurement of how closely different evaluation frameworks align with human assessments of faithfulness, answer relevance, and context relevance. The paper compares two baselines: GPTScore, which requires ChatGPT to rate the three quality dimensions on a scale of 0 to 10, and GPT Ranking, which does not require ChatGPT to choose a preferred answer/context but includes definitions of the considered quality metrics in the prompt. When evaluating answer relevance, a specific prompt is used.
Overall, the RAGAS framework proposed in the paper aligns more closely with human judgment compared to the two baselines.


Some work's main contributions lie at the methods to improve the factuality of LLMs, and their evaluation part may also be informative and important when people reach the related study. So we choose to list evaluation parts in the Table \ref{tab:eval-table2} but not to discuss their evaluation in detail.

\subsection{Evaluating Domain-specific Factuality}


\label{sec:domain-eval}
\begin{table*}[]\small
\centering
\caption{Benchmarks for domain-specific factuality evaluation. The table presents the domain, tasks, datasets, and the LLMs evaluated in the respective study.}
\vspace{-2mm}
\label{tab:domain-eval}
\begin{adjustbox}{max width=\textwidth}
  \setlength{\tabcolsep}{2mm}
{
\begin{tabular}{lllll}
\toprule
Reference &
  Domain &
  Task &
  Metrics &
  Evaluated LLMs \\
  \midrule FLARE \citep{xie2023pixiu} &
  Finance  &
  \begin{tabular}[c]{@{}l@{}}Sentiment analysis, \\ News headline classification \\ Named entity recognition \\ Question answering \\ Stock movement prediction\end{tabular} &
  \begin{tabular}[c]{@{}l@{}}F1, ACC,\\Avg F1, \\ Entity F1, \\ EM, MCC\end{tabular} &
  \begin{tabular}[c]{@{}l@{}}GPT-4 , \\ BloombergGPT, \\  FinMA-(7B, 30B, 7B-full), \\ Vicuna-7B \end{tabular}\\
   \midrule EcomInstruct \citep{li2023ecomgpt} & Finance  
   & 134 E-com tasks
   & \begin{tabular}[c]{@{}l@{}} Micro-F1, \\Macro-F1, \\
  ROUGE\end{tabular}
   & \begin{tabular}[c]{@{}l@{}}BLOOM, BLOOMZ,\\ ChatGPT, EcomGPT\end{tabular}
   \\
   \midrule CMB \citep{wang2023cmb} &
  Medicine &
  Multi-Choice QA &
  ACC &
  \begin{tabular}[c]{@{}l@{}}GPT-4, ChatGLM2-6B,\\ ChatGPT, DoctorGLM,\\ Baichuan-13B-chat,\\ HuatuoGPT, MedicalGPT,\\ ChatMed-Consult,\\ ChatGLM-Med ,\\ Bentsao, BianQue-2\end{tabular} \\
   \midrule Huatuo-26M \citep{li2023huatuo} &
Medicine & Generative-QA
    &
  \begin{tabular}[c]{@{}l@{}}BLEU, \\ROUGE, \\ GLEU\end{tabular} &
  T5, GPT2 \\
   \midrule NCBI \citep{jin2023genegpt} &
  Medicine &
  \begin{tabular}[c]{@{}l@{}}Nomenclature, \\ Genomic location, \\
  Functional analysis, \\
  Sequence alignment \end{tabular} &
  ACC & 
   \begin{tabular}[c]{@{}l@{}}GPT-2, BioGPT, \\
   BioMedLM, 
   GPT-3, \\
   ChatGPT, New Bing
   \end{tabular} \\
   \midrule LegalBench \citep{guha2023legalbench} &
  Law &
  \begin{tabular}[c]{@{}l@{}}Issue-spotting,  \\
  Rule-recall, \\
  Rule-application, \\ Rule-conclusion,
  \\
  Interpretation,
  \\
  Rhetorical-understanding
  \end{tabular}  &
  \begin{tabular}[c]{@{}l@{}} 
          ACC, EM
   \end{tabular}
  & 
   \begin{tabular}[c]{@{}l@{}}
   GPT-4, GPT-3.5, \\
   Claude-1, Incite, OPT\\
   Falcon, LLaMA-2, FLAN-T5...
   \end{tabular} \\
   \midrule 
   LawBench \citep{fei2023lawbench} &
  Law &
  \begin{tabular}[c]{@{}l@{}}Legal QA, NER, \\
  Sentiment Analysis, \\
  Reading Comprehension
  \end{tabular}  &
  \begin{tabular}[c]{@{}l@{}} 
         F1, ACC,\\
        ROUGE-L,\\
         Normalized \\log-distance	...
   \end{tabular}
  & 
   \begin{tabular}[c]{@{}l@{}}GPT-4,\\ 
   ChatGPT, \\
   InternLM-Chat,\\
   StableBeluga2...
   \end{tabular} \\
   \midrule 
   OceanBench \citep{bi2023oceangpt} &
  Geoscience &
  \begin{tabular}[c]{@{}l@{}}15  ocean-\\related tasks
  \end{tabular}  &
  \begin{tabular}[c]{@{}l@{}} 
         ACC
   \end{tabular}
  & 
   \begin{tabular}[c]{@{}l@{}}GPT-4
   \end{tabular} \\
  \bottomrule
\end{tabular}
}
\end{adjustbox}

\end{table*}

In our examination of specialized Large Language Models (LLMs), we've identified an extensive array of datasets and benchmarks tailored to a variety of domains.  These resources not only serve as critical tools for evaluating the capabilities of LLMs but also facilitate advancements in specialized applications. We summarize them in Table~\ref{tab:domain-eval}. The distinction between this subsection and the previous two lies in its focus. This subsection delves deeper into factuality evaluation tailored to specific domains, while Sec \ref{sec:eval_benchmark} and \ref{sec:factuality-eval} primarily concentrate on general factuality evaluation, with only a portion of their content dedicated to datasets evaluating factuality within specific domains.

{\paragraph{Finance}}
\citet{xie2023pixiu} designed a financial natural language understanding and prediction evaluation benchmark dubbed FLARE, based on their collected financial instruction tuning dataset FIT. This benchmark is used to evaluate their FinMA model. It randomly selects validation sets from FIT to choose the best model checkpoint and utilizes distinct test sets for evaluation.
FLARE is a broader variant compared to the existing FLUE benchmark~\cite{shah2022flue} as it also encapsulates financial prediction tasks like stock movement prediction in addition to standard NLP tasks. 
The FLARE dataset includes several subtasks, such as sentiment analysis (FPB, FiQA-SA), news headline  classification (Headline), named entity recognition (NER), question answering (FinQA, ConvFinQA), and stock movement prediction (BigData22, ACL18, CIKM18).
Performance is gauged via a variety of metrics for each task, such as the accuracy and weighted F1 Score for sentiment analysis, entity-level F1 score for named entity recognition, and accuracy and Matthews correlation coefficient for stock movement prediction.
Several methods, including their own FIT-fine-tuned FinMA and other LLMs (BloombergGPT, GPT-4, ChatGPT, BLOOM, GPT-NeoX, OPT-66B, Vicuna-13B) are dedicated to their comparison. BloombergGPT's performance is assessed in various shot scenarios, while the zero-shot performance is reported for the remaining results. Some of the baselines depend on human evaluations given that LLMs without fine-tuning fail to generate instruction-defined answers. Conversely, FinMA's results are conducted on a zero-shot basis and can be evaluated automatically. 
To enable direct comparison between the performance of FinMA and BloombergGPT, despite the former not releasing their test datasets, test datasets were constructed with the same data distribution.

\citet{li2023ecomgpt} propose EcomInstruct benchmark, which experiment investigates the performance of their EcomGPT language model in comparison to baseline models such as BLOOM and BLOOMZ. Categories of these baseline models include pre-trained large models with decoder-only architecture like BLOOM, and instruction-following language models like BLOOMZ and ChatGPT. The evaluation metric of the EcomInstruct involves converting all tasks to generative paradigms and using text generation evaluation metrics like ROUGE-L. Classification tasks were evaluated with precision, recall, and F1 scores. The EcomInstruct dataset, comprising 12 tasks across four major categories, is divided into training and testing sections. The EcomGPT is trained on 85,746 instances of E-commerce data. The performance of the model is assessed based on its capacity to generalize unseen tasks or datasets, with emphasis on cross-language and cross-task paradigm settings.

{\paragraph{Medicine}}
\citet{wang2023cmb} propose a localized medical benchmark called CMB, or the Comprehensive Medical Benchmark in Chinese. CMB is rooted entirely in the native Chinese linguistic and cultural framework. While traditional Chinese medicine is a significant part of this evaluation, it does not make up the entire benchmark.
Both prominent LLMs, such as ChatGPT and GPT-4, and localized Chinese LLMs, including those specializing in the health domain, are evaluated using CMB. However, the benchmark is not designed as a leaderboard competition, but rather as a tool for self-assessment and understanding the progression of models in this field.
CMB embodies a comprehensive, multi-layered medical benchmark in Chinese, comprised of hundreds of thousands of multiple-choice questions and complex case consultation questions. This wide range of queries covers all clinical medical specialties and various professional levels, seeking to evaluate a model's medical knowledge and clinical consultation capabilities comprehensively.
 
\citet{li2023huatuo} introduce Huatuo-26M dataset, the largest Chinese medical Question and Answer (QA) dataset to date, including over 26 million high-quality medical QA pairs. It covers a wide range of topics, such as diseases, symptoms, treatments, and drug information. The dataset is a valuable resource for anyone looking to improve AI applications in the medical field, such as chatbots and intelligent diagnostic systems.
The Huatuo-26M dataset is gathered and integrated from various sources, including online medical encyclopedias, online medical knowledge bases, and online medical consultation records. Each QA pair in the dataset contains a problem description and a corresponding answer from a doctor or expert. Although a significant proportion of the Huatuo-26M dataset is constituted by the online medical consultation records, the text format data from these records is not publicly available, for unspecified reasons.
This dataset is expected to be instrumental for multiple types of research and AI applications in the medical field. These areas of application extend to Natural Language Processing tasks like developing QA systems, text classification, sentiment analysis, and Machine Learning model training tasks like disease prediction and personalized treatment recommendation. Consequently, the dataset is favorable for developing AI applications in the medical field, from intelligent diagnosis systems to medical consultation chatbots.

\citet{jin2023genegpt} use nine tasks related to NCBI resources to evaluate the proposed GeneGPT model. The dataset used is GeneTuring benchmark~\cite{Hou2023GeneTuring}, which contains 12 tasks with 50 question-answer pairs each. The tasks are divided into four modules - Nomenclature  (Gene alias, Gene name conversion),  Genomic location(Gene SNP association, Gene location, SNP location), 
  Functional analysis(Gene disease association, Protein-coding genes), 
  Sequence alignment (DNA to human genome, DNA to multiple species). Two settings of GeneGPT are assessed: a full setting where all prompt components are used, and a slim setting using only two components. The performance of GeneGPT is compared against various baselines including general-domain GPT-based LLMs like GPT-2, GPT-3, and ChatGPT. Additionally, GPT-2-sized biomedical domain-specific LLMs such as BioGPT and BioMedLM are evaluated. The new Bing, a retrieval-augmented LLM with access to relevant web pages, is also assessed. The result evaluation of the compared methods is based on the results reported in the original benchmark and are manually evaluated. However, the evaluation of the proposed GeneGPT method is determined through automatic evaluations.

\paragraph{Law}
LegalBench~\citep{guha2023legalbench} is a benchmark for legal reasoning introduced due to the increasing use of LLMs in the legal field. It consists of 162 tasks covering six types of legal reasoning and was collaboratively constructed through an interdisciplinary process with significant contributions from legal professionals. The tasks designed aim to either demonstrate practical legal reasoning capabilities or measure reasoning skills that are of interest to lawyers. To facilitate discussions between different fields relating to LLMs in law, LegalBench tasks correspond to popular legal frameworks for describing legal reasoning, thus creating a shared language between lawyers and LLM developers. The paper not only describes LegalBench, but also presents an evaluation of 20 different open-source and commercial LLMs and highlights the types of research opportunities that LegalBench can provide.

LawBench~\citep{fei2023lawbench} is an evaluation framework dedicated to assessing the capabilities of LLMs in relation to legal tasks. The context-specificity and high-stakes nature of the law field make it crucial to have a clear grasp of LLMs' legal knowledge and their ability to execute legal tasks. 
LawBench dives deep into three cognitive evaluations of LLMs: the capability to memorize crucial legal details, understanding legal texts, and applying legal knowledge to resolve complicated legal problems. A total of 20 diverse tasks have been put together, covering five main task categories: single-label classification, multi-label classification, regression, extraction, and generation. 
Throughout the evaluation process, 51 LLMs were extensively tested under LawBench, compiling a spectrum of language models including 20 multilingual LLMs, 22 Chinese-oriented LLMs, and 9 law-specific LLMs. The results concluded that GPT-4 ranked as the superior LLM in the law domain, significantly surpassing its competitors. Despite the noted improvements when LLMs were fine-tuned on law-specific texts, the study acknowledged that there is still a long road ahead in achieving highly reliable LLMs for legal tasks.

\citet{bi2023oceangpt}  design OceanBench to evaluate the capabilities of LLMs for oceanography tasks.
OceanBench includes a total of 15 type of ocean-related tasks, such as question-answering and description tasks.
The samples in OceanBench are are automatically generated from the seed dataset and have undergone manual verification by experts.

\section{Analysis of Factuality}
\label{sec:analysis}


In the previous Section \ref{sec:evalution}, we provide quantitative statistics related to evaluating factuality. In this section, we delve deeper, exploring the underlying mechanisms that influence factuality in large language models.
\subsection{Analysis of Factuality}
\label{sec:factuality_analysis}
This subsection delves into intriguing analyses concerning the factuality of LLMs, focusing on aspects that aren't directly tied to evaluation, or enhancement. Specifically, we explore the mechanisms through which LLMs process, interpret, and produce factual content. The subsequent sections offer an in-depth examination of different dimensions of factuality in LLMs, ranging from their knowledge storage, and awareness to their approach to managing conflicting data. 
\footnote{
While some research \citep{wang-etal-2021-generative,neeman-etal-2023-disentqa} applies some similar analyses to Pretrained Language Models. This review excludes them as they are not typically considered Large Language Models.}


\subsubsection{Knowledge Storage}
\label{sec:analy_store}
The language model serves as a repository of knowledge, storing a multitude of information about the world within its parameters ~\cite{petroni-etal-2019-language}. However, the organization of this knowledge within LLMs remains largely mysterious.
\cite{meng2022locating} introduce a methodology called causal tracing to measure the indirect impact of hidden states or activations. This technique was employed to illustrate that factual knowledge is primarily stored in the early-layer feed-forward networks (FFNs) of such models. 
Similarly, \citet{geva-etal-2021-transformer} also suggests that a substantial portion of factual information is encoded within the FFN layers. They conceptualize the input of the FFN as a query, the first layer as keys, and the second layer as values. Consequently, the intermediate hidden dimension of the FFN can be interpreted as the number of memories within the layer, and the intermediate hidden state represents a vector comprising activation values for each memory. As a result, the final output of the FFN can be understood as the weighted sum of activated values. The authors further demonstrate that the value vectors often encapsulate human-interpretable concepts and knowledge~\cite{geva-etal-2022-transformer,DBLP:journals/corr/abs-2304-14767}.
In addition, \citet{chen2023journey} have made an intriguing finding that the language model contains language-independent neurons that express multilingual knowledge and degenerate neurons that convey redundant information by applying the integrated gradients method~\cite{lundstrom2022rigorous}.
Nevertheless, it is important to note that the aforementioned studies primarily focus on the representation of individual facts, and the comprehensive understanding of how factual knowledge is precisely organized and interconnected within these models remains an ongoing challenge.

\subsubsection{Knowledge Completeness and Awareness}
\label{sec:analy_aware}

This subsubsection delves into the intriguing realm of LLMs' self-awareness, their capacity to discern their knowledge gaps, and the balance between their internally generated knowledge and externally retrieved information. We delve into the dichotomy between parametric knowledge and retrieved knowledge, exploring the promises and challenges these models bring to the forefront of knowledge-intensive tasks.

\paragraph{Knowledge Awareness}
Several studies have investigated the knowledge awareness of Large Language Models, specifically assessing whether LLMs can accurately estimate the correctness of their own responses. Most of these studies treat LLMs as "black boxes," prompting the models to report their confidence levels or calculating the perplexity of the model's output as an indicator of response likelihood. \citet{gou2023critic} explore the model's ability to validate and iteratively refine its outputs, akin to how humans interact with tools. The authors find that solely relying on self-correction without external feedback can lead to marginal improvements or even diminished performance. \cite{ren2023investigating} experiment with settings either augmented or not with external document retrieval to determine whether models recognize their own knowledge boundaries. Their findings indicate that LLMs possess an inaccurate perception of their factual knowledge boundaries and tend to be overly confident about their responses. LLMs often fail to fully harness the knowledge they possess; however, retrieval enhancement can somewhat compensate for this shortcoming. \citet{yin-etal-2023-large} introduce a dataset named ``SelfAware" to test if models recognize what they don't know, encompassing both answerable and unanswerable questions. The experiment suggests that models do possess some capacity to discern their own knowledge gaps, but they are still far from human levels. GPT-4 outperforms other models, instructions and In-Context-Learning \cite{ICL} can enhance a model's discriminatory ability. \citet{kadavath2022language} focus on LLM self-assessment based on Language Model calibration using multiple-choice questions. Their findings revealed that the "none of the above" option decreased accuracy, larger models showed better calibration, and RLHF hindered model calibration levels. However, simply adjusting the temperature parameter can rectify this issue.  \citet{azaria2023internal} assess the truthfulness of statements generated by LLMs, by using the model's internal state and hidden layer activations. The authors, employing a feedforward neural network, can classify if the model is misleading by utilizing the hidden output states.

\paragraph{Parametric Knowledge vs Retrieved Knowledge}
\citet{yu2023generate} explore whether the internal knowledge of LLMs can replace the retrieved documents on knowledge-intensive tasks. They ask LLMs, such as InstructGPT, to directly generate contexts given a question rather than retrieving them from the database. They find the generated documents contain the golden answers more often than the top retrieved documents. Then they feed the generated docs and retrieved docs to the Fusion-in-Decoder model \citep{Fusion-in-Decoder} for knowledge-intensive tasks such as Open-domain QA \citep{NQ} and find the generated docs are more effective than the retrieved docs, suggesting that the LLMs contain enough knowledge for knowledge-intensive tasks.

On the contrary, these observations have been contested in subsequent investigations. \citet{kandpal2023large} underscore the dependency of LLMs on the number of associated documents seen during pre-training. They argue that the success in answering fact-based questions is highly linked to the number of documents containing the topic of the question that were encountered in pre-training. The study further posited the necessity of scaling models extensively to achieve competitive performance for questions with minimum representation in the training data.
Adding to these concerns, \citet{sun2023head} critically evaluate the factual knowledge base of LLMs, using a specifically designed Head-to-Tail benchmark comprised of 18K question-answer pairs. The results show that the understanding of factual knowledge, particularly related to torso-to-tail entities, by currently available LLMs is suboptimal.

In summary, while LLMs show promise in handling knowledge-intensive tasks, their dependency on pre-training information and limitations in factual accuracy remain significant hurdles. It underscores the need for further advancements in the field and the importance of incorporating complementary methods, such as retrieval augmentation, to enhance the learning of long-tail knowledge in LLMs.



\subsubsection{Contextual Influence and Knowledge Conflict}
\label{sec:analy_confict}

This sub-subsection examines the interplay between an LLM's inherent parametric knowledge and the provided contextual knowledge, exploring both the model's capacity to utilize context and its behavior when confronted with conflicting information.

\paragraph{Contextual Influence on Generation}
Some works explore the model's capacity to utilize context, for example, \citet{li-etal-2023-large} observe that larger models tend to rely on their parametric knowledge, even when faced with counterfactual contexts. This suggests that as models increase in size, they might grow more confident in their internal knowledge, potentially sidelining external context. However, the introduction of irrelevant contexts can still influence their outputs. The balance between controllability (relying on relevant context) and robustness (resisting irrelevant context) emerges as a challenge in LLM training. The study indicates that reducing context noise improves controllability, but the effect on robustness remains to be seen.
In contrast, \citet{zhou2023context} propose prompt templates to guide LLMs towards more faithful text generation. Among these, opinion-based prompts prove most effective, indicating that when LLMs are queried about opinions, they adhere more closely to the context. Interestingly, the study finds that using counterfactual context enhances the model's faithfulness, while original context, sourced from platforms like Wikipedia, might induce a simplicity bias, leading LLMs to answer questions without heavily relying on the context.
\citet{chen2023benchmarking} conduct a comprehensive evaluation of LLMs' ability to effectively utilize retrieved information. The study reveals that while retrieved documents can boost LLM performance, the presence of noise in these documents can hinder it.
\citet{yue2023automatic} investigate the nature of LLM-generated content in relation to provided references. They categorize the generated content as attributable, contradictory, or extrapolatory to the reference. Both fine-tuned models and instruction-based LLMs struggle to accurately evaluate the alignment between generated content and references, underscoring the challenge of ensuring that LLMs produce content consistent with the provided context.
\citet{shi2023large} study the distractability of LLMs on the GSM-IC dataset derived from GSM8K \citep{cobbe2021training}. They discover that all the prompting techniques are responsive to irrelevant information in the problem definition. They identify various factors of irrelevant information that impact the model's sensitivity to irrelevant context. Moreover, they find that self-consistency prompting and incorporating irrelevant information into the exemplars can enhance models' performance, enabling them to learn to disregard irrelevant information.

\paragraph{Handling Knowledge Conflicts}
A series of studies are interested in LLMs' behavior when confronted with conflicting information. \citet{longpre2021entity} introduce the concept of knowledge conflicts, where the provided context contradicts the model's learned information. Their findings suggest that such conflicts lead to increased prediction uncertainty, especially for in-domain examples. Observations across models, ranging from T5-60M to 11B, indicate that larger models tend to default to their parametric knowledge. Moreover, there's an inverse relationship between retrieval quality and the tendency to rely on internal knowledge: the more irrelevant the evidence, the more the model defaults to its parametric knowledge.
\cite{chen2022rich} conduct experiments on typical ODQA models, including FiD and RAG. Their results show that FiD models rarely resort to memorization (less than 3.6\% for NQ) compared to RAG models. Instead, FiD primarily grounds its answers in the provided evidence. Interestingly, when confronted with conflicting retrieved passages, models tend to fall back on their parametric knowledge.
\cite{xie2023adaptive} explore the behavior of recent LLMs, including ChatGPT and GPT-4. Contrary to findings on smaller LMs, they discover that LLMs can be highly receptive to external evidence, even if it contradicts their parametric memory, provided the external evidence is coherent and convincing. Additionally, LLMs exhibit a strong confirmation bias, especially when presented with evidence that aligns with their parametric memory. This bias becomes even more pronounced for widely accepted knowledge. In scenarios where no relevant evidence is provided, LLMs tend to express uncertainty. However, when presented with both relevant and irrelevant evidence, they demonstrate an ability to filter out irrelevant information.
\citet{wang2023resolving} argue that LLMs should not rely solely on either parametric or non-parametric information, but grant LLM users the agency to make informed decisions. They introduce a framework including three tasks ((1) Contextual knowledge conflict detection; (2) QA-span knowledge conflict detection; (3) Distinct answers generation) to simulate knowledge conflicts and evaluate whether LLMs' behaviors align with the goal. 

In conclusion, while studies like \citet{li-etal-2023-large} and \citet{zhou2023context} emphasize the challenges and potential solutions in making LLMs more context-aware, others like \citet{yue2023automatic} and \citet{xie2023adaptive} highlight the inherent biases and limitations of LLMs. The overarching theme is the need for a balanced approach, where LLMs effectively leverage both their internal knowledge and external context to produce accurate and coherent outputs.

\subsection{Causes of Factual Errors}
\label{sec:causes}
Understanding the root causes of these factual inaccuracies is crucial for refining these models and ensuring their reliable application in real-world scenarios. In this subsection, we delve into the multifaceted origins of these errors, categorizing them based on the stages of model operation: Model Level, Retrieval Level, Generation Level, and other miscellaneous causes. Table~\ref{tab:causes} shows examples of factuality errors caused by different factors.

\subsubsection{Model-level Causes}
\label{sec:model-causes}
This subsection delves into the intrinsic factors within large language models that contribute to factual errors, originating from their inherent knowledge and capabilities. 

\paragraph{Domain Knowledge Deficit} The model may lack comprehensive expertise in specific domains, leading to inaccuracies. Every LLM has its limitations based on the data it was trained on. If an LLM hasn't been exposed to comprehensive data in a specific domain during its training, it's likely to produce inaccurate or generalized outputs when queried about that domain. For instance, while an LLM might be adept at answering general science questions, it might falter when asked about niche scientific subfields \citep{ScienceQA}.

\paragraph{Outdated Information} The model's dependence on older datasets can make it unaware of recent developments or changes. LLMs are trained on datasets that, at some point, become outdated. This means that any events, discoveries, or changes post-dating the last training update won't be known to the model. For example, ChatGPT and GPT-4 are both trained on data up to 2021.09 might not be aware of events or advancements after then.

\paragraph{Immemorization} The model does not always retain knowledge from its training corpus. While it's a misconception that LLMs ``memorize" data, they do form representations of knowledge based on their training. However, they might not always recall specific, less-emphasized details from their training parameters, especially if such details were rare or not reinforced through multiple examples. For example, ChatGPT has pretrained with Wikipedia, but it still fail in answering some questions from NaturalQuestions \citep{NQ} and TriviaQA \citep{TQ}, which are constructed from Wikipedia \citep{wang2023evaluating}.

\paragraph{Forgetting} The model might not retain knowledge from its training phase or could forget prior knowledge as it undergoes further training. As models are further fine-tuned or trained on new data, there's a risk of ``catastrophic forgetting" ~\citep{goodfellow2015empirical, wang2022preserving, chen2020recall, zhai2023investigating} where they might lose certain knowledge they were previously know. This is a well-known challenge in neural network training, where networks forget previously learned information when exposed to new data, which also happens in large language models \citep{luo2023empirical}.

\paragraph{Reasoning Failure} While the model might possess relevant knowledge, it can sometimes fail to reason with it effectively to answer queries. Even if an LLM has the requisite knowledge to answer a question, it might fail to connect the dots or reason logically. For instance, ambiguity in the input~\citep{liu2023we} can potentially lead to a failure in understanding by LLMs, consequently resulting in reasoning errors. In addition, \citet{berglund2023reversal} find the LLM traps in the reversal curse, for example, it knows that A is B's mother but fail to answer who is B's son. This is especially evident in complex multi-step reasoning tasks or when the model needs to infer based on a combination of facts \citep{tan2023chatgpt,kotha2023understanding}. 



\subsubsection{Retrieval-level Causes}
\label{sec:retrieval-causes}
The retrieval process plays a pivotal role in determining the accuracy of LLMs' responses, especially in retrieval-augmented settings. Several factors at this level can lead to factual errors:

\paragraph{Insufficient Information}
If the retrieved data doesn't provide enough context or details, the LLM might struggle to generate a factual response. This can result in generic or even incorrect outputs due to the lack of comprehensive evidence.

\paragraph{Misinformation Not Recognized by LLMs}
LLMs can sometimes accept and propagate misinformation present in the retrieved data. This is especially concerning when the model encounters knowledge conflicts, where the retrieved information contradicts its pre-trained knowledge, or multiple retrieved documents contradict each other~\cite{TruthDiscovery}. For instance, \cite{longpre2021entity} observed that the more irrelevant the evidence, the more likely the model is to rely on its intrinsic knowledge. Recent studies, such as \cite{pan2023risk}, have also shown that LLMs are susceptible to misinformation attacks within the retrieval process.

\paragraph{Distracting Information}
LLMs can be misled by irrelevant or distracting information in the retrieved data. For example, if the evidence mentions a "Russian movie" and "Directors", the LLM might incorrectly infer that "The director is Russian". \cite{luo2023sail} highlighted this vulnerability, noting that LLMs can be significantly impacted by distracting retrieval results. They further proposed instruction tuning as a potential solution to enhance the model's ability to sift through and leverage retrieval results more effectively.

Additionally, when dealing with long retrieval inputs, the models tend to show their best performance when processing information given at the beginning or end of the input context, according to \citet{liu2023lost}. In contrast, the models are likely to experience a significant decrease in performance when they are required to extract relevant data from the middle of these extensive contexts.

\paragraph{Misinterpretation of Related Information}
Even when the retrieved information is closely related to the query, LLMs can sometimes misunderstand or misinterpret it. While this might be less frequent when the retrieval process is optimized, it remains a potential source of error. For instance, in the ReAct study \citep{yao2023react}, the rate of errors dropped significantly when the retrieval process was improved.

\subsubsection{Inference-level Causes}
\label{sec:inference-causes}
\paragraph{Snowballing} During the generation process, a minor error or deviation at the beginning can compound as the model continues to generate content. 
For instance, if an LLM misinterprets a prompt or starts with an inaccurate premise, the subsequent content can veer further from the truth \citep{Snowball,varshney2023stitch}.

\paragraph{Erroneous Decoding} The decoding phase is crucial for translating the model's internal representations into human-readable content~\citep{chuang2023dola, massarelli2020decoding}. Mistakes during this phase, whether due to issues like beam search errors or suboptimal sampling strategies, can lead to outputs that misrepresent the model's actual "knowledge" or intention. This can manifest as inaccuracies, contradictions, or even nonsensical statements.

\paragraph{Exposure Bias}
LLMs are a product of their training data. If they've been exposed more frequently to certain types of content or phrasing, they might have a bias toward generating similar content, even when it's not the most factual or relevant. This bias can be especially pronounced if the training data has imbalances or if certain factual scenarios are underrepresented~\citep{felkner2023winoqueer, gallegos2023bias}. The model's outputs, in such cases, reflect its training exposure rather than objective factuality. For example, research by ~\cite{hossain-etal-2023-misgendered} suggests that LLMs can correctly identify the gender of individuals who conform to the binary gender system, but they perform poorly when determining non-binary or neutral genders.

\section{Enhancement}
\label{sec:enhancement}

\begin{figure*}[h]
    \centering
    \includegraphics[width=5.5in]{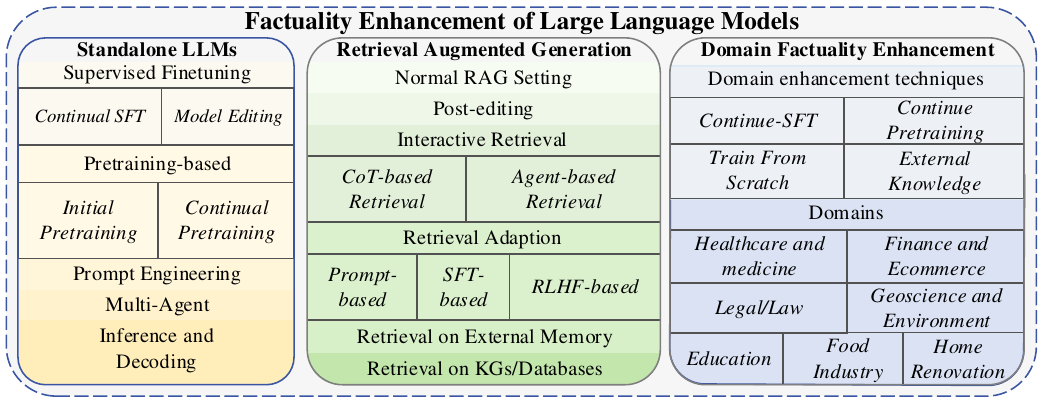}
    \vspace{-2mm}
    \caption{An overview of methods to enhance factuality in large language models. This encompasses three primary areas: enhancement techniques for pure LLMs, strategies for retrieval-augmented LLMs, and methods employed by domain-specific LLMs to boost their factual accuracy within their respective domains. 
    }
    \vspace{-3mm}
    \label{fig:enhancement}
\end{figure*}

This section discusses methods to enhance factuality in LLMs across different phases, including LLM generation, retrieval-augmented generation, inference-phase enhancements, and domain-specific factuality improvements, outlined in Figure~\ref{fig:enhancement}.

\begin{table*}[t]\small
\centering
\caption{Performance of Selected Factuality Enhancement Methods. The table displays performance metrics on various datasets for both baseline models and their enhanced counterparts, denoted with a $\rightarrow$. Due to space constraints, only a subset of datasets, metrics, and models from each work is presented.}
\label{tab:enhance-table}
\vspace{-2mm}
\begin{adjustbox}{max width=\textwidth}
  \setlength{\tabcolsep}{.3mm}
    \renewcommand{\arraystretch}{1.2}
{\begin{tabular}{ccclc|c|ccl}
\toprule 
Reference & Dataset & Metrics & Baselines $\rightarrow$ Theirs &&& Dataset & Metrics & Baselines $\rightarrow$ Theirs \\     
\midrule  
\citet{li2022decoupled} & NQ & EM & 34.5 $\rightarrow$ 44.35 (T5 11B) & & & GSM8K & ACC & 77.0 $\rightarrow$ 85.0 (ChatGPT) \\
\midrule  
GenRead \citep{yu2023generate} & NQ & EM & 20.9 $\rightarrow$ 28.0 (InstructGPT) &&& TriviaQA & EM & 57.5 $\rightarrow$ 59.0 (InstructGPT) \\
& & & &&& WebQA & EM & 18.6 $\rightarrow$ 24.6 (InstructGPT) \\
\midrule  
DoLa \citep{chuang2023dola} & FACTOR News & ACC & 58.3 $\rightarrow$ 62.0 (LLaMa-7B) &&& FACTOR News & ACC & 61.1 $\rightarrow$ 62.5 (LLaMa-13B) \\
& FACTOR News & ACC& 63.8 $\rightarrow$ 65.4 (LLaMa-33B) &&& FACTOR News & ACC& 63.6 $\rightarrow$ 66.2 (LLaMa-65B) \\
& FACTOR Wiki  & ACC & 58.6 $\rightarrow$ 62.2 (LLaMa-7B) &&& FACTOR Wiki  & ACC & 62.6 $\rightarrow$ 66.2 (LLaMa-13B) \\
& FACTOR Wiki  & ACC& 69.5 $\rightarrow$ 70.3 (LLaMa-33B) &&& FACTOR Wiki  & ACC& 72.2 $\rightarrow$ 72.4 (LLaMa-65B) \\
& TruthfulQA & \%Truth * Info & 32.4 $\rightarrow$ 44.6 (LLaMa-13B) &&& TruthfulQA & \%Truth * Info  & 34.8 $\rightarrow$ 49.2 (LLaMa-65B) \\
\midrule
CD \citep{li2022contrastive} & TruthfulQA & \%Truth * Info & 32.4 $\rightarrow$ 44.4 (LLaMa-13B) &&& TruthfulQA & \%Truth * Info & 31.7 $\rightarrow$ 36.7 (LLaMa-33B) \\
& TruthfulQA & \%Truth * Info & 34.8 $\rightarrow$ 43.4 (LLaMa-65B) &&& \\
\midrule
ITI \citep{li2023inferencetime} & NQ & ACC & 46.6 $\rightarrow$ 51.3 (LLaMA-7B) &&& TriviaQA & ACC & 89.6 $\rightarrow$ 91.1 (LLaMA-7B) \\
& MMLU& ACC  & 35.7 $\rightarrow$ 40.1 (LLaMA-7B) &&& TruthfulQA & \%Truth * Info & 32.5 $\rightarrow$ 65.1 (Alpaca) \\
&  TruthfulQA & \%Truth * Info & 26.9 $\rightarrow$ 43.5 (LLaMa-7B) &&&  TruthfulQA & \%Truth * Info & 51.5 $\rightarrow$ 74.0 (Vicuna) \\
\midrule
FactTune \citep{ovadia2023finetuning} & Biographies  & ACC & 74.8 $\rightarrow$ 83.1 (Llama-2-Chat) &&& Medical QA & ACC & 63.6 $\rightarrow$ 68.2 (Llama-2-Chat)  \\
\midrule
LM vs LM \citep{cohen2023lm} & LAMA & F1 & 50.7 $\rightarrow$ 80.8 (ChatGPT) &&& TriviaQA & F1 & 56.2 $\rightarrow$ 82.6 (ChatGPT) \\
& NQ & F1 & 60.6 $\rightarrow$ 79.1 (ChatGPT) &&& PopQA & F1 & 65.2 $\rightarrow$ 85.4 (ChatGPT) \\
& LAMA & F1 & 42.5 $\rightarrow$ 79.3 (GPT-3) &&& TriviaQA & F1 & 46.7 $\rightarrow$ 77.2 (GPT-3) \\
& NQ & F1 & 52.0 $\rightarrow$ 78.0 (GPT-3) &&& PopQA & F1 & 43.7 $\rightarrow$ 77.4 (GPT-3) \\
\midrule
\citet{weller2023according} & TriviaQA & QUIP & 31.6 $\rightarrow$ 33.6 (ChatGPT) &&& NQ & QUIP & 32.8 $\rightarrow$ 34.3 (ChatGPT) \\
& HotpotQA & QUIP & 28.3 $\rightarrow$ 29.2 (ChatGPT) &&& ELI5  & QUIP & 24.1 $\rightarrow$ 26.5 (ChatGPT) \\
& TriviaQA & EM & 77.8 $\rightarrow$ 78.8 (ChatGPT) &&& NQ & EM & 32.9 $\rightarrow$ 34.8 (ChatGPT) \\
& HotpotQA & F1 & 35.7 $\rightarrow$ 36.6 (ChatGPT) &&& ELI5  & R-L & 22.7 $\rightarrow$ 21.7 (ChatGPT) \\
\midrule
Chain-of-Verification \citep{dhuliawala2023chain} & MultiSpanQA & F1 & 39.0 $\rightarrow$ 48.0 (LLaMA 65B) &&& - & FactScore & 55.9 $\rightarrow$ 71.4 (LLaMA 65B) \\
&&&&&& - & Avg. \# facts & 16.6$\rightarrow$ 12.3  (LLaMA 65B)  \\
\midrule
ReAct \citep{yao2023react} & HotpotQA & EM & 28.7 $\rightarrow$ 35.1 (PaLM-540B) &&& FEVER & ACC & 57.1 $\rightarrow$ 62.0 (PaLM-540B)\\
\midrule
\citet{jiang2023active} & 2WikiMultihopQA & EM & 28.2 $\rightarrow$ 51 (ChatGPT) &&& StrategyQA & EM & 72.9 $\rightarrow$ 77.3 (ChatGPT) \\
& WikiAsp & UniEval & 47.1 $\rightarrow$ 53.4 (ChatGPT) &&& ASQA & EM & 33.8 $\rightarrow$ 41.3 (ChatGPT) \\
&&&&&& ASQA-hint & EM & 40.1 $\rightarrow$ 46.2 (ChatGPT) \\
\midrule
Atlas \citep{Atlas} & MMLU & ACC & 42.4 $\rightarrow$ 56.3 (T5-770M) &&& MMLU & ACC & 50.4 $\rightarrow$ 59.9 (T5-3B) \\
&&&&&& MMLU & ACC & 54 $\rightarrow$ 65.8 (T5-11B) \\
\midrule 
REPLUG \citep{REPLUG} & MMLU & ACC & 68.3 $\rightarrow$ 73.2 (Codex) &&& NQ & EM & 40.6 $\rightarrow$ 45.5(Codex) \\
&&&&&& TriviaQA & EM & 73.6 $\rightarrow$ 77.3 (Codex) \\
\midrule 
Self-RAG \citep{asai2023selfrag} & TriviaQA & ACC & 30.0 $\rightarrow$ 72.4 (LLaMA2-7B) &&& ALCE-ASQA & EM & 15.2 $\rightarrow$ 30.0 (LLaMA2-13B)\\
& TriviaQA & ACC &30.2 $\rightarrow$ 74.5 (LLaMA2-7B)&&&   ALCE-ASQA & EM & 16.3 $\rightarrow$ 31.7 (LLaMA2-13B) \\
\bottomrule
\end{tabular}
}
\end{adjustbox}
\end{table*}

Table~\ref{tab:enhance-table} provides a summary of enhancement methods and their respective improvements over baseline LLMs. It's essential to recognize that various research papers may employ distinct experimental settings, such as zero-shot, few-shot, or full settings. Consequently, when examining this table, it's important to note that performance metrics for different methods, even when evaluating the same metric on the same dataset, may not be directly comparable.

\subsection{On Standalone LLM Generation}
\label{sec:enhace_pure}

When focusing on standalone LLM generation, enhancement strategies can be broadly grouped into three main categories:

\emph{(1)
Improving Factual Knowledge from Unsupervised Corpora} (Sec \ref{sec:enhace_pure_pre}):
This involves refining the training data during pretraining, such as through deduplication and emphasizing informative words \citep{lee-etal-2022-deduplicating}. Techniques like TOPICPREFIX~\citep{lee2022factuality} and sentence completion loss are also explored to enhance this approach.

\emph{(2)
Enhancing Factual Knowledge from Supervised Data} (Sec \ref{sec:enhace_pure_sft}):
Examples in this category include supervised fine-tuning strategies~\citep{chung2022scaling,LIMA} focus on finetuning with labelled data or integrating structured knowledge such as knowledge graphs (KGs) or making precise adjustments to model parameters \citep{li2023inferencetime}.

\emph{(3)
Optimally Eliciting Factual Knowledge from the Model} (Sec \ref{sec:enhace_pure_multi}, \ref{sec:enhace_pure_prompt}, \ref{sec:enhace_pure_decoding}):
This category encompasses methods like Multi-agent collaboration~\citep{multiagent_debate} and innovative prompts~\citep{yu2023generate}. Additionally, novel decoding methodologies, such as factual-nucleus sampling, are introduced to further improve factuality \citep{lee2022factuality, chuang2023dola}.


Some work \citep{liu2023mitigating,sun2023aligning} aim for improve the factuality of large multi-modal models, we choose to not emphasize them.

\subsubsection{Pretraining-based}
\label{sec:enhace_pure_pre}
Pretraining plays a pivotal role in equipping the model with the factual knowledge derived from the corpus. By emphasizing strategies during this phase, the model's inherent factuality can be significantly enhanced. This approach is particularly crucial for addressing challenges like immemorization and forgetting.

\paragraph{\textbf{Initial Pretraining}}
Methods employed during the foundational pretraining of the model.

\citet{lee-etal-2022-deduplicating} develop two distinct tools for deduplicating training data, addressing the issue of redundant texts and long repeated substrings present in the training sets of current LLMs. These tools effectively reduce the recall of memorized texts in model outputs. Remarkably, they achieve similar or even superior accuracy with fewer training steps.

\citet{sadeq-etal-2023-unsupervised} introduce a modification to the Masked Language Model (MLM) training objective used in the pretraining of LLMs. They discover that high-frequency words do not consistently contribute to the model's ability to learn factual knowledge. To address this, they devise a strategy that encourages the language model to prioritize informative words during unsupervised training. This is achieved by masking tokens more frequently based on their informative relevance. To quantify this relevance, they utilize Pointwise Mutual Information (PMI) \citep{PMI}, positing that words with elevated PMI values, in relation to their adjacent words, are likely to be more informatively pertinent. Experimental results indicate that this innovative approach significantly bolsters the efficacy of pretrained language models across various tasks, including factual recall, question answering, sentiment analysis, and natural language inference, in a closed-book setting.

\paragraph{\textbf{Continual Pretraining}}
Iterative pretraining processes that allow the model to progressively refine and update its knowledge base.

\citet{lee2022factuality} introduce TOPICPREFIX as a pre-processing method and the sentence completion loss as training objective. Some factual sentences may be unclear when the LM training corpus is chunked, especially when these sentences contain pronouns (e.g., she, he, it). So they prepend TOPICPREFIX (e.g., Wikipedia document name) to sentences in the factual documents to transform each sentence into an independent factual statement. They also introduce the sentence completion loss, with the aim of enabling the model to capture facts from entire sentences rather than just focusing on the associations between subwords. For implementation, they establish a pivot $t$ for each sentence and require zero-masking for all token prediction losses before $t$. This pivot is only necessary during the training phase. Experiments show that such methods can further reduce the factual errors than standard factual-domain adaptive training.

\subsubsection{Supervised Finetuning}
\label{sec:enhace_pure_sft}

Supervised fine-tuning leverages labeled datasets to refine the model's performance. This approach serves a dual purpose: it imparts specific task or knowledge base-oriented information to the model and addresses challenges like immemorization and forgetting. Several studies, such as \citet{chung2022scaling,LIMA}, emphasize the pivotal role of supervised fine-tuning in eliciting the inherent knowledge of the base model, subsequently enhancing its reasoning capabilities.

\paragraph{\textbf{Continual SFT}}
A cyclic fine-tuning approach, where the model undergoes consistent refinement using sequential sets of labeled data.

\citet{moiseev-etal-2022-skill} investigate how to inject structured knowledge from a knowledge graph into LLMs. The approach involves directly training T5 using triplets containing relationship knowledge. Previous methods often describe the triplets using prompts and then train LLMs with masked language model task, but some triplets are not easy to describe. They compare three knowledge-enhancement fine-tuning methods, which are MLM training on the C4 corpus \citep{C4}, masking the subject or object in KG triplets, and masking the subject or object in the KELM corpus \citep{KELM}. Experiments show that effectiveness of the latter two methods achieved better exact match scores in closed-book QA tasks \citep{jiang-etal-2019-freebaseqa, wikihop, TQ, NQ}. This demonstrates that training directly based on KG triplets is an effective way to inject knowledge into the model

\citet{sun2023contrastive} introduce negative samples and contrastive learning to supervised finetuning process with MLE loss. The sources of these negative samples are either from the Knowledge Graph or are generated by large language models. While traditional contrastive learning can only function at the token or sentence level, the intrinsic value of span information cannot be overlooked, they employ Named Entity Recognition to extract this critical span data. The training process incorporates a blend of MLE loss, standard contrastive learning loss, and span-based contrastive learning loss, with the parameters finely tuned to optimize results. Experiments indicate that the method delivers performance comparable with SOTA KB-based methods but offers significant benefits in efficiency and scalability.

\citet{yang2023chatgpt} presents a development framework for Knowledge Graph-enhanced Large Language Models (KGLLMs), drawing from established technologies. They detail various enhancement strategies, including a before-training enhancement that refines input quality by integrating factual data; a during-training enhancement that synergizes textual and structural knowledge, utilizing tools like graph neural networks and attention mechanisms; multi-task learning which focuses on knowledge-guided pre-training tasks to bolster the factual knowledge acquisition of LLMs; and a post-training enhancement that fine-tunes LLMs for domain-specific tasks using knowledge-rich data. Additionally, the significance of prompt learning in LLMs is highlighted, emphasizing the importance of choosing appropriate prompt templates. They also suggest knowledge graphs as a valuable resource for crafting these templates to harness domain-specific knowledge.

FactTune \citep{ovadia2023finetuning} aims to use Reinforcement Learning to improve the factuality of LLM, so scoring the factuality of the content generated by the model is needed. However, human labels are too costly, so authors use two methods to check the factuality of the content generated by LLM, namely 1) Reference-based, which uses external knowledge base retrieval and Factscore \citep{Min2023-FActScore} method to score; 2) Reference-free, which uses Another LLM's confidence to score. Then, they utilized the Directly Preference Optimization \citep{rafailov2023direct} algorithm to optimize the model. 50\% and 20+\% error rate reductions were observed in Llama's bio generation tasks and medical QA respectively.

\paragraph{\textbf{Model Editing}}
Instead of directly finetuning the model, model edit is a more precise approach to enhance the model's factuality. By editing specific areas that are related to the fact, the model can correctly express that fact without compromising other unrelated knowledge. Current editing methods~\citep{yao2023editing} can be categorized into weight-preserved and weight-modified paradigms. 
When modifying the model's weight, KN~\citep{dai-etal-2022-knowledge} and ROME~\citep{meng2022locating} first analyze representations to locate those underlying factual errors and then directly update the relevant weights. Meanwhile, KE~\citep{de-cao-etal-2021-editing} and MEND~\citep{mitchell2022fast} employ a hypernetwork to learn the necessary weight changes. While effective, the robustness and generalization of directly updating weights remains an open question.

\citet{li2023inferencetime} present a technique known as Inference-Time Intervention (ITI) designed to boost the factual accuracy of large language models.  ITI adjusts model activations during inference to enhance the truthfulness of responses. This method, categorized under activation editing, is both adjustable and less intrusive, setting it apart from weight editing techniques. Drawing inspiration from prior research, they utilize steering vectors, which are proven effective for style transfer, to guide model activations. They detail a multi-step model selection process involving the calibration of intervention strength hyperparameters, pinpointing truth-telling related heads, and determining their truth-telling directions. The TruthfulQA dataset serves as the foundation for training and validation, with a rigorous 2-fold cross-validation employed to prevent data leakage. Experiments on the TruthfulQA benchmark demonstrate that ITI improves Alpaca's \citep{alpaca} truthfulness from 32.5\% to 65.1\%.





\subsubsection{Multi-Agent}
\label{sec:enhace_pure_multi}

Engaging multiple models in a collaborative or competitive manner, enhancing factuality through their collective prowess, helps in the immemorization and reasoning failure problems.

\citet{multiagent_debate} propose an approach to enhance the performance of language models by treating different LLMs as intelligent agents engaged in multi-agent debates. In this method, multiple instances of language models present and debate their respective answers and reasoning processes, ultimately reaching a consensus on the final answer after multiple rounds of debate. If the debate's answers fail to converge, prompts are modified to reduce the stubbornness of the two agents. This approach has been demonstrated to significantly improve mathematical and reasoning abilities across various tasks while enhancing the factual accuracy of generated content. Moreover, this method can be directly applied to existing black-box models, making it applicable to all research tasks using the same prompts.

\citet{cohen2023lm} develop a fact-checking mechanism. Drawing parallels to a scenario where a witness is interrogated for the veracity of their claims, they utilize a LLM to gather statements from a QA dataset, which are either factually correct or incorrect. During the statement generation, the model is provided with a golden answer, prompting it to produce both accurate and inaccurate statements, inherently labeling each statement. For every QA pair, another LLM, acting as an interrogator, generates a series of questions. A separate LLM, playing the role of the respondent, answers these queries. This iterative questioning and answering continues until the interrogator is satisfied, culminating in a conclusion. The authors conduct experiments on datasets like LAMA \citep{LAMA}, PopQA \citep{PopQA}, NaturalQA \citep{NQ}, and TriviaQA \citep{TQ}. The precision is the portion of incorrect claims, out of the claims rejected by the examiner and the recall is the portion of incorrect claims rejected by the examiner, out of all the incorrect claims. Measured by the F1 score, the LM vs LM method consistently outperforms the baseline by a significant margin, ranging from ten to over twenty points across all datasets.

\subsubsection{Novel Prompt}
\label{sec:enhace_pure_prompt}

Introducing innovative or tailored prompts to extract more factual and precise responses from the LLM, can better assist the model elicit the knowledge in its parameters and improve the reasoning ability.

\citet{yu2023generate} introduce a novel approach called Generate-then-Read (GENREAD). Document retrievers are replaced by LLM generators. In this article, the LLM is prompted to generate multiple  contextual documents on a given question. The authors clustered these document embeddings, and sampled documents from different clusters to ensure diversity of contextual documents. With these generated in-context demonstrations, LLMs achieved better results on knowledge-intensive tasks than retrieving from external corpus such as Wikipedia.

\citet{weller2023according} introduce a metric called QUIP-Score to measure the dependency on pre-trained data. They index Wikipedia to swiftly determine the dependency of an LLM's response. By using specific prompts, like "Based on evidence from Wikipedia:" they aim to evoke the LLM's recall of content from its training dataset. In addition to grounding prompts, they also introduce anti-grounding prompts to encourage the LLM to respond without referencing its training data, for instance, "Respond without using any information from Wikipedia." The motivation behind this approach is the belief that guiding the LLM to reference more of the knowledge it acquires during pre-training can reduce the generation of incorrect information. To quantify this grounding, they propose the QUIP-Score metric to gauge the similarity between the model's generated content and the most relevant content in Wikipedia. In their experiments conducted on datasets like TQ, NQ, HotpotQA, and ELI5, the results show that while adding the said prompt doesn't significantly improve traditional QA metrics, it notably boosts the scores on the QUIP metric.

\citet{khot2023decomposed} present Decomposed Prompting. Complex tasks are broken down into multiple simpler tasks via prompting and then it can be addressed by task-specific LLMs. For instance, for tasks involving extremely long input sequences, this technique systematically decomposes the input into shorter sequences for individual processing. Notably, the authors observed that when combined with a retrieval module, this approach significantly enhances performance on open domain multi-hop QA tasks.

\citet{dhuliawala2023chain} present Chain-of-Verification (CoVe) to reduce factual errors.The CoVe strategy involves the model initially crafting a response, subsequently formulating verification queries to assess its initial draft, independently responding to these queries to maintain unbiased answers, and ultimately producing a validated reply. The CoVe method encompasses four pivotal stages: 1) Drafting an initial reply based on a query using the LLM, 2) Generating verification questions from the query and initial answer to pinpoint potential errors, 3) Responding to each verification question and comparing these answers with the initial response to detect discrepancies, and 4) If discrepancies are found, producing a revised answer that integrates the verification outcomes. The entire procedure is executed by prompting the same LLM differently to achieve the intended results. Experiments show that Chain-of-Verification can reduce errors in diverse tasks, including list-based questions from Wikidata \citep{wikidata}, closed book MultiSpanQA \citep{multispanqa} and longform text generation \citep{Min2023-FActScore}.

\subsubsection{Decoding}
\label{sec:enhace_pure_decoding}

Decoding methodologies, such as beam search and nucleus sampling, play a crucial role in directing the model to produce outputs that are both factual and coherent. By refining the decoding process, challenges like snowballing errors or erroneous decoding, as detailed in Sec \ref{sec:inference-causes}, can be effectively addressed.

\citet{lee2022factuality} propose a new decoding sampling algorithm called factual-nucleus sampling that achieves a better trade-off between generation quality and factuality when compared to prevailing decoding algorithms. They postulate that the randomness of sampling is more harmful to factuality when applied to the generation of the latter portion of a sentence as opposed to its initial segment. So the factual-nucleus sampling algorithm, an adaptation of nucleus sampling, can dynamically adjust the 'nucleus' probability throughout the generation of each sentence, progressively reducing randomness with each successive generation step. The $\omega$-bound parameter is provided to prevent the p-value from becoming too small and hurting diversity. Experimental results show that a factual-nucleus sampling algorithm can improve the factuality of generation while maintaining generation quality, e.g.,  diversity and repetition.

\citet{chuang2023dola}  propose ``Decoding by Contrasting Layers" to mitigate these hallucinations. This approach leverages the differences in logits obtained from projecting the later layers versus earlier layers to the vocabulary space, taking advantage of the known localization of factual knowledge in LLMs. The results of this study demonstrate that DoLa consistently enhances the truthfulness of LLM-generated content across various tasks, such as multiple-choice and open-ended generation tasks, showcasing its potential to significantly improve the reliability of LLMs in generating accurate and truthful facts.

\subsection{On Retrieval-Augmented Generation}
\label{sec:enhace_rag}

Retrieval-Augmented Generation (RAG) has emerged as a widely adopted approach to address certain limitations inherent to standalone LLMs, such as outdated information and the inability to memorize \citep{LlamaIndex,LangChain}. These challenges are elaborated upon in Sec \ref{sec:model-causes}. Yet, while RAG offers solutions to some issues, it introduces its own set of challenges, including the potential for insufficient information and the misinterpretation of related data, as detailed in Sec \ref{sec:retrieval-causes}. This subsection delves into various strategies devised to mitigate these challenges. Within the realm of retrieval-augmented generation, enhancement techniques can be broadly categorized into several pivotal areas:

\emph{(1)
The Normal Setting of Utilizing Retrieved Text for Generations} (Sec \ref{sec:enhace_rag_normal}):

\emph{(2)
Interactive Retrieval and Generation} (Sec \ref{sec:enhace_rag_inter}):
Examples here include the integration of Chain-of-Thoughts steps into query retrieval \citep{he2022rethinking} and the use of an LLM-based agent framework that taps into external knowledge APIs \citep{yao2023react}.

\emph{(3)
Adapting LLMs to the RAG Setting} (Sec \ref{sec:enhace_rag_adapt}):
This involves methods like the one proposed by \citet{peng2023check}, which combines a fixed LLM with a plug-and-play retrieval module. Another notable approach is REPLUG \citep{REPLUG}, a retrieval-augmented framework that treats the LLM as a black box and fine-tunes retrieval models using language modeling scores.

\emph{(4)
Retrieving from Additional Knowledge Bases} (Sec \ref{sec:enhace_rag_kg} and Sec \ref{sec:enhace_rag_mem}):
This category includes methods that retrieve from external parametric memories \citep{chen2023purr} or knowledge graphs \citep{zhang2023mitigating} to enhance the model's knowledge base.

\subsubsection{Normal RAG Setting}
\label{sec:enhace_rag_normal}


\paragraph{Workflow of Normal RAG Setting}
A normal RAG setting works by retrieving external data and passing it to a LLM during the generation phase. We follow the framework proposed by LlamaIndex~\citep{LlamaIndex} and LangChain~\citep{LangChain} to decouple the process into the following modules and steps:

\emph{(1) Document loaders} are used to load documents from various sources. These loaders can fetch different types of documents (HTML, PDF, code) from various locations (private s3 buckets, public websites).
    
\emph{(2) Document transformers} are employed to extract relevant parts of the documents. This may involve splitting or chunking large documents into smaller chunks. Different algorithms for this task are employed, optimized for specific document types like code or markdown.

\emph{(3) Text embedding models} are employed to capture the semantic meaning of the text. 

\emph{(4) Vector stores} are employed to efficiently store and search the embeddings. 

\emph{(5) Retrievers}, such as the Parent Document Retriever, Self Query Retriever, and Ensemble Retriever, are used to retrieve the data from the database. 

Here, the Parent Document Retriever allows for the creation of multiple embeddings per parent document to retrieve smaller chunks while maintaining a larger context. The Self Query Retriever separates the semantic part of a query from other metadata filters, allowing for more accurate retrieval. The Ensemble Retriever enables the retrieval of documents from multiple sources or using different algorithms.

\citet{borgeaud2022improving} suggest scaling the size of the text database for retrieval as a complementary path to scaling language models. There is a pre-collected text database with a total of over 5 trillion tokens. These chunks are stored in the form of key-value pairs, with each chunk as a unit, and similarity retrieval is performed on  $k$-nearest neighbours from key-value database using the $L_2$ distance on Bert embeddings. The input sequence is splited into chunks, Retrieval Transformer (Retro) model retrieves text similar to the previous chunk to improve the predictions in the current chunk. The model calculate cross-attention between the input text and the retrieved text chunks to generate better answers. With only 25× fewer parameters of GPT-3, its performance on Pile is quite comparable. 

\citet{lazaridou2022internetaugmented} present a method that capitalizes on the few-shot capabilities of large-scale language models to enhance their grounding in factual and current information. Drawing from semi-parametric language models, the approach conditions LMs using few-shot prompts based on data sourced from Google Search. For any given query, the method retrieves pertinent documents from the web, extracts the top 20 URLs, and processes them to obtain clear text. These documents are segmented into paragraphs, and the most relevant ones are chosen using TF-IDF based on their similarity to the query. The LMs are then conditioned using few-shot prompts that incorporate the retrieved paragraphs. This k-shot prompting technique is augmented with an evidence paragraph, creating a prompt structure that encompasses evidence, query, and response. The method also involves generating multiple answers from the model and reranking them using different probabilistic factorizations. The experimental results indicate that by conditioning on retrieved evidence, the 7B Gopher LM \citep{Gopher} surpassed the performance of the 280B Gopher LM, with relative improvements reaching up to 30\% on the NQ \citep{NQ} dataset.



\subsubsection{Interactive Retrieval}
\label{sec:enhace_rag_inter}

While retrieval systems are designed to source relevant information, they may occasionally fail to retrieve accurate or comprehensive data. Additionally, LLMs might struggle to recognize, or even be misled by, the retrieved content, as detailed in Sec \ref{sec:retrieval-causes}. Implementing an interactive retrieval mechanism can address these challenges, aiding in sourcing more appropriate information and guiding the LLM towards improved content generation.
In this subsubsection, we explore methods that employ the Chain-of-Thoughts and Agents mechanisms to achieve effective interactive retrieval.

\paragraph{\textbf{CoT-based Retrieval}}
In recent studies, there is a growing interest in integrating Chain-of-Thoughts \citep{CoT} steps into query retrieval. \citet{he2022rethinking} introduce a method that generates multiple reasoning paths and their corresponding predictions for each query. This process involves retrieving relevant knowledge from external sources like Wikidata \citep{wikidata}, WordNet \citep{wordnet}, and ConceptNet \citep{conceptnet}. The faithfulness of each reasoning path determines based on entailment scores, contradiction scores, and MPNet similarities \citep{mpnet} with the retrieved knowledge. The prediction with the highest faithfulness score is chosen as the final result. This method demonstrates superior performance in tasks such as commonsense reasoning \citep{geva2021did}, temporal reasoning \citep{TempQuestions}, and tabular reasoning \citep{gupta2020infotabs}, outperforming the baseline CoT reasoning and self-consistency methods \citep{wang2023selfconsistency}. On a related note, \citet{trivedi-etal-2023-interleaving} propose IRCoT, an innovative retrieval technique that interweaves the CoT process. In this approach, each generated sentence during the CoT combines with the question to form a retrieval query. The subsequent reasoning step then produces by the Language Model using both the retrieval results and the prior reasoning. This interleaving method finds to enhance the performance of both retrieval and CoT in Open-domain QA. Experiments proves that this is beneficial for models of different sizes, including GPT-3(175B) \citep{GPT-3} and Flan-T5 \citep{chung2022scaling} families.

FLARE \citep{jiang2023active} is a dynamic solution to address the limitations of previous RAG works which either retrieve only once at the onset of generation or do so based on fixed intervals. Single retrievals are insufficient for long-form generation due to the evolving information needs during the process, and fixed intervals for previous token queries can be inappropriate, FLARE determines "when and what to retrieve" dynamically. The decision of "when" is based on whether the current sentence contains a token with a generation probability below a set threshold. If it doesn't, the sentence is accepted and moves to the next generation step; otherwise, retrieval augmented generation occurs. For the "what", the current sentence is used as a query. To address the challenge of low-probability tokens affecting retrieval accuracy, two solutions are proposed: masking low-probability tokens, and using LLM for generation of these tokens as queries. Testing on tasks like Multihop QA \citep{2WikiMultihopQA}, Commonsense Reasoning \citep{StrategyQA}, Long-form QA \citep{ASQA}, and Open-domain Summarization \citep{wikiasp} using GPT3.5, results show FLARE outperforms baselines, with both query generation methods showing comparable performance.

\paragraph{\textbf{Agent-based Retrieval}} 
Using an LLM-based agent framework that leverages external knowledge APIs as tools or requesting such APIs as actions.

\citet{yao2023react} present a new framework named ReAct that integrates Chain-of-Thoughts reasoning with action. Through in-context learning, the LLM's CoT output is transformed into descriptions of reasoning processes and action behaviors. Subsequently, these action descriptions are standardized and executed, with the results being incorporated into the next prompt. In terms of results, for tasks like Fact checking and QA, while CoT has 14\% of its correct answers containing incorrect reasoning steps or facts, ReAct only has 6\%. In the incorrect answers, 56\% of CoT's errors were due to errors in reasoning steps or facts, whereas ReAct has no factual errors. However, ReAct have 23\% of its errors resulting from search errors.

\citet{shinn2023reflexion} propose a prompt engineering framework named Reflextion to enable LLMs to reflect on and correct previous errors. They use linguistic feedback to strengthen an agent's actions instead of adjusting model weights. Specifically, the Reflextion Agent, an LLM, first interacts with its environment by generating an "Action" via ReAct \citep{yao2023react} or Chain-og-Thoughts \citep{CoT} in few-shot scenarios, which results in an "Observation". This "Observation", whether a reward, error message, or natural language feedback, provides insights on the Agent's current "Action". When the Agent receives a failure signal, it triggers a self-reflextion mechanism, utilizing the LLM, to summarize the reasons for the failure into its Memory module, creating a "long-term memory". On subsequent generations, the Agent can review all past reflextion memories to prevent mistakes. Experimental findings indicate that Reflextion achieves a 10-20\% performance increase over the baseline methods ReACT and CoT on datasets like AlfWorld \citep{alfworld}, HotPotQA \citep{hotpotqa}, and HumanEval \citep{chen2021evaluating}.

\citet{varshney2023stitch} introduce a comprehensive framework aimed at reducing factual inaccuracies. They use models to recognize entities and generate questions, we recognize these models as tools for the LLM-based agent. During the generation process, pivotal concepts encompassing names, geographical locales, and temporal references are ascertained within the contextual sentence employing entity extraction, keyword distillation, or directives to the LLM. The logit output values corresponding to these discerned concepts act as surrogates for confidence estimates. Should these values fall beneath a predetermined threshold, the mechanism then fetches a pertinent document to corroborate the generated information. The query methodology employed for such a retrieval hinges on posing a binary (Yes/No) query to the LLM. In scenarios where the validation is unsuccessful, the framework directs the model to rectify the erroneous output, either by omission or by substitution, drawing upon the knowledge from the consulted document. Empirical evaluations, specifically in the domain of article generation, underscore the efficacy of the delineated approach. Notably, factual error rates exhibited by GPT-3 witnessed a substantial decline, from 47.5\% to a mere 14.5\%, when subjected to this methodology. The diagnostic facet of their approach manifests an 80\% recall, and the rectification mechanism adeptly rectifies 57.6\% of the factually incorrect outputs that were accurately pinpointed.

Self-RAG \citep{asai2023selfrag} builds upon the Retrieval-Augmented Generation (RAG) framework by incorporating a self-reflective approach. This includes on-demand retrieval, prompting the LLM to consider whether the retrieved documents are relevant and supportive of the argument, thereby enhancing the factuality of the LLM's output. Self-RAG achieves this by having the model produce special tokens (i.e., reflection tokens) during the output process. The authors employ an end-to-end training method that enables the model to generate these special tokens. At the inference level, Self-RAG decodes a retrieval token for each input and each segment of the previously generated content. If the token is 'no,' the model proceeds with the normal output; if 'yes,' it performs a retrieval. The input, previous output, and each retrieved document are then fed into another session of the model, which assesses the relevance of each document. If a document is relevant, the model further evaluates whether it is supportive and should be used in generation. Based on this document, the model produces a segment and applies soft and hard constraints using the generated reflection token to select the most appropriate document. Self-RAG then incorporates the most suitable document into the continued generation of content, providing citations as needed. This process repeats until the entire content is generated. Self-RAG requires training data, which is annotated by GPT-4. Experiments based on the Llama-2 model on Short-form QA (PopQA \citep{PopQA}, TQA \citep{TQ}), Closed-set QA (Pub \citep{zhang2023interpretable}, ARC \citep{clark2018think}), and Long-form NLG with Citations (Bio Generation \citep{Min2023-FActScore}, ALCE-ASQA \citep{gao2023enabling}) have demonstrated its effectiveness.

\subsubsection{Retrieval Adaptation}
\label{sec:enhace_rag_adapt}

Recent research \citep{wang2023evaluating,ren2023investigating} has highlighted that merely using the retrieved information in LLMs doesn't always enhance their ability to answer factual questions. This underscores the importance of enabling LLMs to better adapt to the retrieved data to produce more accurate content. In this section, we delve into various strategies that facilitate this adaptation. Specifically, we explore three methodological approaches: prompt-based methods, SFT-based methods, and RLHF-based methods.

\paragraph{\textbf{Prompt-based}}
Leveraging prompts to navigate the retrieval process, ensuring the extraction of pertinent and factual data.

\citet{peng2023check} introduce the LLM-Augmenter system, a system using a fixed LLM combined with a plug-and-play retrieval module to help the LLM perform better in tasks that are particularly sensitive to factual errors. This system enhances its performance by enabling the LLM to use a series of modules (e.g. allowing the LLM to interact with external knowledge) to assist the LLM in generating results grounded in evidence. And they use automated feedback generated by utility functions (e.g., the factuality score of a LLM-generated response) to modify LLM's candidate response options. The author evaluates the system's performance on information-seeking dialog \citep{galley2019grounded} and Wiki QA \citep{OTT-QA} and experiments show that the system can significantly reduce ChatGPT's errors without sacrificing the fluency and informativeness of the generated content.

\paragraph{\textbf{SFT-based}}
Optimizing the LLM or retrieval system through training to enhance the alignment between generation tasks and the retrieved content.

\citet{Atlas} introduce a comprehensive architecture named ATLAS which is composed of the Contriever \citep{contriever} retriever of the dual-encoder architecture and the T5 \citep{t5} language model with Fusion-in-Decoder \citep{Fusion-in-Decoder}. The training objectives for the retriever consist of four components: Attention Distillation \citep{izacard_attention_dist}, where the retriever is trained on the average attention scores from the language model for each article; End-to-end training of Multi-Document Reader and Retriever (EMDR2) \citep{EMDR2}, which involves using the query and the top-K retrieved articles from the current retriever as input and loss computation against the standard answers to train the retriever; Perplexity Distillation (PDist), where the retriever is trained to predict how much the perplexity of standard answers would improve for each document; Leave-one-out Perplexity Distillation (LOOP), which trains the retriever to predict how much worse the prediction of the language model gets when removing a document from the top-K results. The LM's training objectives consist of three parts: prefix language modeling, masked language modeling, and title to section generation. Besides, they optimize and accelerate retriever training using techniques such as full index update, Re-ranking, and Query-side fine-tuning. Atlas achieves notable accuracy on Natural Questions using only 64 examples, outperforming a 540B model with 50x fewer parameters.

\citet{REPLUG} introduce REPLUG, a retrieval-augmented framework that considers the LLM as a black box, freezes its parameters and tunes retrieval models with supervision signals using language modeling scores. In this framework, the input context and the document are encoded through the dual encoder architecture,  cosine similarity is then calculated to retrieve related documents. The likelihood for each retrieved document and the language model scores are computed. Then they can update the retrieval model parameters by minimizing the KL divergence between retrieved document likelihood and the language model"s score distribution.  Ablation experiments demonstrate that this method significantly improves the performance of the original language models and the improvements are not coming from ensembling random documents.

\citet{luo2023sail} focus on utilizing instruction-tuning to denoise the retrieval results. They gather retrieval outcomes from various search APIs and domains, leading to the creation of a new search-grounded dataset. This dataset encompasses instructions, grounding information, and responses. Notably, it includes both pertinent results and those that are unrelated or disputed. The model needs to learn to ground on useful search results. After fine-tuning the LLaMa-7B model on this dataset, the resulting model, named SAIL-7B, exhibits superior performance in transparency-sensitive tasks such as open-ended QA and fact-checking.

\paragraph{\textbf{RLHF-based}}
\citet{GopherCite} use reinforcement learning from human preferences (RLHP) to train a 280 billion parameter model named GopherCite that generate answers along with high quality supporting evidence. They firstly collect data from existing models and have it rated by humans. The data is used for fine-tuning and reward model training. A supervised fine-tuning model is trained to produce accurate quotes with proper syntax. A reward model is created to rank model outputs based on overall quality. Finally, a reinforcement learning policy is optimized to align model behavior with human preferences, improving quoting performance. The model may decline to answer when the reward model score is too low. According to human evaluation, the model achieves better supported and plausible rating on the subset of Natural Questions dataset~\citep{NQ} than the previous SOTA (FiD-DPR) \citep{Fusion-in-Decoder}.


\subsubsection{Retrieval on External Memory}
\label{sec:enhace_rag_mem}

Currently, most LLMs enhance their factuality by retrieving knowledge in the form of text snippets from external storage and incorporating them into the context. Some researchers are exploring the storage of knowledge in non-textual forms and integrating this knowledge into models through specialized methods.

\citet{li2022decoupled} store knowledge in the form of key-value pairs in memory. The key is obtained by encoding knowledge using a Doc Retrieval Embedder, while the value is encoded using a Transformer encoder. Similar to traditional retrieval-based LLMs, the model encodes the input using a Query retrieval embedder and retrieves knowledge from memory. The retrieved value is then integrated into the model's multi-head attention layer through cross-attention to enhance factuality.

G-MAP \citep{wan2022gmap} does not explicitly store knowledge in storage but uses a general domain PLM as external memory. To mitigate catastrophic forgetting during adaptive pretraining, G-MAP introduces a frozen-parameter general domain PLM (PLM-G) during the fine-tuning of the domain-specific PLM (PLM-D). During fine-tuning, the input is provided to both PLM-G and PLM-D. The hidden states from each layer of PLM-G are stored in a cache, and a Memory-Augmented Strategy is used to extract hidden states from certain layers, which are then concatenated and integrated into PLM-D's Memory-Augmented Layer. The study also compared four Memory-Augmented Strategies, with Chunk-based Gated Memory Transfer performing the best.

Fine-tuning methods and some model editing methods store new knowledge in new model parameters through continual pretraining. The difference is that fine-tuning methods store many pieces of knowledge in a matrix parameter, while model editing establishes a new neuron for each piece of knowledge.

\citet{houlsby2019parameterefficient} propose to add Adapter modules to fine-tune pre-trained deep learning models on a new task with minimal changes to the original model by inserting small, trainable modules between existing layers. During fine-tuning, the main body of the pre-trained model is frozen, and the Adapter module learns knowledge specific to downstream tasks. The Adapter method reduces the computational requirements for model fine-tuning while enhancing the model's factuality for specific domains. However, the addition of the Adapter module also increases the overall parameter count of the model, somewhat reducing the model's inference performance.

To enhance the model's understanding of entities and thereby improve factuality, KALA \citep{KALA}, EaE \citep{EaE}, and Mention Memory \citep{jong2022mention} store encoded entities in external memory. During generation, the retrieved entity embeddings are integrated into the model layers.

\citet{KALA} introduced KALA to reduce overhead and catastrophic forgetting during Adaptive Training. KALA not only establishes a memory for entities and their encodings but also uses a KG to store relationships between these entities. For a given mention in the input, the corresponding entity is first determined. Based on the KG, the encoding of this entity and its neighboring entities is retrieved from memory. Through GNN weighted aggregation, the encoding for this mention's corresponding entity is obtained. Finally, the Knowledge-conditioned Feature Modulation (KFM) is introduced in the model layer, integrating the encoding result into the representation of all tokens involved in the mention.

\citet{EaE} introduce mention detection, entity linking, and MLM during model training. The model queries the top 100 entity embeddings from the entity storage that is closest to the current mention and integrates them using attention.

\citet{jong2022mention}'s TOME model is an improvement over the EaE model. Instead of storing entity embeddings in memory, TOME stores the embeddings of entity mentions. For marked entity mentions in the input, TOME retrieves all related entity mention embeddings from memory and integrates them into the model through a memory attention layer.

Similar to the three methods mentioned above, knowledge plugin \citep{zhang-etal-2023-plug} also introduce entity-related knowledge. However, instead of integrating the knowledge directly into the model layers, they utilize a pre-trained mapping network. This network maps the entity embeddings to the token embedding space of the Pre-trained Language Model (PLM). Ultimately, the mapped entity embeddings are injected at the input embedding level, facilitating the knowledge insertion process.

\citet{xue2023improving} address the challenge of improving factual consistency in knowledge-grounded dialogue systems by introducing extended feed-forward networks (FFNs) and leveraging reinforcement learning, resulting in more accurate and reliable responses, as demonstrated on the WoW \citep{WoW} and CMU\_DoG \citep{cmu_dog_emnlp18} datasets.


To further tackle the factual correction task \citep{thorne2021evidence}, \citet{gao2023rarr} explore the integration of LLMs, such as GPT-3, with search engines to improve their precision and memory. The goal is to use search engines to search for evidence and correct sentences generated by LLMs. The proposed method RARR is that, for each input sentence, a set of questions is generated, and web pages are searched to verify the consistency of information with the input sentence. The paper evaluates the modifications based on attribution and preservation criteria, with both manual and automatic verification methods. The primary evaluation metric is F1, considering both attribution and preservation aspects to assess the effectiveness of the approach in enhancing LLM-generated sentences.

\citet{chen2023purr} is a follow-up work to \citep{gao2023rarr} and \citep{thorne2021evidence}. Similar to EFEC, this paper fine-tunes a T5 model to serve as an editor, but it introduces negative samples during fine-tuning. The model PURR is trained to take user questions, perform Google searches to retrieve the top 5 web page summaries (used as positive samples), and generate noise by replacing some content in these positive samples using the language model. A sequence-to-sequence model is then trained to correct the noisy sentences back to their correct versions. This approach differs from EFEC, which used a mask-and-fill approach. PURR represents an improvement over EFEC, focusing on directly training a language model to edit incorrect sentences into correct ones using Google search for generating positive samples, ultimately leading to increased F1 scores.

\subsubsection{Retrieval on Structured Knowledge Source}
\label{sec:enhace_rag_kg}

We discuss studies that retrieves on structured repositories, such as knowledge graphs and databse, to source factual data during generation in this subsubsection.

\citet{zhang2023mitigating} utilize knowledge graphs (KG) for retrieval to tackle factual errors. They observe that there can be inconsistencies between a user's request and the content in the KG. For instance, when a user mentions a full name, the KG might only have its abbreviation, leading to imperfect retrieval results. To rectify this, they propose a method to rephrase the user's request. Their approach involves generating an SQL query based on the user's input and the database metadata using a LLM. And then they query the database and ask the LLM to identify which entity in the sentence corresponds to an entity in the database, thereby creating a mapping. Using the entity names from the database, the LLM is prompted to reformulate the question. If a database query results in multiple rows for a selected column and item, a new question is generated using a greedy approach, prompting the user for more specific details until a conclusive answer is reached. Experiments show that this method exhibits notable enhancements in comparison to contemporary state-of-the-art techniques in mitigating language model inaccuracies.

StructGPT \citep{StructGPT} is a general prompt framework to support LLMs reasoning on structured data (e.g., KG, Table, and Database).
In general, the core of this framework is that they construct the specialized interfaces to collect relevant evidence from structured data (i.e., reading), and let LLMs concentrate on the reasoning task based on the collected information (i.e., reasoning). 
Specially, they propose an invoking-linearization-generation procedure to support LLMs in reasoning on the structured data with the help of the interfaces.
By iterating this procedure with provided interfaces, our approach can gradually approach the target answers to a given query.
Experiments conducted on three types of structured data, including KGQA, TableQA, and Text-to-SQL, show that StructGPT greatly improves the performance of LLMs, under the few-shot or zero-shot settings.

\citet{baek2023knowledgeaugmented} propose to inject the factual knowledge from knowledge graphs into (large) language models (up to GPT-3.5), by retrieving the relevant facts from knowledge graphs based on their textual similarities with the input question and then injecting them as the prompt of language models. This approach improves the performance of language models on knowledge graph question answering tasks \citep{Mintaka} by up to 48\% on average, compared to baselines without knowledge graphs.

\subsection{Domain Factuality Enhanced LLMs}
\label{sec:domain_llms}

Domain Knowledge Deficit is not only an important reason for limiting the application of LLM in specific fields, but also a subject of great concern to both academia and industry. In this subsection, we discuss how those Domain-Specific LLMs enhance their domain factuality.

\begin{table*}[]\small
\centering
\caption{LLMs Enhanced for Domain-Specific Factuality. In the `domain' column, we utilize the following abbreviations: healthcare/medicine (H), finance (F), law/legal (L), geoscience/environment (G), education (E), food testing (FT), and home renovation (HR).}
\label{tab:domain_llms}
\vspace{-2mm} \begin{adjustbox}{max width=0.95\textwidth}
  \setlength{\tabcolsep}{1.mm}
{
\begin{tabular}{lllllcccc}
\toprule
Reference &
  \begin{tabular}[c]{@{}l@{}}Do- \\main\end{tabular} &
  Base-Model &
  \begin{tabular}[c]{@{}l@{}}Eval\\Task\end{tabular}&
  \begin{tabular}[c]{@{}l@{}}Eval\\Dataset\end{tabular} &
  \begin{tabular}[c]{@{}l@{}}Continual \\ Pretrained?\end{tabular} &
  \begin{tabular}[c]{@{}l@{}}Continual \\SFT?\end{tabular} &
  \begin{tabular}[c]{@{}l@{}}Train From \\ Scratch?\end{tabular} &
  \begin{tabular}[c]{@{}l@{}}External \\ Knowledge\end{tabular} \\

  \midrule
HuatuoGPT \citep{zhang2023huatuogpt} &
  H &
  \begin{tabular}[c]{@{}l@{}}Baichuan-7B,  \\Ziya-LLaMA-13B\end{tabular}&
  QA &
  \begin{tabular}[c]{@{}l@{}}cMedQA2,\\ WebMedQA,\\ Huatuo-26M\end{tabular} &
  \checkmark &
   &
   &
   \\
Zhongjing \citep{yang2023zhongjing} &
  H &
  Ziya-LLaMA-13B &
  QA &
   \begin{tabular}[c]{@{}l@{}}CMtMedQA, \\huatuo-26M\end{tabular} &
  \checkmark &
  \checkmark &
   &
   \\
\citet{wang2023augmenting} &
  H &
  \begin{tabular}[c]{@{}l@{}}GPT-3.5-Turbo, \\ LLaMA-2-13B\end{tabular} &
  QA &
  \begin{tabular}[c]{@{}l@{}}MedQAUSMLE,\\ MedQAMCMLE,\\ MedMCQA\end{tabular} &
   &
   &
   &
  \checkmark \\
MOLFORMER \citep{ross2022large} &
  H &
  MOLFORMER &
  \begin{tabular}[c]{@{}l@{}}Molecule\\ properties \\ prediction\end{tabular} &
   &
   &
   &
  \checkmark &
   \\
DISC-MedLLM \citep{bao2023disc} &
  H &
  Baichuan-13B &
  QA &
  \begin{tabular}[c]{@{}l@{}}CMB-Clin, \\ CMD, \\ CMID\end{tabular} &
   &
  \checkmark &
   &
   \\
CohortGPT \citep{guan2023cohortgpt} &
  H &
  ChatGPT &
  \begin{tabular}[c]{@{}l@{}}IU-RR, \\ MIMIC-CXR\end{tabular} &
   &
   &
   &
   &
  \checkmark \\
DeID-GPT \citep{liu2023deid} &
  H &
  GPT-4 &
  \begin{tabular}[c]{@{}l@{}}Medical \\ Text \\ De-Identification\end{tabular} &
   &
   &
   &
   &
  \checkmark \\

ChatDoctor \citep{li2023chatdoctor} &
  H &
  LLaMA &
  QA &
   &
   &
  \checkmark
   &
   & \\
BioMedLM \citep{venigalla2022biomedlm} &
  H &
  GPT (2.7b) &
  QA &
   &
   &
   &
  \checkmark &
   \\
DoctorGLM \citep{xiong2023doctorglm} &
  H &
  ChatGLM-6B &
  QA &
   &
   &
  \checkmark &
   &
   \\
MedChatZH \citep{tan2023medchatzh} &
  H &
  Baichuan-7B &
  QA &
  C-Eval, MMLU &
   &
  \checkmark &
   &
   \\
BioGPT \citep{luo2022biogpt} &
  H &
  GPT-2 &
  QA, DC, RE &
   &
   &
   &
  \checkmark &
   \\
GeneGPT \citep{jin2023genegpt} &
  H &
  Codex &
  QA &
  GeneTuring &
   &
   &
   &
  \checkmark \\
Almanac \citep{hiesinger2023almanac} &
  H &
  text-davinci-003 &
  QA &
  ClinicalQA &
   &
   &
   &
  \checkmark \\
MolXPT \citep{liu2023molxpt} &
  H &
  GPT-2medium &
  \begin{tabular}[c]{@{}l@{}}Molecular \\ Property \\ Prediction,  \\ Molecule-text \\ translation\end{tabular} &
   &
   &
  \checkmark &
  \checkmark &
   \\
LawGPT \citep{nguyen2023brief} &
  L &
  GPT3 &
   &
   &
   &
  \checkmark &
   &
   \\
\citet{savelka2023explaining} &
  L &
  GPT-4 &
   &
   &
   &
   &
   & \checkmark
   \\
Lawyer LLaMA \citep{huang2023lawyer} &
L &
LLaMA &
CN Legal Tasks &
&
\checkmark &
\checkmark &
& 
\\
ChatLaw \citep{cui2023chatlaw} &
  L &
  Ziya-LLaMA-13B &
  QA &
  \begin{tabular}[c]{@{}l@{}}national judicial \\ examination \\  question\end{tabular} &
  \checkmark &
   &
   &
  \checkmark \\
EcomGPT \citep{li2023ecomgpt} &
  F &
  BLOOMZ &
  \begin{tabular}[c]{@{}l@{}}4 major tasks\\  12 subtasks\end{tabular} &
  EcomInstruct &
   &
  \checkmark &
   &
   \\
BloombergGPT \citep{BloombergGPT} &
  F &
BLOOM &
  \begin{tabular}[c]{@{}l@{}}Financial NLP \\ (SA, BC, \\NER, \\NER+NED, QA)\end{tabular} &
 \begin{tabular}[c]{@{}l@{}} Financial\\Datasets \end{tabular}&
   &
   &
  \checkmark &
   \\
K2 \citep{deng2023learning} &
  G &
  LLaMA-7B &
   &
  GeoBench &
  \checkmark &
   &
   &
   \\
HouYi \citep{bai2023houyi} &
  G &
  ChatGLM-6B &
   &
   &
  \checkmark &
   &
   &
   \\
OceanGPT \citep{bi2023oceangpt} &
  G &
  LLaMA-2-7B &
   &
  OceanBench &
  \checkmark &
   \checkmark&
   &\checkmark
   \\
GrammarGPT \citep{fan2023grammargpt} &
  E &
  phoenix-inst-chat-7b &
  \begin{tabular}[c]{@{}l@{}}Chinese \\ Grammatical Error\\ Correction\end{tabular} &
  \begin{tabular}[c]{@{}l@{}}ChatGPT-\\generated,\\ Human-\\annotated\end{tabular} &
   &
  \checkmark &
   &
  \\
FoodGPT  \citep{qi2023foodgpt} &
  FT &
  Chinese-LLaMA2-13B &
  QA &
   &
  \checkmark &
   &
   &  \checkmark 
   \\

ChatHome  \citep{wen2023chathome} &
  HR &
  Baichuan-13B &
   &
   \begin{tabular}[c]{@{}l@{}}
   C-Eval,\\ CMMLU, \\EvalHome \end{tabular}
   &
   &\checkmark
   &
   &  
   \\
  \bottomrule
\end{tabular}}
\end{adjustbox}
\end{table*}
Table~\ref{tab:domain_llms} lists the domain-factuality enhanced LLMs. Here, we include several domains, including healthcare/medicine (H), finance (F), law/legal (L), geoscience/environment (G), education (E), food testing (FT), and home renovation (HR). 

Based on the actual scenarios of Domain-Specific LLMs and our previous categorization of enhancement methods, we have summarized several commonly used enhancement techniques for Domain-Specific LLMs: 

\emph{(1) Continual Pretraining: } A method that involves continuously updating and fine-tuning a pre-trained language model using domain-specific data. This process ensures that the model stays up-to-date and relevant within a specific domain or field. It starts with an initial pre-trained model, often a general-purpose language model, and then fine-tunes it using domain-specific text or data. As new information becomes available, the model can be further fine-tuned to adapt to the evolving knowledge landscape. Continual pretraining is a powerful approach for maintaining the accuracy and relevance of AI models in rapidly changing domains, such as technology or medicine \citep{zhang2023huatuogpt,yang2023zhongjing}.

\emph{(2)
Continual SFT: }  Another strategy for enhancing the factuality of AI models. In this approach, the model is fine-tuned using labeled or annotated data specific to the domain of interest. This fine-tuning process allows the model to learn and adapt to the nuances and specifics of the domain, improving its ability to provide accurate and contextually relevant information. It can be particularly useful in applications where access to domain-specific labeled data is available over time, such as in the case of legal databases, medical records, or financial reports \citep{bao2023disc,li2023chatdoctor}.

\emph{(3) Train From Scratch:} It involves starting the learning process with minimal prior knowledge or pretraining. This approach can be likened to teaching a machine learning model with a blank slate. While it may not have the advantage of leveraging pre-existing knowledge, training from scratch can be advantageous when dealing with completely new domains or tasks where there is limited relevant data available. It allows the model to build its understanding from the ground up, although it may require substantial computational resources and time \citep{ross2022large,venigalla2022biomedlm}.

\emph{ (4)
External knowledge:} involves augmenting a language model's internal knowledge with information from external sources. This method allows the model to access databases, websites, or other structured data repositories to verify facts or gather additional information when responding to user queries. By integrating external knowledge, the model can enhance its fact-checking capabilities and provide more accurate and contextually relevant answers, especially when dealing with dynamic or rapidly changing information. Below, we introduce these methods \citep{wang2023augmenting,fan2023grammargpt}.

For each Domain-specific LLM, we list its respective enhancement methods, which are presented in Table \ref{tab:domain_llms}.

\subsection{Healthcare domain-enhanced LLMs} 
These LLMs have emerged as powerful tools in the medical field, offering a diverse range of capabilities. These models, such as CohortGPT\citep{guan2023cohortgpt}, ChatDoctor\citep{li2023chatdoctor}, DeID-GPT\citealp{liu2023deid}, BioMedLM\citealp{venigalla2022biomedlm}, DoctorGLM\citealp{DoctorGLM}, MedChatZH\citealp{tan2023medchatzh}, BioGPT\citep{luo2022biogpt}, GeneGPT\citep{jin2023genegpt}, Almanac\citep{hiesinger2023almanac}, and MolXPT\citep{liu2023molxpt}, harness the potential of LLMs to revolutionize healthcare. They are equipped with features like classifying unstructured medical text into disease labels, improving performance with knowledge graphs and sample selection strategies, fine-tuning on large datasets of patient-doctor dialogues, enabling automatic medical text de-identification, excelling in medical question-answering tasks, handling traditional Chinese medical question-answering, and outperforming in various biomedical NLP tasks. Some models interact with web APIs for genomics questions, while others specialize in clinical guidelines and treatment recommendations. These LLMs not only demonstrate state-of-the-art performance on healthcare domain but also emphasize the importance of domain-specific training and evaluation, showcasing their potential in transforming healthcare and clinical decision-making.

\citet{zhang2023huatuogpt} present HuatuoGPT, a medical language model that uses data from ChatGPT and doctors, resulting in state-of-the-art performance in medical consultations. It is based on Baichuan-7B and Ziya-LLaMA-13B-Pretrain-v1, continually pre-trained on both distilled data (from ChatGPT) and real-world data (from Doctors). 

Likewise, \citet{yang2023zhongjing} introduce Zhongjing, the first Chinese medical language model based on LLaMA, which utilizes a comprehensive training pipeline and a multi-turn medical dialogue dataset. Specifically, it is enhanced with a multi-turn medical dialogue dataset called CMtMedQA, consisting of 70,000 authentic doctor-patient dialogues, enabling complex dialogue and proactive inquiry. The backbone model used is Ziya-LLaMA-13B-v1, and the evaluation dataset is CMtMedQA and huatuo-26M~\citep{li2023huatuo}. 

\citet{wang2023augmenting}
unveil a system, LLM-AMT, that improves large-scale language models like GPT-3.5-Turbo and LLaMA-2-13B  with medical textbooks, notably enhancing open-domain medical question-answering tasks. At the same time, the external knowledge source is a Hybrid Textbook Retriever comprising 51 textbooks from the MedQA dataset and Wikipedia.


\citet{bao2023disc} present DISC-MedLLM, a solution that uses LLMs to provide accurate medical responses in conversational healthcare services, utilizing strategies like medical knowledge graphs, real-world dialogue reconstruction, and human-guided preference rephrasing to create high-quality SFT datasets, applied on Baichuan-13B-Base. The paper uses various datasets for fine-tuning, including Re-constructed AI Doctor-Patient Dialogue, MedDialog, cMedQA, Knowledge Graph QA pairs (CMeKG), Behavioral Preference Dataset (Manual selection), MedMCQA, MOSS6, and Alpaca-GPT.

Similarly, \citet{guan2023cohortgpt} introduce CohortGPT, a model that uses LLMs for participant recruitment in clinical research by classifying complex medical text into disease labels. CohortGPT enhances ChatGPT performance with the use of a knowledge graph as auxiliary information and a CoT sample selection strategy. The tasks involve IU-RR (Preparing a collection of radiology examinations for distribution and retrieval) and MIMIC-CXR (Mimic-cxr, a de-identified publicly available database of chest radiographs with free-text reports). \citet{li2023chatdoctor} introduce ChatDoctor, a refined LLaMA-based model. It is fine-tuned using a dataset of 100,000 patient-doctor dialogues, and equipped with a self-directed information retrieval mechanism.

\citet{liu2023deid} introduce DeID-GPT, a framework leveraging GPT-4 for automatic medical text de-identification. Additionally, the paper mentions the use of HIPAA identifiers as an extra knowledge source to enhance the de-identification process.

\citet{venigalla2022biomedlm} present BioMedLM, a domain-specific LLM trained on PubMed data, for medical QA tasks. \citet{xiong2023doctorglm} introduce DoctorGLM, a Chinese-focused language model fine-tuned for healthcare-specific tasks. Both \citet{tan2023medchatzh} and \citet{luo2022biogpt} introduce dialogue and generative Transformer language models for traditional Chinese medical question-answering and biomedical NLP tasks, respectively.

\citet{jin2023genegpt} present GeneGPT, a method for teaching LLMs to answer genomics questions using NCBI Web APIs. 
\citet{hiesinger2023almanac} introduce Almanac, a LLM with retrieval capabilities for medical guideline recommendations. Lastly, \citet{liu2023molxpt} introduce MolXPT, a unified language model adept in molecular property prediction and molecular generation.

\subsection{Legal domain enhanced LLMs} These LLMs, such as LawGPT \citep{nguyen2023brief}, and ChatLaw \citep{ChatLaw}, have been fine-tuned to provide comprehensive legal assistance, from answering intricate legal queries and generating legal documents to offering expert legal advice. Leveraging extensive corpora of legal text, these models ensure context-aware and accurate responses. Moreover, their continual development involves injecting domain knowledge, designing supervised fine-tuning tasks, and incorporating retrieval modules to address issues like hallucination and ensure high-quality legal assistance. These innovations not only pave the way for more accessible and reliable legal services but also open up new avenues for research and exploration within the legal domain.

\citet{nguyen2023brief} 
introduce LawGPT 1.0, a fine-tuned GPT-3 language model for the legal domain to provide conversational legal assistance, including answering legal questions, generating legal documents, and offering legal advices.
 The paper mentions the use of a large corpus of legal text for fine-tuning the model to adapt it to the legal domain.

\citet{savelka2023explaining}
 evaluate the performance of GPT-4 in generating explanations of legal terms in legislation, comparing a baseline approach to an augmented approach that uses a legal information retrieval module to provide context from case law, revealing improvements in quality and addressing issues of factual accuracy and hallucination.

\citet{huang2023lawyer} address the challenge of enhancing LLMs like LLaMA for domain-specific tasks, particularly in the legal domain, by injecting domain knowledge during continual training, designing appropriate supervised finetune tasks and incorporating a retrieval module to improve factuality during text generation. They release their data and model for further research in Chinese legal tasks.

\citet{cui2023chatlaw} introduce ChatLaw, an open-source legal LLM, designed for the Chinese legal domain. The paper introduces a method to improve model factuality during data screening, and a self-attention method for error handling.
The paper uses various datasets for fine-tuning ChatLaw, including a collection of original legal data, data constructed based on legal regulations and judicial interpretations, and crawled real legal consultation data.
The primary model used in this paper is Ziya-LLaMA-13B, which serves as the backbone for ChatLaw, tailored for the Chinese legal domain and optimized to handle legal questions and tasks. Additionally, the paper uses of a vector database retrieval method, keyword retrieval, and a self-attention method to enhance the model's performance in the legal domain.

\subsection{Finance Domain-enhanced LLMs} These LLMs combine sophisticated language models designed specifically for commercial and financial tasks to deliver robust processing capabilities. They focus on creating tailor-made solutions optimized for both financial text analysis and e-commerce settings, trained on datasets containing myriad business-related tasks and copious financial tokens. They are designed to perform a plethora of functions, ranging from understanding and generating instructions for various E-commerce assignments to identifying sentiment, recognizing named entities, and answering questions in financial contexts. The models are further fine-tuned for zero-shot generalization on diverse tasks and benchmarks. 

\citet{li2023ecomgpt} introduce EcomGPT, a language model tailored for E-commerce scenarios, trained on the newly created EcomInstruct dataset, which consists of 2.5 million instruction data spanning various E-commerce tasks and data types.
The dataset covers product information, user reviews, and more. It defines atomic tasks and Chain-of-Task tasks to enable comprehensive training for E-commerce scenarios.
The backbone model used is BLOOMZ, which is fine-tuned with the EcomInstruct dataset. The evaluation dataset includes 12 tasks, encompassing classification, generation, extraction, and other E-commerce-related tasks.

\citet{BloombergGPT} introduce BloombergGPT, a specialized 50 billion-parameter language model for the financial domain, trained on a massive 363 billion token dataset, which combines Bloomberg's extensive financial data sources with general-purpose datasets.
 The dataset used in this paper is an extensive 363 billion token dataset, which includes a significant portion of financial data from Bloomberg's sources (51.27\% of the training data).
  BloombergGPT is based on a decoder-only causal language model architecture known as BLOOM. The evaluation includes various financial NLP tasks such as sentiment analysis, named entity recognition, binary classification, and question answering.
  

\subsection{Other Domain-Enhanced LLMs}

\paragraph{Geoscience and  Environment domain-enhanced LLMs} are expertly designed, leveraging vast corpora to provide precise and robust results pertaining to geoscience and renewable energy. K2, a trailblazer in geoscience LLM, was trained on a massive geoscience text corpus and further refined using the GeoSignal dataset. Meanwhile, the HouYi model, another pioneering LLM focusing on renewable energy, harnessed the Renewable Energy Academic Paper dataset, containing over a million academic literature sources. These LLMs are fine-tuned to deliver adept performance in their respective fields, showing substantial capabilities in aligning their responses with user queries and renewable energy academic literature. 

\citet{deng2023learning}
 introduce K2, the first LLM designed specifically for geoscience, which is a LLaMA-7B continuously trained on a 5.5 billion token geoscience text corpus and fine-tuned using the GeoSignal dataset.
 The paper also presents resources like GeoSignal, a geoscience instruction tuning dataset, and GeoBench, the first geoscience benchmark for evaluating LLMs in the context of geoscience.
 
\citet{bai2023houyi} present the development of the HouYi model, the first LLM specifically designed for renewable energy, utilizing the newly created Renewable Energy Academic Paper (REAP) dataset, which contains over 1.1 million academic literature sources related to renewable energy, and the HouYi model is fine-tuned based on general LLMs such as ChatGLM-6B.

\citet{bi2023oceangpt} present OceanGPT, the first-ever LLM in the ocean domain, which is expert in various ocean science tasks. They also propose  a novel framework called DoInstruct to automatically obtain a large volume of ocean domain data.  OceanGPT is evaluated in OceanBench and  shows a higher level of knowledge expertise for oceans science tasks.

\paragraph{Education domain-enhanced LLMs} are used for assisting education scenarios. An example is GrammarGPT \citep{fan2023grammargpt}, which provides an innovative approach to language learning, particularly focusing on error correction in Chinese grammar. It is an open-source LLM designed for native Chinese grammatical error correction, which leverages a hybrid dataset of ChatGPT-generated and human-annotated data, along with heuristic methods to guide the model in generating ungrammatical sentences. The backbone model used is phoenix-inst-chat-7b.

\paragraph{Food domain-enhanced LLMs} are language models specifically designed to meet the distinct requirements of food testing protocols. For example, \citet{qi2023foodgpt} introduce FoodGPT, a LLM for food testing that incorporates structured knowledge and scanned documents using an incremental pre-training approach, with a focus on addressing machine hallucination by constructing a knowledge graph as an external knowledge base, utilizing the Chinese-LLaMA2-13B as the backbone model and collecting food-related data for training.

\paragraph{Home renovation domain-enhanced LLMs} are domain-specific language models tailored for home renovation tasks. For example, 
 \citet{wen2023chathome} introduce ChatHome, which uses a dual-pronged methodology involving domain-adaptive pretraining and instruction-tuning on an extensive dataset comprising professional articles, standard documents, and web content relevant to home renovation. The backbone model is Baichuan-13B, and the evaluation datasets include C-Eval, CMMLU, and the newly created "EvalHome" domain dataset, while the fine-tuning data sources encompass National Standards, Domain Books, Domain Websites, and WuDaoCorpora.







\section{Conclusion}
Throughout this survey, we have systematically explored the intricate landscape of factuality issues within large language models (LLMs). We began by defining the concept of factuality (Sec \ref{sec:definition}) and proceeded to discuss its broader implications (Sec \ref{sec:impact}). Our journey took us through the multifaceted realm of factuality evaluation, encompassing benchmarks (Sec \ref{sec:eval_benchmark}), metrics (Sec \ref{sec:eval_metrics}), specific evaluation studies (Sec \ref{sec:factuality-eval}), and domain-specific evaluations (Sec \ref{sec:domain-eval}). We then delved deeper, probing the intrinsic mechanisms that underpin factuality in LLMs (Sec \ref{sec:analysis}). Our exploration culminated in the discussion of enhancement techniques, both for standalone LLMs (Sec \ref{sec:enhace_pure}) and retrieval-augmented LLMs (Sec \ref{sec:enhace_rag}), with a special focus on domain-specific LLM enhancements (Sec \ref{sec:domain_llms}).

Despite the advancements detailed in this survey, several challenges loom large. The evaluation of factuality remains an intricate puzzle, complicated by the inherent variability and nuances of natural languages. The core processes governing how LLMs store, update, and produce facts are yet not fully revealed. And while certain techniques, like continual training and retrieval, show promise, they are not without limitations.
Looking ahead, the quest for fully factual LLMs presents both challenges and opportunities. Future research might delve deeper into understanding the neural architectures of LLMs, develop more robust evaluation metrics, and innovate on enhancement techniques. As LLMs become increasingly integrated into our digital ecosystem, ensuring their factual reliability will remain paramount, with implications that impact across the AI community and beyond.

\bibliographystyle{ACM-Reference-Format}
\bibliography{references, references_domain}












\end{document}